\documentclass[journal]{IEEEtran}
\usepackage[english]{babel}
\usepackage{microtype}
\ifCLASSINFOpdf
\else
\fi
\usepackage{url}        
\usepackage{xcolor}
\usepackage[colorlinks=false,linkbordercolor=red,citebordercolor=green,urlbordercolor=cyan]{hyperref}
\renewcommand{\UrlFont}{\color{blue}}
\makeatletter
\renewcommand{\url@samestyle}{\def\UrlFont{\color{blue}}}
\makeatother
\usepackage{booktabs}
\usepackage{makecell}
\usepackage{siunitx}
\usepackage{multirow}
\usepackage{array} 
\usepackage{tabularx}
\usepackage{threeparttable}
\usepackage{graphicx} 
\usepackage[cmex10]{amsmath}
\usepackage{mathtools}

\usepackage{amsfonts} 
\allowdisplaybreaks
\begin{document}
\raggedbottom
%
\title{Hurdle--RMIL: Addressing Zero Inflation and Long-Tailed Imbalance in Infrared Rainfall Retrieval}
%
%


\author{Fangjian~Zhang,
        Xiaoyong~Zhuge,
        Wenlan~Wang,
        Haixia~Xiao,
        Yuying~Zhu,
        Siyang~Cheng,
        and~Ali~Mamtimin
\thanks{Corresponding author: Xiaoyong Zhuge.}%
\thanks{Fangjian Zhang, Xiaoyong Zhuge, Wenlan Wang, Haixia Xiao, and Yuying Zhu are with the Nanjing Innovation Institute for Atmospheric Sciences, Chinese Academy of Meteorological Sciences-Jiangsu Meteorological Service, Nanjing 210041, China, and also with the Jiangsu Key Laboratory of Severe Storm Disaster Risk / Key Laboratory of Transportation Meteorology of CMA, Nanjing 210041, China (e-mail: zhangfj@cma.gov.cn; zhugexy@cma.gov.cn).}%
\thanks{Siyang Cheng is with the Institute of Tibetan Plateau Meteorology, Chinese Academy of Meteorological Sciences, Beijing 100081, China, and also with the Heavy Rain and Drought-Flood Disasters in Plateau and Basin Key Laboratory of Sichuan Province, Institute of Tibetan Plateau Meteorology, China Meteorological Administration, Chengdu 610213, China.}
\thanks{Ali Mamtimin is with the Institute of Desert Meteorology, China Meteorological Administration, Urumqi 830002, China; the National Field Observation and Research Station (Xinjiang Taklimakan) for Desert Meteorology, Urumqi 830002, China; the Xinjiang Key Laboratory of Desert Meteorology and Sandstorm, Urumqi 830002, China; the Taklimakan Desert Meteorology Field Experiment Station of China Meteorological Administration, Urumqi 830002, China; and the Tazhong National Climate Observatory, Qiemo 841999, China.}%
}

\maketitle

\begin{abstract}
    
    \normalfont Imbalanced labels can cause frequent samples to dominate AI-based quantitative remote sensing, degrading rare-event retrieval. In rain-rate retrieval based on satellite infrared brightness temperatures, this imbalance leads to systematic underestimation of rare high-intensity rainfall. In this study, Hurdle--Retrieval Model Imbalanced Learning (RMIL) is proposed. Following a divide-and-conquer strategy, Hurdle--RMIL separates zero inflation from the long-tailed distribution of positive rain. A hurdle model handles zero inflation, whereas RMIL exploits invariance under fixed observation conditions of the rainfall-to-satellite forward process to derive a Bayes-based transformation linking conditional distributions under naturally long-tailed and hypothetical balanced rainfall. This transformation enables the balanced-distribution model to be learned from natural samples without constructing a balanced dataset. Comparisons with conventional learning, classification--regression modeling, cost-sensitive learning, and generative learning using test data from multiple regions in China show that Hurdle--RMIL mitigates systematic underestimation and improves detection of rare high-intensity and extreme rainfall without markedly degrading lower-threshold accuracy. At 0.1--10 $\mathrm{mm}\cdot\mathrm{h}^{-1}$, its root mean square error remains close to those of the best baselines, and it yields the highest equitable threat score (ETS) at most evaluated thresholds, with its advantage becoming more pronounced at high thresholds. At 30 $\mathrm{mm}\cdot\mathrm{h}^{-1}$, its ETS is 0.051 versus 0.015 for the best baseline, and its mean error is $-25.41$ $\mathrm{mm}\cdot\mathrm{h}^{-1}$ versus $-28.98$ $\mathrm{mm}\cdot\mathrm{h}^{-1}$. Case studies further show improved representations of rainfall intensity and spatial extent, demonstrating that Hurdle--RMIL effectively addresses rainfall-distribution imbalance and improves the retrieval of rare high-intensity rainfall.
\end{abstract}


%

\section{Introduction}
\IEEEPARstart{I}n recent years, artificial intelligence (AI) has been applied widely to quantitative remote sensing (QRS), substantially advancing the field \cite{yuanDeepLearningEnvironmental2020}. However, when the label distribution is imbalanced, conventional learning favors common samples while underrepresenting rare ones, which can result in underfitting of AI models for rare events \cite{renBalancedMseImbalanced2022}. The distributions of most environmental variables are inherently imbalanced, meaning that this problem is widespread in AI-based QRS tasks. For example, conventionally trained AI-based retrieval models often misestimate extreme values, underestimating high concentrations of atmospheric particulate matter and trace gases \cite{liRobustDeepLearning2020,liSpatiotemporalEstimationSatelliteborne2021a}, high land-surface temperatures \cite{s19132987}, deep snow \cite{weiFineResolutionSnowDepth2022}, and high-intensity rainfall \cite{taoDeepNeuralNetworks2016a}. In QRS, rare samples often correspond to impactful events, including haze events driven by high concentrations of particulate matter, heat waves associated with extreme temperature, and flood events triggered by high-intensity rainfall. Addressing this issue is crucial for improving the performance of AI-based QRS models in rare event retrieval, enhancing disaster monitoring capabilities, and advancing the development of AI applications in QRS.
Among the various environmental variables, rain displays one of the most pronounced imbalances, and studying rainfall retrieval will provide critical guidance for tackling similar challenges. This study focuses on rain-rate retrieval from satellite infrared brightness-temperature measurements. The broad spatial coverage of satellite observations makes this task valuable for rainfall monitoring.
Existing AI-based infrared rainfall retrieval algorithms that show promise in enhancing the retrieval of rare high-intensity rainfall can be broadly classified into three categories.

The first category is classification--regression modeling, which employs separate models for first detecting rain occurrence and then estimating rain rate \cite{taoDeepNeuralNetworks2016a,minEstimatingSummertimePrecipitation2019,huangPrecipitationEstimationUsing2024,wangFY4AAGRIInfrared2024,taoTwoStageDeepNeural2018,wangInfraredPrecipitationEstimation2020}.
In this arrangement, the occurrence model produces a hard rain/no-rain decision, and only this classification result is used to route samples to a separate rain-rate regression model; the occurrence probability itself is not retained for the subsequent estimation. This design aims to handle zero inflation—the severe imbalance arising from the overwhelming dominance of non-rain samples.
However, this process introduces a secondary challenge in the form of a long-tailed imbalance at the second stage, where the frequency of positive-rain samples decreases rapidly as rain rate increases.
Researchers have employed strategies such as data resampling, ensemble learning, and the incorporation of constraints into the loss function to counteract this issue \cite{minEstimatingSummertimePrecipitation2019,wangInfraredPrecipitationEstimation2020,Yun2026-mt}.
Despite these efforts, studies indicate that classification--regression modeling still exhibits substantial shortcomings in high-intensity rainfall retrieval.
Yang et al.~\cite{yangMultiTaskCollaborationDeep2021} argued that error propagation across the two stages may limit the effectiveness of classification--regression modeling.
To mitigate this limitation, they proposed the Multi-Task Collaboration Deep Learning Framework (MTCF), an enhanced classification--regression method. MTCF retains rain-occurrence classification and rain-rate regression as two distinct tasks but integrates them into a jointly trained, end-to-end multi-task framework, allowing the two tasks to mutually reinforce each other. Nevertheless, their method does not explicitly tackle the long-tailed imbalance, resulting in only a marginal 4\% improvement over conventional learning for rain exceeding 10 $\mathrm{mm} \cdot \mathrm{h}^{-1}$.


Another category is cost-sensitive learning, which reformulates the loss function. Ma et al.~\cite{maImprovementNearRealTimePrecipitation2022,maImprovedDeepLearningBasedPrecipitation2024} decomposed the mean squared error (MSE) into a non-rain term (observation $=0$) and a rain term (observation $>0$), with the latter further divided into underestimation (retrieval $<$ observation) and overestimation (retrieval $>$ observation) terms. Weights were assigned to each term to regulate model bias. Although this method improves the agreement between the frequency distributions of retrieval and observation, notable discrepancies persist for rain rates above 15 $\mathrm{mm}\cdot\mathrm{h}^{-1}$. Berthomier and Perier~\cite{berthomierEspressoGlobalDeep2023} proposed a sample-distribution-based weighted objective function that assigns weights according to rain rate, strengthening the model’s focus on higher rain rates. Although superior to operational products such as the Integrated Multi-satellitE Retrievals for GPM (IMERG), their method still underestimates high-intensity rainfall in high-latitude regions due to persistent imbalance and shows limited gains for the rarest and most intense events.


More recently, generative learning has emerged as a new direction \cite{guilloteauGenerativeDiffusionModel2025, PERSIANN-cGAN}. Guilloteau et al.~\cite{guilloteauGenerativeDiffusionModel2025} introduced a generative model for rainfall retrieval. Evaluations demonstrated the generative model’s superior ability to reconstruct the spatial fine structure of rain, which is a feature closely linked to higher-intensity rainfall. The generative model achieves a significantly higher retrieval rate for samples at higher rain rates compared to conventional learning. However, as the generative model does not inherently address the underlying imbalance, it fails to fully resolve the problem. This is evidenced by the cloud retrieval experiment \cite{Xiao2025-it}, in which cloud property super-resolution based on generative learning still severely underestimates extreme values.



This study develops Hurdle--Retrieval Model Imbalanced Learning (RMIL), an imbalanced learning framework. Here, imbalanced learning refers to learning from data in which certain target values have substantially fewer observations than others, with the aim of improving generalization across the entire target range, particularly for sparsely represented target values. In this study, these values correspond to rare high-intensity rainfall. The framework builds on the effective divide-and-conquer principle, which decomposes the overall rainfall imbalance into zero inflation and the long-tailed distribution of positive rain, thereby reducing the complexity of the learning problem. Classification--regression methods \cite{taoDeepNeuralNetworks2016a,minEstimatingSummertimePrecipitation2019,huangPrecipitationEstimationUsing2024,wangFY4AAGRIInfrared2024,taoTwoStageDeepNeural2018,wangInfraredPrecipitationEstimation2020} implement this principle directly by separating rain-occurrence detection from rain-rate estimation, whereas the loss-function decompositions of \cite{maImprovementNearRealTimePrecipitation2022,maImprovedDeepLearningBasedPrecipitation2024} constitute an indirect implementation. A hurdle model is a probabilistic mixture model that represents a zero-inflated conditional distribution as a probability mass at zero and a conditional distribution over positive values. It has been applied to various types of zero-inflated data, including rainfall forecasting \cite{wilsonPointPredictionCapturing2022}. In rainfall retrieval, it is closely related to the classification--regression methods discussed above, as both distinguish rain occurrence from rain-rate estimation. Unlike classification--regression modeling, however, the hurdle formulation retains the occurrence probability and the positive-rain component in a unified probabilistic model rather than using a hard occurrence threshold to route samples between independent models. The core methodological innovation of Hurdle--RMIL is the newly proposed RMIL for the positive-rain component, for which conventional learning from a naturally long-tailed rainfall distribution systematically underestimates rare high-intensity rainfall. RMIL exploits the invariance, under fixed observation conditions, of the causal relationship from rainfall to the corresponding satellite observation, which is governed by radiative-transfer physics. Based on this invariance, RMIL derives a Bayes-based probability-density transformation that relates the conditional rain-rate densities associated with long-tailed and balanced rainfall distributions. This transformation enables the conditional density associated with a balanced rainfall distribution to be estimated by maximum likelihood directly from naturally collected long-tailed samples, without constructing a balanced training dataset. Under the empirical lognormal formulation adopted in this study, the corresponding negative log-likelihood admits a closed-form expression for AI model training.

Hurdle--RMIL is evaluated using multiregional test data distributed across a broad area of China spanning 106.25°--120.65°E and 23.3°--37.7°N through threshold-based statistical metrics and rainfall-event case studies, with comparisons against conventional, classification--regression, cost-sensitive, and generative learning. The results demonstrate that the proposed framework mitigates systematic underestimation, improves rain-event detection at high rainfall thresholds, and better captures rainfall intensity and spatial extent.

The remainder of this paper is organized as follows. Section~\ref{Hurdle--RMIL} introduces Hurdle--RMIL. Section~\ref{experiments} details the data, experimental design, AI models, and evaluation methods. Section~\ref{results} presents the experimental results, Section~\ref{discussion} discusses the findings, and Section~\ref{conclusion} concludes the study.

\section{Hurdle--RMIL}
\label{Hurdle--RMIL}
\subsection{Hurdle Model}
 Let $R$ denote the rain rate and $S$ denote the infrared brightness-temperature features. Here, $R$ and $S$ are random variables, whereas $r$ and $s$ denote their respective values. Given $S=s$, the hurdle model specifies the conditional distribution of $R$ through its probability mass at zero and its density over the positive domain:
\begin{equation} \label{eq:hurdle_pdf}
P_{R\mid S}(r\mid s)=
\begin{cases}
p(s), & r=0, \\
[1-p(s)]\,\mathrm{f}_{R\mid S}(r\mid s), & r>0.
\end{cases}
\end{equation}
Here, $P_{R\mid S}(\cdot\mid s)$ denotes the conditional probability distribution of $R$ given $S=s$. For $r>0$, $\mathrm{f}_{R\mid S}(r\mid s)$ is the density of the rain rate in the positive-rain component, given $S=s$. The probability mass at zero is denoted by $p(s)$. Thus, the hurdle model represents the conditional distribution of $R$ given $S=s$ as a point mass at zero and a positive-rain conditional distribution. The occurrence component is represented by $p(s)$, whereas the positive-rain-rate component is specified by $\mathrm{f}_{R\mid S}$. While the hurdle model addresses zero inflation by separating zero and positive values, its positive-rain-rate component might still underestimate high-intensity rainfall owing to the long tail if conventional learning is employed.

Additionally, classification--regression modeling is closely related to the hurdle model but differs in a key aspect. In classification--regression modeling, an occurrence model first produces a hard rain/no-rain decision, after which a separate model estimates the rain rate for samples classified as rainy. Structurally, classification--regression modeling and the hurdle model both comprise two stages—occurrence detection and rain rate estimation—thus, they appear similar. The key difference lies in uncertainty handling: classification--regression modeling introduces a hard threshold after occurrence probability estimation, breaking the continuity of uncertainty propagation, whereas the hurdle model performs continuous modeling within a unified probabilistic framework. 

\subsection{RMIL}

The second component of the hurdle model suffers from underestimation of heavy rain owing to the long tail. 
 To clarify this phenomenon, the forward and retrieval processes are first introduced. The forward process follows the causal order of a rainfall event and its satellite observation: rainfall with rate $R=r$ occurs first, after which the satellite observes the corresponding signal $S=s$. It therefore proceeds from $R$ to $S$. The forward model is a probabilistic representation of this process and is specified by the conditional density $\mathrm{f}_{S\mid R}$. Conversely, rainfall retrieval uses the observed signal to infer the preceding rain rate, reversing this causal observation sequence. Its probabilistic representation is the retrieval model, also termed the inversion model, and is specified by the conditional density $\mathrm{f}_{R\mid S}$. In this study, the retrieval model learned from the naturally long-tailed rainfall distribution is termed the ``imbalanced retrieval model.'' Accordingly, the phenomenon can be summarized schematically as follows, where CL denotes conventional learning:
\begin{center}
\textnormal{long-tailed rainfall distribution} $\xrightarrow{\mathrm{CL}}$ \textnormal{imbalanced retrieval model}
\end{center}

According to Bayes’ theorem, the forward and retrieval models relate to the marginal densities $\mathrm{f}_{R}$ and $\mathrm{f}_{S}$ as follows:
\begin{equation} \label{eq:bayes_true_revised}
\mathrm{f}_{R\mid S}(r \mid s) = \frac{\mathrm{f}_{S\mid R}(s \mid r) \cdot \mathrm{f}_{R}(r)}{\mathrm{f}_{S}(s)}
\end{equation}

As shown in Eq.~\eqref{eq:bayes_true_revised}, the retrieval model depends not only on the long-tailed dataset, represented by the marginal densities $\mathrm{f}_{R}$ and $\mathrm{f}_{S}$, but also on the forward model. Hence, the factors involved in learning the imbalanced retrieval model can be summarized as follows:
\begin{center}
\textnormal{long-tailed rainfall distribution} $+$ \textnormal{forward model} $\xrightarrow{\mathrm{CL}}$ \textnormal{imbalanced retrieval model}
\end{center}

Now, an ideal scenario is examined, characterized by a balanced dataset featuring uniform marginal densities for both $R$ and $S$—denoted by $\mathrm{f}_{\mathrm{ideal},R}$ and $\mathrm{f}_{\mathrm{ideal},S}$, respectively, together with an ideal forward model specified by the density $\mathrm{f}_{\mathrm{ideal},S\mid R}$. In this ideal scenario, the ideal retrieval model obtained via conventional learning, specified by the density $\mathrm{f}_{\mathrm{ideal},R\mid S}$, is free from frequency-driven preference.
According to Bayes’ theorem, the following equality holds:
\begin{equation} \label{eq:bayes_ideal_revised}
\mathrm{f}_{\mathrm{ideal},R\mid S}(r \mid s) = \frac{\mathrm{f}_{\mathrm{ideal},S\mid R}(s \mid r) \cdot \mathrm{f}_{\mathrm{ideal},R}(r)}{\mathrm{f}_{\mathrm{ideal},S}(s)}
\end{equation}

In practice, it is impossible to obtain a perfectly balanced dataset owing to theoretical and technical limitations. However,  the causal direction in infrared rainfall retrieval is $R \rightarrow S$: rainfall affects the satellite observation through radiative-transfer physics. Data sampling can change the marginal densities $\mathrm{f}_{R}$ and $\mathrm{f}_{S}$, but it does not alter this radiative-transfer physics. It therefore does not change the causal direction, the forward process, or the forward model under fixed observation conditions. This invariance can be expressed as follows:
\begin{equation} \label{eq:forward_invariance_revised}
\mathrm{f}_{\mathrm{ideal},S\mid R}(s \mid r) = \mathrm{f}_{S\mid R}(s \mid r)
\end{equation}

Based on Eqs.~\eqref{eq:bayes_true_revised}, \eqref{eq:bayes_ideal_revised}, and \eqref{eq:forward_invariance_revised}, Eq.~\eqref{eq:new} can be derived:
\begin{equation}\label{eq:new}
\mathrm{f}_{R\mid S}(r \mid s)
= \mathrm{f}_{\mathrm{ideal},R\mid S}(r \mid s) \cdot
\frac{\mathrm{f}_{R}(r)}{\mathrm{f}_{\mathrm{ideal},R}(r)} \cdot
\frac{\mathrm{f}_{\mathrm{ideal},S}(s)}{\mathrm{f}_{S}(s)}
\end{equation}

Considering $\int_{0}^{+\infty}\mathrm{f}_{R\mid S}(r\mid s)\,\mathrm{d}r=1$ and Eq.~\eqref{eq:new}, Eq.~\eqref{eq:jifen} can be further obtained:
\begin{equation}\label{eq:jifen}
{\int_0^{+\infty} \mathrm{f}_{\mathrm{ideal},R\mid S}(r' \mid s) \cdot
\frac{\mathrm{f}_{R}(r')}{\mathrm{f}_{\mathrm{ideal},R}(r')} \, \mathrm{d}r'} \cdot
\frac{\mathrm{f}_{\mathrm{ideal},S}(s)}{\mathrm{f}_{S}(s)} = 1
\end{equation}

Combining the above derivation steps yields the following:
\begin{equation}
\mathrm{f}_{R\mid S}(r \mid s)
= \frac{\mathrm{f}_{\mathrm{ideal},R\mid S}(r \mid s) \cdot
\frac{\mathrm{f}_{R}(r)}{\mathrm{f}_{\mathrm{ideal},R}(r)}}
{\int_0^{+\infty} \mathrm{f}_{\mathrm{ideal},R\mid S}(r' \mid s) \cdot
\frac{\mathrm{f}_{R}(r')}{\mathrm{f}_{\mathrm{ideal},R}(r')} \, \mathrm{d}r'}
\end{equation}

Because the ideal marginal density $\mathrm{f}_{\mathrm{ideal},R}$ is uniform, it can be omitted in both the numerator and the denominator, resulting in the following expression:
\begin{equation} \label{eq:transform_eq_revised}
\mathrm{f}_{R\mid S}(r \mid s) =
\frac{\mathrm{f}_{\mathrm{ideal},R\mid S}(r \mid s) \cdot \mathrm{f}_{R}(r)}
{\int_0^{+\infty} \mathrm{f}_{\mathrm{ideal},R\mid S}(r' \mid s) \cdot \mathrm{f}_{R}(r') \, \mathrm{d}r'}
\end{equation}

Thus, Eq.~\eqref{eq:transform_eq_revised} establishes a probability-density relationship between the imbalanced and ideal retrieval models. The imbalanced retrieval model tends to underestimate high-intensity rainfall. RMIL uses this relationship to estimate the ideal retrieval model by maximum likelihood directly from naturally collected long-tailed samples, as described below.

Assuming the ideal retrieval model is parameterized by $\theta$, and given a long-tailed dataset containing $n$ independent and identically distributed positive-rain samples $\{(r_i,s_i)\}_{i=1}^{n}$, the likelihood function $L$ is defined as follows:
\begin{equation} \label{eq:likelihood_def_revised}
L = \prod_{i=1}^n \mathrm{f}_{R\mid S}(r_i \mid s_i)
\end{equation}

Substituting Eq.~\eqref{eq:transform_eq_revised} into Eq.~\eqref{eq:likelihood_def_revised} yields the following equation:
\begin{equation} \label{eq:likelihood_transformed_revised}
L(\theta) = \prod_{i=1}^n \frac{\mathrm{f}_{\mathrm{ideal},R\mid S}(r_i \mid s_i; \theta) \cdot \mathrm{f}_{R}(r_i)}{\int_0^{+\infty} \mathrm{f}_{\mathrm{ideal},R\mid S}(r' \mid s_i; \theta) \cdot \mathrm{f}_{R}(r') \, \mathrm{d}r'}
\end{equation}

The values of the parameter $ \theta $ are estimated through the following equation:
\begin{equation} \label{eq:mle_revised}
\hat{\theta} = \operatorname*{arg\,max}_{\theta} \log L(\theta)
\end{equation}

where $ \hat{\theta} $ denotes the parameters of the ideal retrieval model that maximize the likelihood given the long-tailed dataset. This procedure enables learning the ideal retrieval model directly from a naturally collected long-tailed dataset. This learning method is referred to as RMIL.

\subsection{Empirical Distribution and Optimization Objective}
\label{EDOO}

The hurdle model and RMIL jointly constitute the complete Hurdle--RMIL framework. The hurdle formulation provides the probabilistic vehicle for applying RMIL to the positive-rain component while retaining the occurrence component within the same conditional distribution. To apply Hurdle--RMIL to rainfall retrieval, this section introduces the empirical distribution and derives the corresponding optimization objective. 

The lognormal distribution is not only employed widely to model the distribution of $R$, but has also been used successfully to model the conditional distribution of $R$ given satellite observations \cite{kirstetterProbabilisticPrecipitationRate2018a}. According to Eqs.~\eqref{eq:hurdle_pdf} and \eqref{eq:transform_eq_revised}, the complete conditional distribution of $R$ given $S=s$ can be expressed as follows:
\begin{equation}
    \label{eq:full_model_pdf_final}
    P_{R\mid S}(r\mid s)=
    \begin{cases}
        p(s), & r=0, \\
        [1-p(s)]\dfrac{\mathrm{Num}(r,s)}{\mathrm{Den}(s)}, & r>0.
    \end{cases}
\end{equation}
where the numerator and denominator are defined as:
{\footnotesize
\begin{gather}
    \label{eq:N_def}
    \mathrm{Num}(r,s) = \mathrm{f}_{\mathrm{ideal},R\mid S}(r \mid s; \mu, \sigma) \cdot \mathrm{f}_{R}(r; \mu_R, \sigma_R) \\ 
    \label{eq:D_def}
    \mathrm{Den}(s) = \int_0^{+\infty} \mathrm{f}_{\mathrm{ideal},R\mid S}(r' \mid s; \mu, \sigma) \cdot \mathrm{f}_{R}(r'; \mu_R, \sigma_R) \, \mathrm{d}r' 
\end{gather}
}
where $\mu$ and $\sigma$ denote the parameters of the ideal retrieval model, and $\mu_R$ and $\sigma_R$ denote the parameters of the marginal density $\mathrm{f}_{R}$, which can be estimated statistically from the training dataset.

The negative log-likelihood $\mathrm{NLL}(p, \mu, \sigma)$ of this mixed conditional distribution—that is, the objective function of the AI model—admits a closed-form expression and can be decomposed into a sum of terms:
\begin{align}
    \label{eq:nll_full_combined}
    \begin{split}
    \mathrm{NLL}(p, \mu, \sigma) &= \sum_{i=1}^N \Big( \text{DryT}_i + \text{WetT}_i \\
    & \quad + \text{LogNormT}_i + \text{CorrT}_i \Big)
    \end{split} \\
    \text{DryT}_i &= -\mathbb{I}[r_i = 0] \log(p_i) \notag \\
    \text{WetT}_i &= -\mathbb{I}[r_i > 0] \log(1 - p_i) \notag \\
    \text{LogNormT}_i &= \mathbb{I}[r_i > 0] \left( \frac{(\log(r_i) - \mu_{i})^2}{2\sigma_{i}^2} + \log(\sigma_{i}) \right) \notag \\
    \begin{split} 
    \text{CorrT}_i &= \mathbb{I}[r_i > 0] \cdot \left[ - \frac{1}{2 (\sigma_R^2 + \sigma_{i}^2)} \right. \\ 
    & \quad \cdot \Big( \mu_R^2 + \mu_{i}^2 + \sigma_R^2 (2\mu_{i} - \sigma_{i}^2) \\ 
    & \quad \left. + 2\mu_R (\sigma_{i}^2 - \mu_{i}) \Big) - \frac{1}{2} \log(\sigma_R^2 + \sigma_{i}^2) \right] \notag 
    \end{split}
\end{align}
where $N$ denotes the total number of training samples; $r_i$ denotes the observed rain rate of the $i$-th sample; $ \mathbb{I}[\cdot] $ denotes the indicator function; $ p_i$, $\mu_i$, and $\sigma_i $ correspond to the satellite observation $s_i$; and CorrT represents a correction term that embodies RMIL’s adjustment and redirection of conventional learning.

For a satellite observation $s$, the predicted rain rate is taken as the conditional expectation of $R$ given $S=s$:
\begin{equation} \label{eq:qpe_expectation_final}
\mathbb{E}[R\mid S=s]
= [1-p(s)]\exp\left(\mu(s)+\frac{\sigma^2}{2}\right)
\end{equation}

Additionally, as the conditional distribution is fitted, it also enables quantification of the uncertainty. Nevertheless, this lies beyond the scope of the present study and is not elaborated on further.

\subsection{Parameter Estimation Scheme}

According to subsection~\ref{EDOO}, three parameters must be estimated: $p$, $\mu$, and $\sigma$. Based on observed phenomena, a hybrid estimation scheme has been designed.

It is observed that without range constraints on the AI model’s output, the estimated value of $\sigma$ tends to diverge to infinity, leading to model failure. Conversely, when range constraints are imposed, all estimates converge uniformly to the set upper bound. This phenomenon suggests that the current RMIL is insufficient to guide differentiated dynamic estimation of $\sigma$. To address this, $\sigma$ is treated as a hyperparameter and assigned a fixed value that is selected from a set of candidates through external optimization methods (e.g., grid search). Although this hyperparameterization is a compromise, subsequent rainfall retrieval results have demonstrated that it is an effective strategy.

For $p$ and $\mu$, dynamic estimation is performed. Two implementation options exist: one employs two independent AI models to estimate $p$ and $\mu$ sequentially, while the other uses a single multioutput AI model for joint estimation. Experimental results with a deep neural network show that the single-model approach is more effective overall.

\section{Experiments}
\label{experiments}
\subsection{Data}

\begin{figure*}[!t]
    \centering
    \includegraphics[width=1\linewidth]{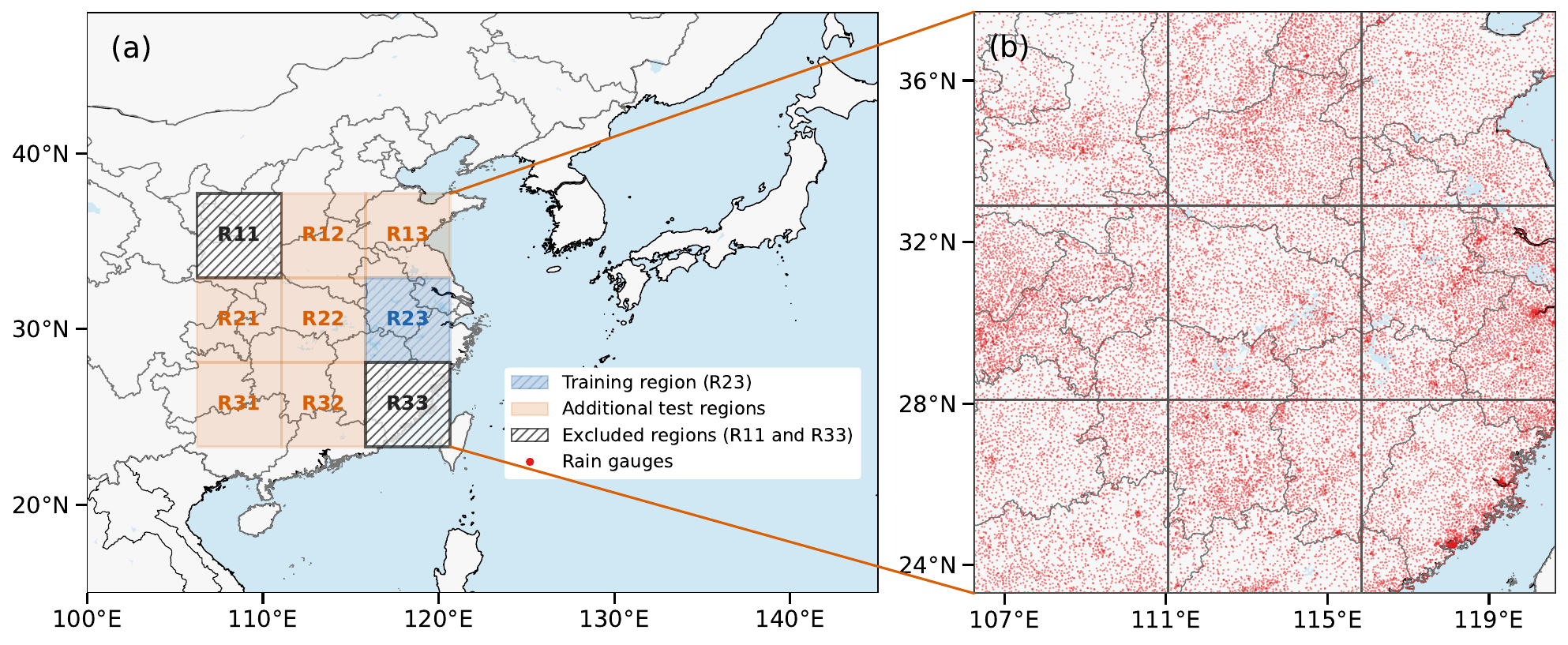}
    \caption{Study domain and experimental partition. Each region contains $96\times96$ grid points. (a) Locations of the R11–R33 regions; R23 is used for model training, validation, and testing; R12, R13, R21, R22, R31, and R32 are used for testing; and R11 and R33 are excluded. (b) Rain-gauge distribution across the study domain.}
    \label{fig:sampling area} 
\end{figure*}

As shown in Fig.~\ref{fig:sampling area}, the study domain spans 23.3°–37.7°N and 106.25°–120.65°E. It is partitioned into nine contiguous $96\times96$ grid regions, denoted R11–R33 from north to south and west to east. R11 is excluded because its relatively sparse gauge network does not support a sufficiently reliable interpolated reference product, while R33 is excluded because ocean areas without rain-gauge observations account for a large proportion of the region. The geographic coverage, sample-collection period, and rain-gauge count of each region are summarized in Table~\ref{tab:regional_gauges}. The regions retained for model training, validation, and testing have dense rain-gauge networks, supporting reliable gridded rainfall products through interpolation. Radiance data from Japan's Himawari-8 satellite are also used. The Advanced Himawari Imager (AHI) onboard Himawari-8 operates across 16 channels (band01–band16) covering visible, near-infrared, and infrared wavelengths. Details of the infrared channels are listed in Table~\ref{tab:table1}. Under standard mode, AHI provides full-disk observations every 10 min.

\begin{table*}[!t]
    \renewcommand{\arraystretch}{1.2}
    \centering
    \caption{Geographic coverage, sample-collection period, and rain-gauge count of each region.}
    \label{tab:regional_gauges}
    \footnotesize
    \begin{tabular}{@{}c c c c c@{}}
        \toprule
        Region & Longitude ($^\circ$E) & Latitude ($^\circ$N) & Gauges & Period \\
        \midrule
        R11 & 106.25--111.05 & 32.90--37.70 & 2323 & -- \\
        R12 & 111.05--115.85 & 32.90--37.70 & 3969 & May--August, 2019--2021 \\
        R13 & 115.85--120.65 & 32.90--37.70 & 2605 & May--August, 2019--2021 \\
        R21 & 106.25--111.05 & 28.10--32.90 & 4144 & May--August, 2019--2021 \\
        R22 & 111.05--115.85 & 28.10--32.90 & 3447 & May--August, 2019--2021 \\
        R23 & 115.85--120.65 & 28.10--32.90 & 5041 & May--August, 2016--2021 \\
        R31 & 106.25--111.05 & 23.30--28.10 & 2905 & May--August, 2019--2021 \\
        R32 & 111.05--115.85 & 23.30--28.10 & 3786 & May--August, 2019--2021 \\
        R33 & 115.85--120.65 & 23.30--28.10 & 3193 & -- \\
        \bottomrule
    \end{tabular}
\end{table*}


\begin{table}[!t] 
\renewcommand{\arraystretch}{1.3} 
\caption{Information regarding the Himawari-8/AHI infrared channels.} 
\label{tab:table1}
\centering 
\footnotesize 
\begin{tabularx}{\columnwidth}{@{}l c >{\raggedright\arraybackslash}X@{}}
    \toprule
    Channel & \makecell[c]{Central Wavelength\\ (\si{\micro\meter})} & Principal application \\
    \midrule
    Band08 & 6.2 & Water vapor (WV) in upper troposphere \\ 
    Band09 & 6.9 & WV in upper and middle troposphere \\ 
    Band10 & 7.3 & WV in middle and lower troposphere \\ 
    Band11 & 8.6 & Cloud phase \\ 
    Band12 & 9.6 & Total ozone \\ 
    \midrule
    Band13 & 10.4 & \multirow{4}{*}{Cloud top temperature} \\ 
    Band14 & 11.2 & \\ 
    Band15 & 12.4 & \\ 
    Band16 & 13.3 & \\ 
    \bottomrule
\end{tabularx}
\end{table}

Following the setup of \cite{hiroseHighTemporalRainfall2019}, one infrared channel and five groups of brightness-temperature differences are used as input features (Table~\ref{tab:table2}). Using nearest-neighbor interpolation, the satellite observations and hourly accumulated rain-gauge measurements are mapped onto a 0.05°$\times$0.05° grid, yielding a $96\times96$ grid for each region. In R23, 4883 spatiotemporally matched samples collected from May to August during 2016–2021 are obtained and divided into 3907 training samples, 488 validation samples, and 488 test samples. Samples collected from May to August during 2019–2021 from R12, R13, R21, R22, R31, and R32 are also used for testing. Consequently, the training, validation, and test sets contain 3907, 488, and 2177 samples, respectively. The trained model is applied to all test regions without regional retraining or fine-tuning. Across the 3907 R23 training samples, grid points with $R=0$ account for approximately 78.3\% of all training grid points, whereas those with $R>0$ account for approximately 21.7\%. This substantial probability mass at zero indicates the presence of zero inflation in the training data. Among the positive-rain grid points, $\mu_R\approx0.46$ and $\sigma_R\approx1.28$, and the sample frequency decreases rapidly as rain rate increases, indicating a long-tailed positive-rain distribution.


\begin{table}[!t] 
\renewcommand{\arraystretch}{1.3} 
\caption{Signal (model input) utilized for AI-based rain retrieval algorithms.} 
\label{tab:table2}
\centering 
\footnotesize 
\begin{tabular}{cc} 
    \toprule
    Signal & Indicative information \\
    \midrule
    Band13 & \multirow{2}{*}{Cloud top height} \\
    Band10--Band16 & \\
    \midrule
    Band11--Band13 & \multirow{2}{*}{Cloud water path} \\
    Band13--Band15 & \\
    \midrule
    Band08--Band09 & \multirow{2}{*}{Water vapor} \\
    Band09--Band10 & \\
    \bottomrule
\end{tabular}
\end{table}

\subsection{Baselines}

Five baselines (see Table~\ref{tab:table3}) were selected to evaluate the performance of Hurdle--RMIL relative to the baseline methods. 
Multi-Task Collaboration Deep Learning Framework (MTCF), developed by Yang et al.~\cite{yangMultiTaskCollaborationDeep2021}, is included as an enhanced classification--regression method implemented through multi-task learning. It retains rain-occurrence classification and rain-rate regression as distinct tasks while jointly optimizing them within a unified framework.
The nonlinear weighted MSE (NWMSE) and linear weighted MSE (LWMSE) are included as representatives of cost-sensitive learning methods. Specifically, LWMSE assigns weights to the objective function via a linear function. Although not previously used for rainfall retrieval, LWMSE is an established method for handling imbalanced label distribution and has shown superior performance in a depth-estimation task \cite{jiaoLookDeeperDepth2018}. NWMSE was introduced by Berthomier et al.~\cite{berthomierEspressoGlobalDeep2023} specifically for rainfall retrieval, demonstrating enhanced capability in detecting heavy rain compared to operational products like IMERG. As a representative generative learning method, the Diffusion model was configured based on \cite{xiaoRetrievalTotalPrecipitable2025}, despite its original application being cloud property and atmospheric precipitable water estimation. Because inference with the Diffusion model is stochastic, the ensemble mean of the generated members is used as its deterministic rainfall estimate for evaluation. Finally, the original MSE (OMSE) is trained by minimizing the standard, unmodified MSE, the most commonly used objective function in conventional regression. Unlike LWMSE and NWMSE, OMSE applies no rain-rate-dependent weighting and is optimized directly over the naturally imbalanced training samples without resampling or distribution correction. It therefore represents conventional learning and provides a reference for evaluating the benefits of imbalance-aware methods.

    

\begin{table}[!t] 
\centering
\begin{threeparttable}
\renewcommand{\arraystretch}{1.3} 

\caption{Information regarding the baselines.}
\label{tab:table3}

\begin{tabularx}{\columnwidth}{@{}l c >{\raggedright\arraybackslash}X@{}}
    \toprule
    Method & Source & Category \\
    \midrule
    OMSE & & Conventional learning \\
    MTCF & \cite{yangMultiTaskCollaborationDeep2021} & Classification--regression \\
    NWMSE & \cite{berthomierEspressoGlobalDeep2023} & Cost-sensitive learning \\
    LWMSE & \cite{jiaoLookDeeperDepth2018} & Cost-sensitive learning \\
    Diffusion & \cite{Xiao2025-it, xiaoRetrievalTotalPrecipitable2025} & Generative learning \\
    \bottomrule
\end{tabularx}

\vspace{0.5em} 

\begin{tablenotes}
    \footnotesize
    \item[Note:] OMSE: Original MSE; NWMSE: Nonlinear weighted MSE; LWMSE: Linear weighted MSE; Diffusion: Diffusion model; MTCF: Multi-Task Collaboration deep learning Framework.
\end{tablenotes}
\end{threeparttable}
\end{table}

\subsection{AI Model}

\begin{figure*}[!t] 
    \centering
    \includegraphics[width=1\linewidth]{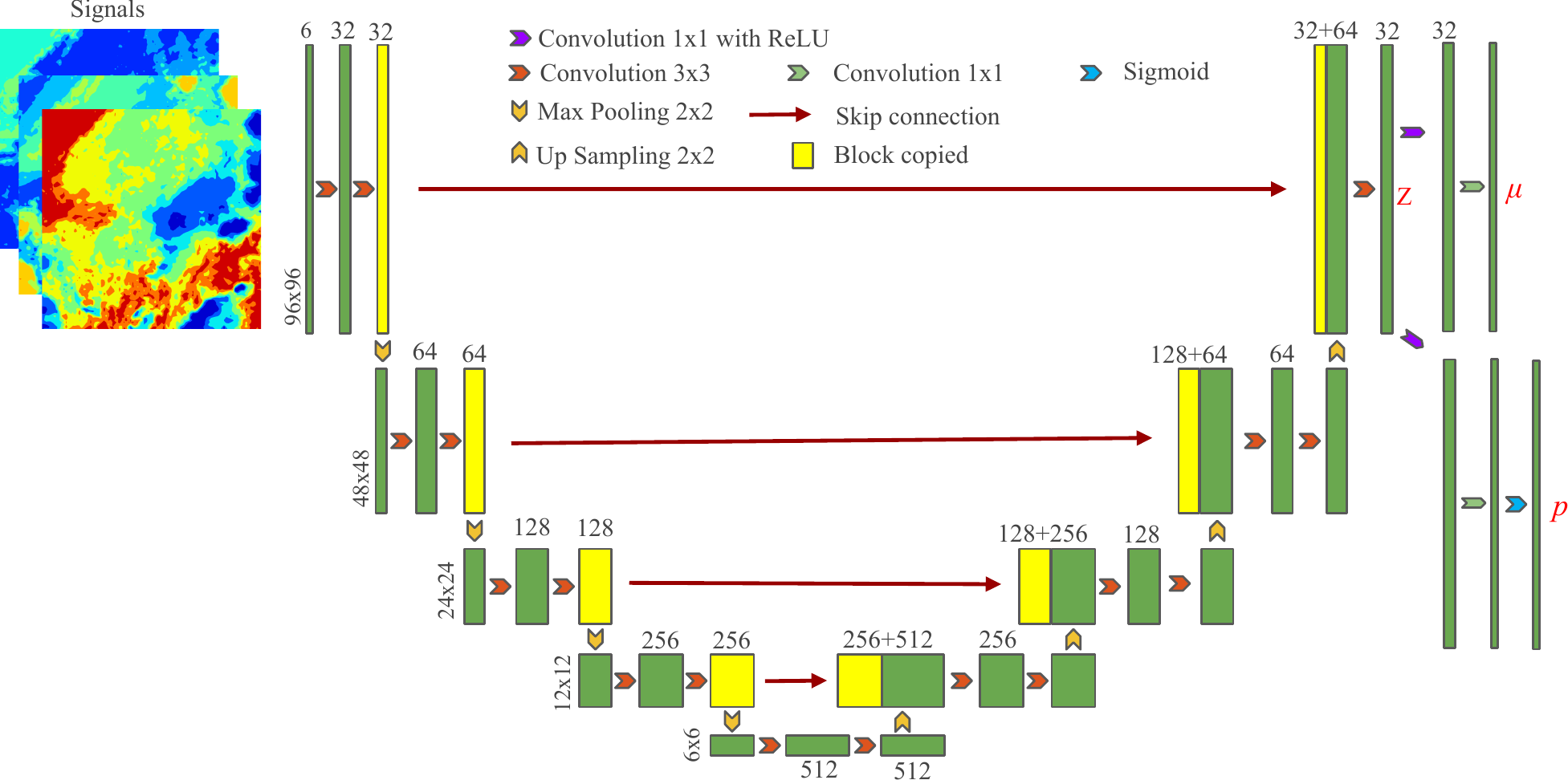} 
    \caption{U-Net architecture for jointly estimating $p$ and $\mu$. ReLU denotes the activation function; the green and yellow rectangles represent feature maps.}
    \label{model_arch}
\end{figure*}

A modified U-Net was employed to jointly estimate  $p$ and $\mu$. Originally developed for medical image segmentation \cite{ronnebergerUNetConvolutionalNetworks2015b}, U-Net has since been applied widely in the fields of computer vision, weather forecasting, and remote sensing. Inspired by the Mixture Density Network \cite{860813}, we extended U-Net with additional output modules to estimate both $p$ and $\mu$. The architecture of the model is illustrated in Fig.~\ref{model_arch}. The model extracts deep features $Z$, which are then processed by two branches: $\mu$ is predicted via stacked $1 \times 1$ convolutions with a Rectified Linear Unit activation function after the first layer, and $p$ is predicted through a parallel sequence ending with sigmoid activation. OMSE, LWMSE, NWMSE, and Diffusion employ the original U-Net because no estimation of multiple parameters is required, whereas MTCF uses the modified U-Net of \cite{yangMultiTaskCollaborationDeep2021}.

For all models, standard training strategies—including batch normalization, Kaiming initialization, early stopping, the Adam optimizer, weight decay, and data normalization—were applied consistently to promote efficient convergence and reliable generalization. The objective function of MTCF consists of three components, with their relative weights tuned to balance their magnitudes; in this study, they were set to 3, 1, and 3. The threshold hyperparameter exerts decisive influence on MTCF’s performance. Based on a systematic evaluation of multiple decision thresholds, 0.5 was selected as a representative configuration for the baseline; detailed results are provided in Appendix~\ref{app1}.

\subsection{Evaluation}

To quantitatively evaluate rainfall-retrieval performance, two continuous error metrics and three categorical verification metrics are adopted: root mean square error (RMSE), mean error (ME), probability of detection (POD), false alarm ratio (FAR), and equitable threat score (ETS). These metrics are defined as follows. Here, $y_i$ and $\hat{y}_i$ denote the observed and retrieved rain rates, respectively, and $n$ denotes the number of valid observation--retrieval pairs meeting the corresponding rainfall threshold. For a given threshold, true positives (TP) are observed events with retrieved rain rates greater than or equal to the threshold, whereas false negatives (FN) are observed events with retrieved rain rates below the threshold. False positives (FP) are observed nonevents with retrieved rain rates greater than or equal to the threshold, whereas true negatives (TN) are observed nonevents with retrieved rain rates below the threshold. The term $H$ denotes the number of hits expected by chance.

\begin{equation} \label{RMSE}
    \mathrm{RMSE} = \sqrt{\frac{1}{n} \sum_{i=1}^{n} (y_i - \hat{y}_i)^2}
\end{equation}

\begin{equation} \label{ME}
    \mathrm{ME} = \frac{1}{n} \sum_{i=1}^{n} (\hat{y}_i - y_i)
\end{equation}

\begin{equation} \label{POD}
    \mathrm{POD} = \frac{\mathrm{TP}}{\mathrm{TP} + \mathrm{FN}}
\end{equation}

\begin{equation} \label{FAR}
    \mathrm{FAR} = \frac{\mathrm{FP}}{\mathrm{TP} + \mathrm{FP}}
\end{equation}

\begin{equation} \label{ETS}
    \mathrm{ETS} = \frac{\mathrm{TP} - H}{\mathrm{TP} + \mathrm{FP} + \mathrm{FN} - H}
\end{equation}

\begin{equation} \label{eq:H_def}
    H = \frac{(\mathrm{TP} + \mathrm{FP})(\mathrm{TP} + \mathrm{FN})}{\mathrm{TP} + \mathrm{FP} + \mathrm{FN} + \mathrm{TN}}
\end{equation}

RMSE measures the magnitude of retrieval errors and ranges from 0 to $+\infty$, with smaller values indicating higher accuracy. ME measures systematic error: negative and positive values indicate underestimation and overestimation, respectively, and a larger absolute value indicates a greater bias magnitude. POD is the fraction of observed events that are correctly detected and ranges from 0 to 1, with higher values indicating fewer misses. FAR is the fraction of retrieved events that are not observed and also ranges from 0 to 1, with lower values indicating fewer false detections. ETS measures detection skill after accounting for hits expected by chance and ranges from $-1/3$ to 1, with higher values indicating better performance; an ETS of zero or less indicates no detection skill beyond chance.

A hierarchical evaluation strategy is adopted. RMSE and ME are evaluated at a threshold of 0 and at 11 positive thresholds (i.e., 0.1, 0.5, 1, 2, 3, 5, 7, 10, 15, 20, and 30 $\mathrm{mm}\cdot\mathrm{h}^{-1}$). The threshold of 0 represents the complete set of valid samples. At each positive threshold, RMSE and ME are computed using observed samples with rain rates greater than or equal to that threshold, while a rainfall event for the categorical metrics is defined as an observed or retrieved rain rate greater than or equal to that threshold. The numbers of observed grid-point samples meeting the 10, 15, 20, and 30 $\mathrm{mm}\cdot\mathrm{h}^{-1}$ thresholds are reported in Table~\ref{tab:threshold_counts}. Sampling uncertainty in RMSE, ME, POD, FAR, and ETS at the 15, 20, and 30 $\mathrm{mm}\cdot\mathrm{h}^{-1}$ thresholds is quantified using a paired nonparametric bootstrap. Each spatiotemporally matched regional sample is treated as a resampling block, with all grid points retained together to preserve the spatial dependence within the sample. In each of \num{5000} bootstrap replicates, regional samples are drawn with replacement, and the same draw is applied to all models to preserve the paired structure of model comparisons. The metrics are then recalculated at these three thresholds, and the 2.5th and 97.5th percentiles of their bootstrap distributions are taken as the bounds of the 95\% confidence intervals. These intervals are used to assess whether the performance differences at high rainfall thresholds remain consistent under variations in the composition of the test samples.
Additionally, two representative rainfall events are analyzed as case studies.

\begin{table}[!t]
    \renewcommand{\arraystretch}{1.2}
    \centering
    \caption{Observed grid-point sample counts at selected high rainfall thresholds.}
    \label{tab:threshold_counts}
    \footnotesize
    \begin{tabular}{cr}
        \toprule
        Threshold ($\mathrm{mm}\cdot\mathrm{h}^{-1}$) & Samples \\
        \midrule
        10 & \num{497908} \\
        15 & \num{279246} \\
        20 & \num{162986} \\
        30 & \num{55562}  \\
        \bottomrule
    \end{tabular}
\end{table}

\section{Results}
\label{results}
This section evaluates Hurdle--RMIL from the perspectives of model configuration, component effectiveness, and retrieval performance. Sensitivity analysis is first conducted to characterize how the hyperparameter $\sigma$ affects the performance metrics across rainfall thresholds, after which an ablation experiment isolates the contribution of RMIL. The joint and sequential parameter-estimation strategies are then compared to determine which strategy provides better overall performance. Finally, Hurdle--RMIL is compared with the baseline methods through case studies that assess the spatial representation of individual rainfall events and through quantitative evaluation over the entire test set that examines overall retrieval performance.

\subsection{Selection of \texorpdfstring{$\sigma$}{sigma}} \label{section:sigma_ana}

\begin{figure*}[!t]
    \centering
        \includegraphics[width=0.46\textwidth]{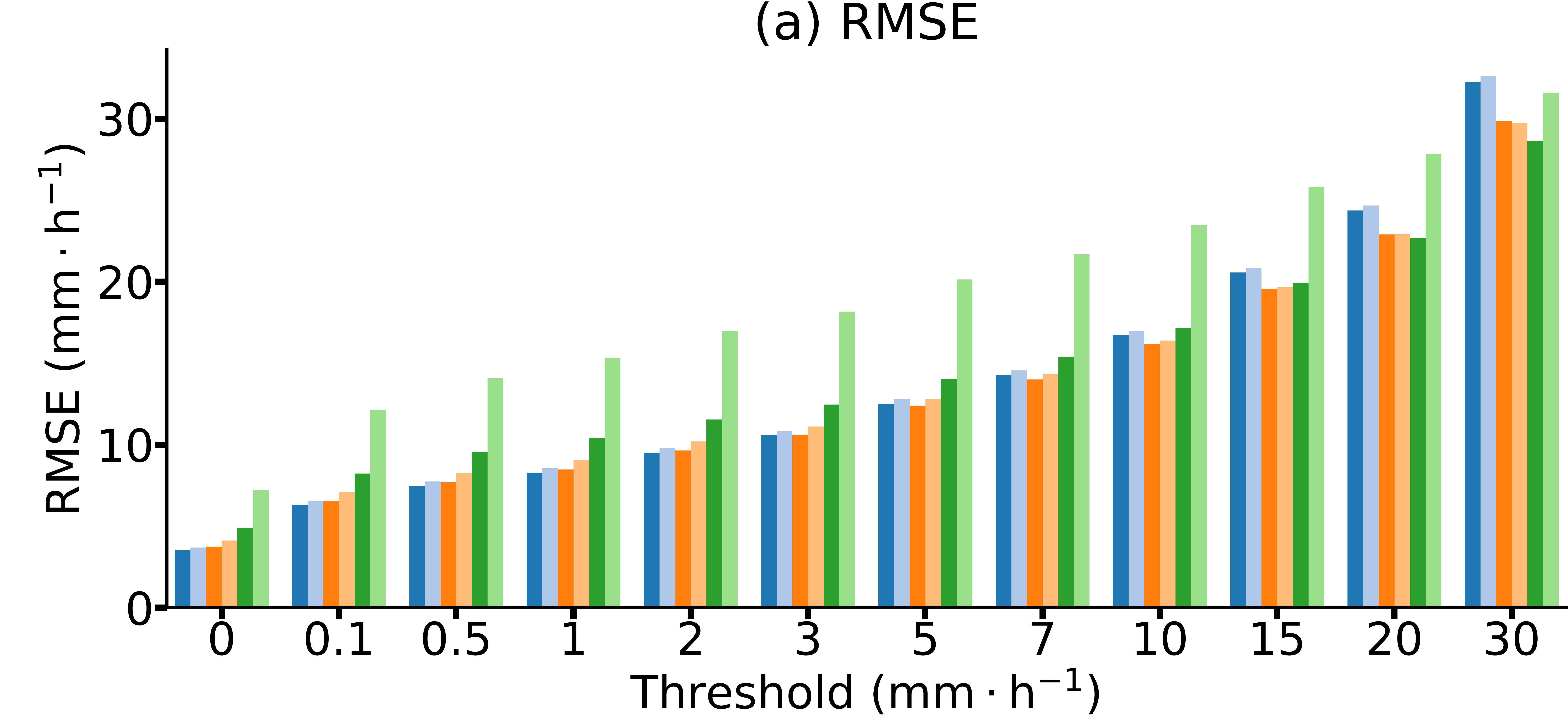}
        \includegraphics[width=0.46\textwidth]{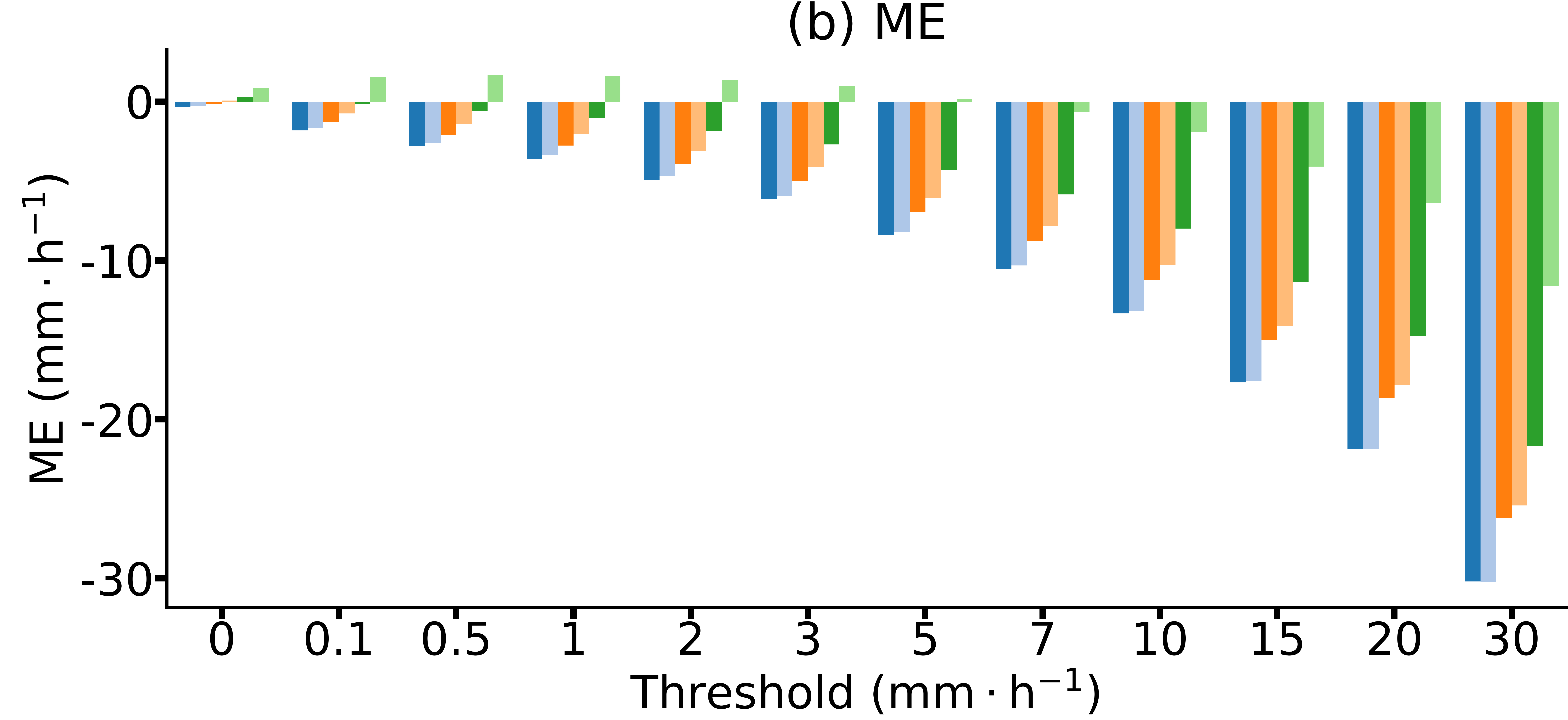}
        \par\vspace{1.5ex}
        \includegraphics[width=0.46\textwidth]{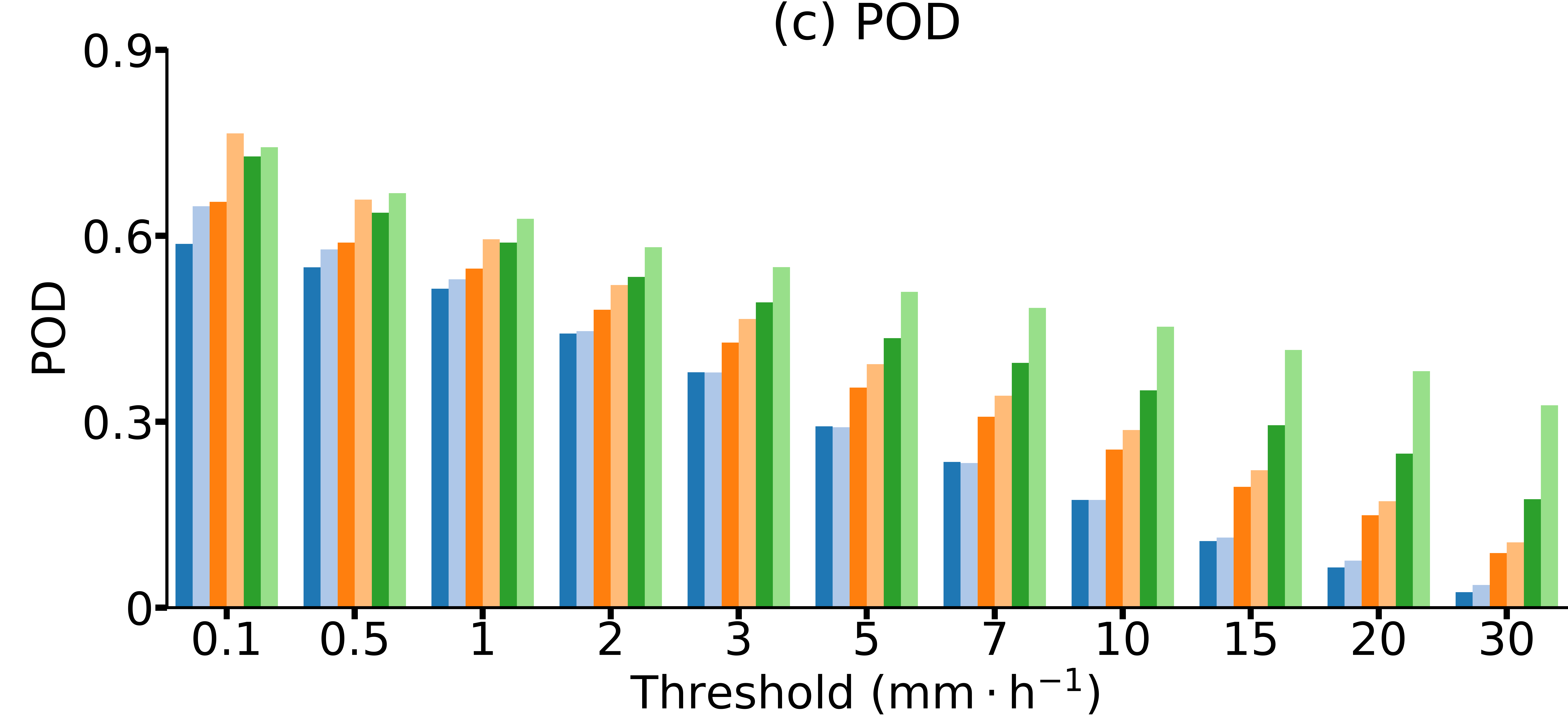}
        \includegraphics[width=0.46\textwidth]{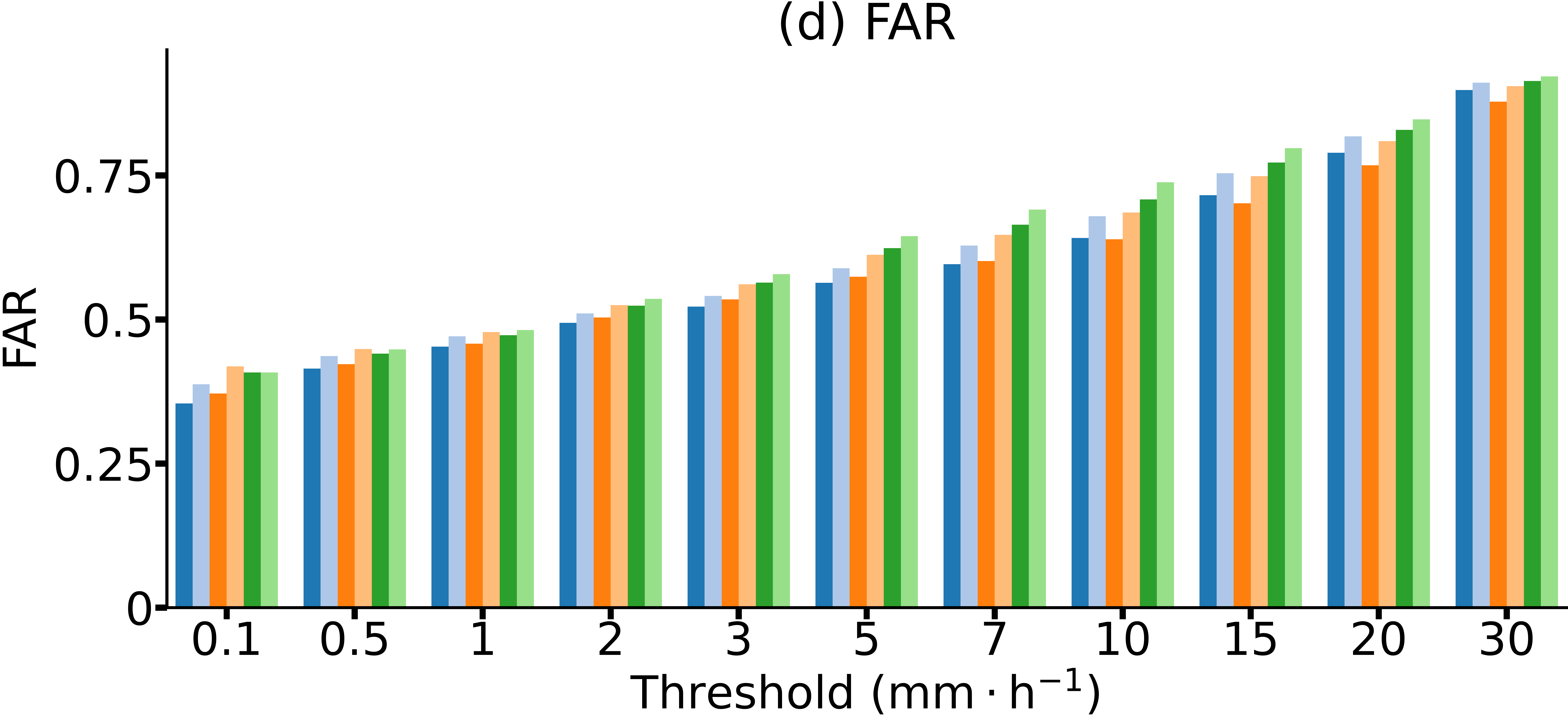}
        \par\vspace{1.5ex}
        \includegraphics[width=0.46\textwidth]{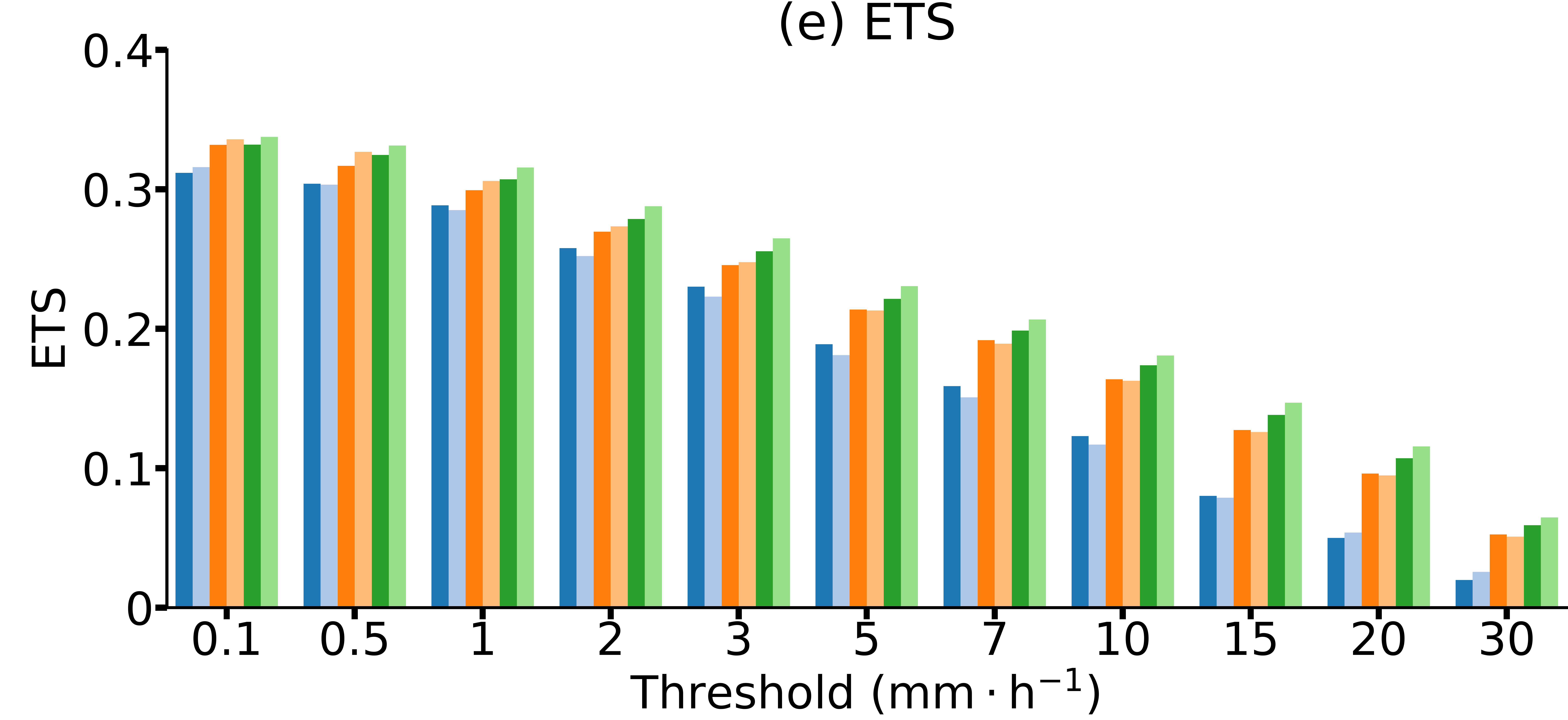}
        \includegraphics[width=0.46\textwidth]{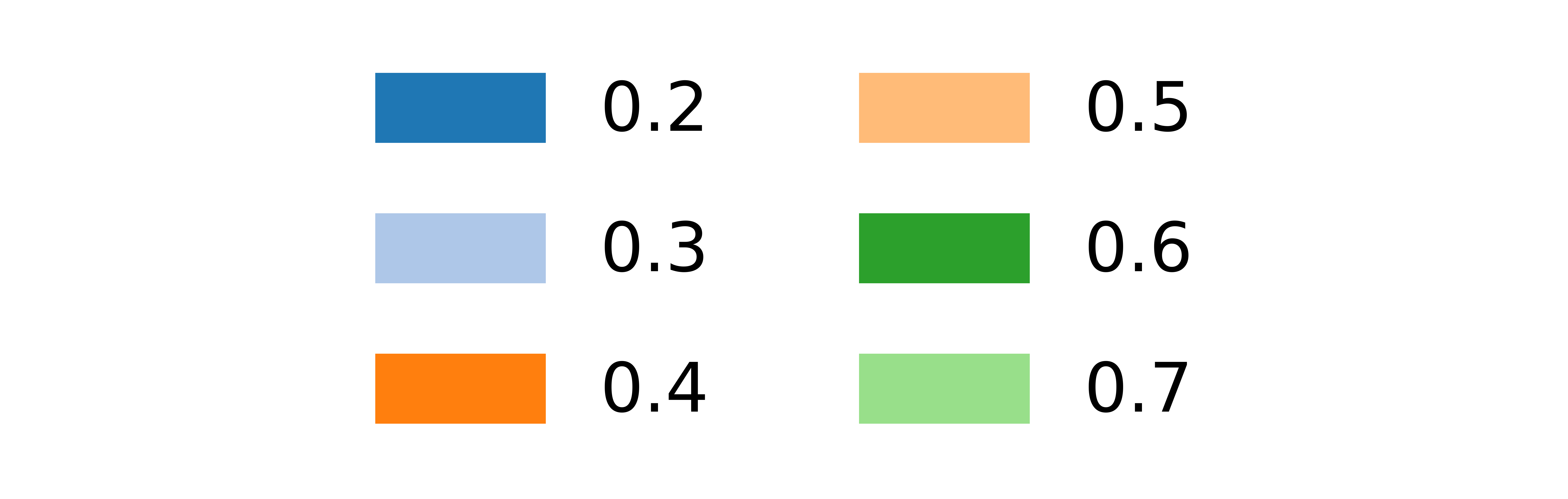}
    \caption{Performance of Hurdle--RMIL with $\sigma$ ranging from 0.2 to 0.7 over the pooled test set: (a) RMSE, (b) ME, (c) POD, (d) FAR, and (e) ETS.}
    \label{fig:sigma_ana}
\end{figure*}

\begin{figure*}[!t]
    \centering
    \includegraphics[width=0.46\textwidth]{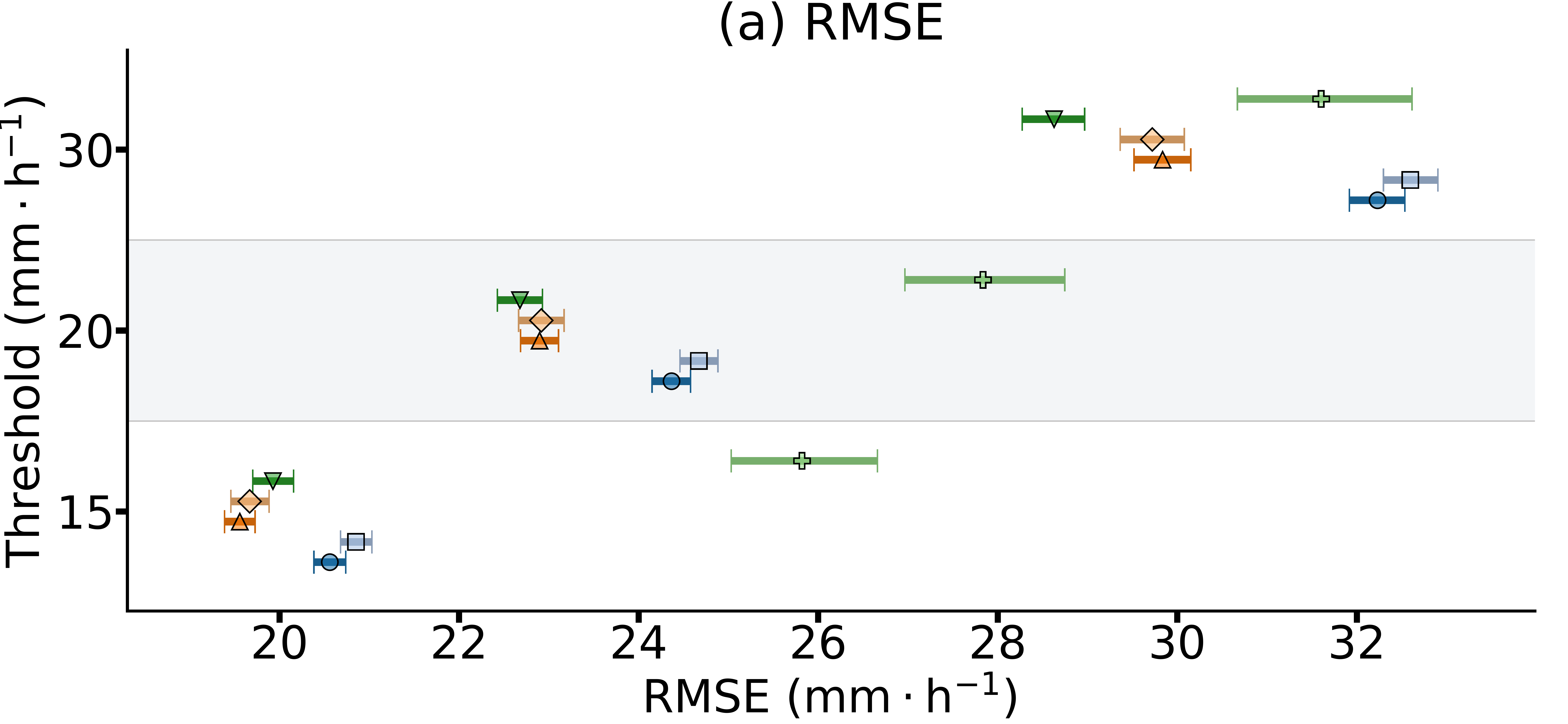}
    \includegraphics[width=0.46\textwidth]{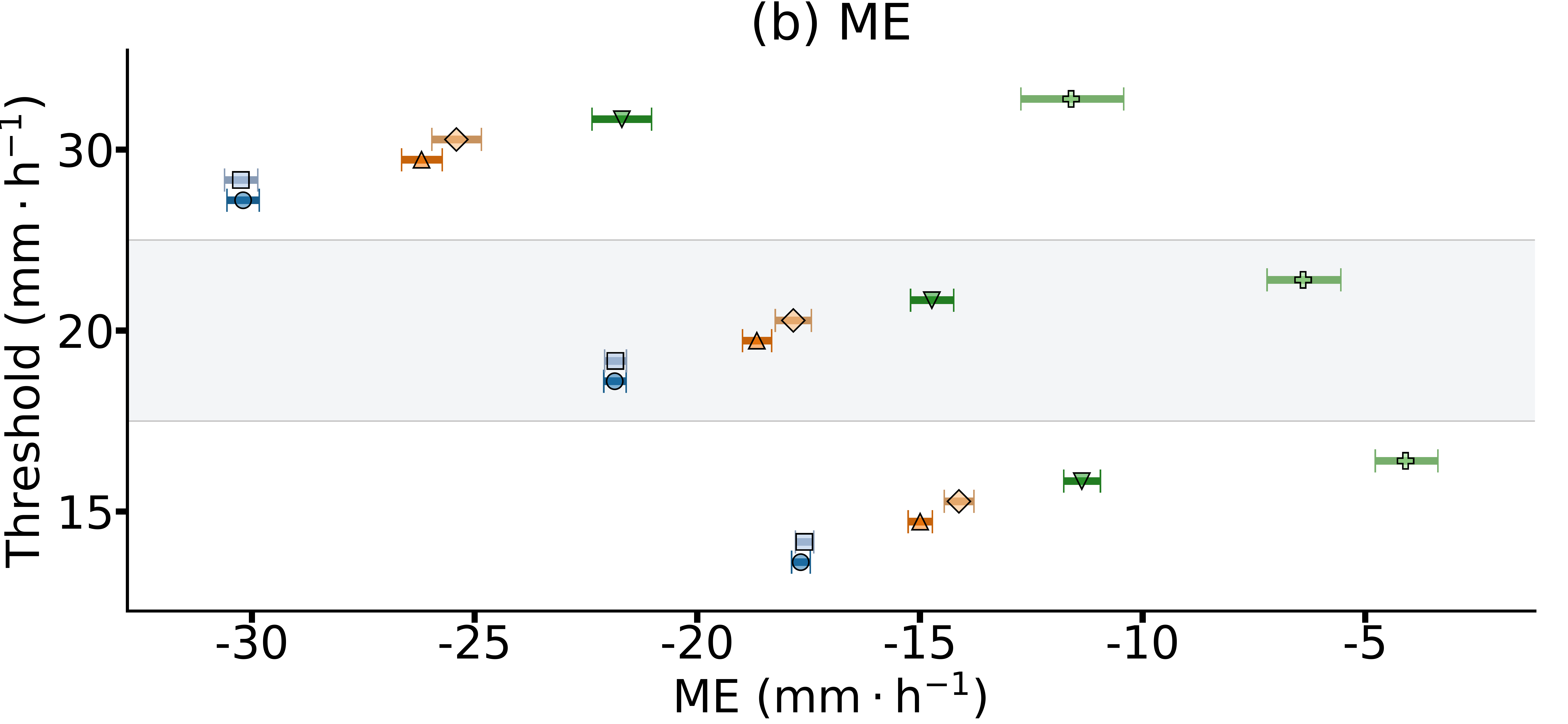}
    \par\vspace{1.5ex}
    \includegraphics[width=0.46\textwidth]{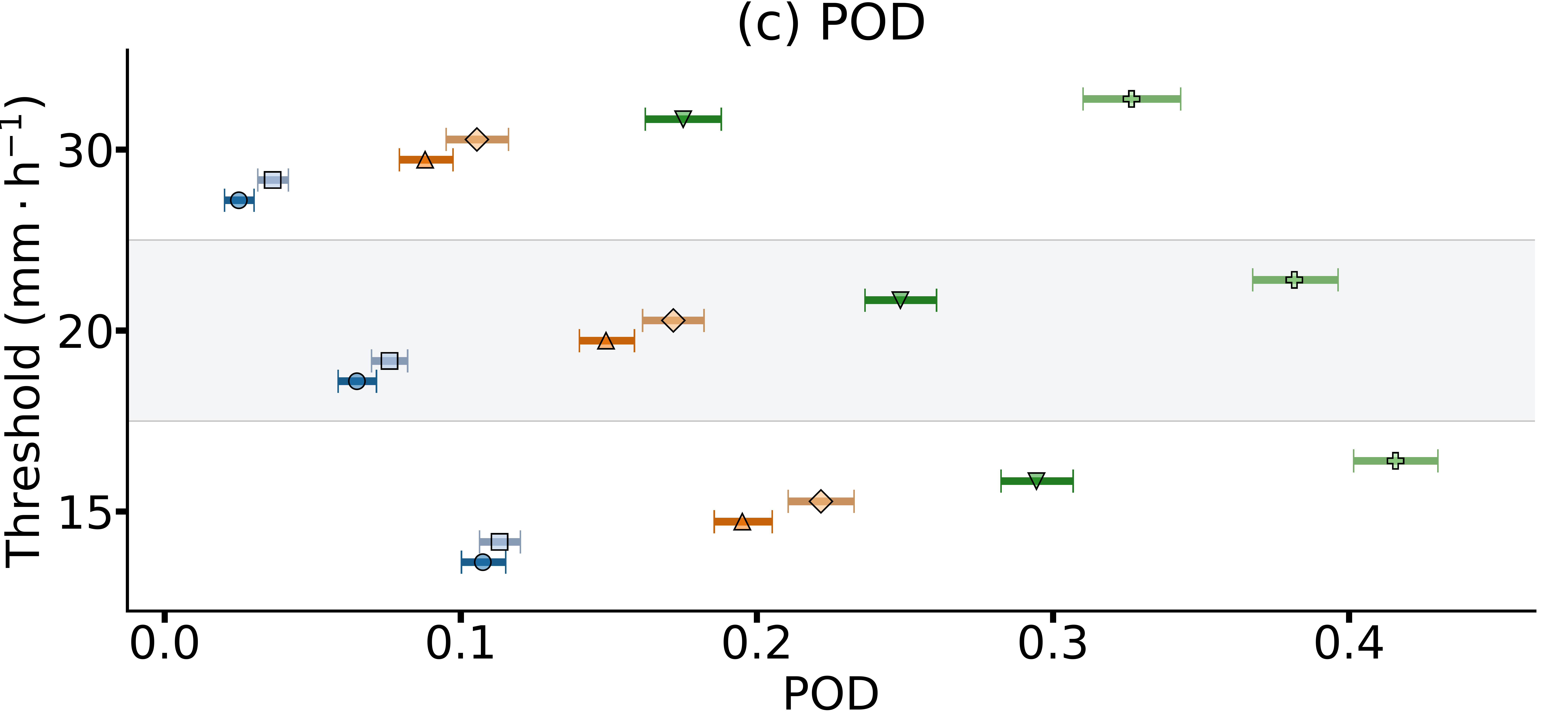}
    \includegraphics[width=0.46\textwidth]{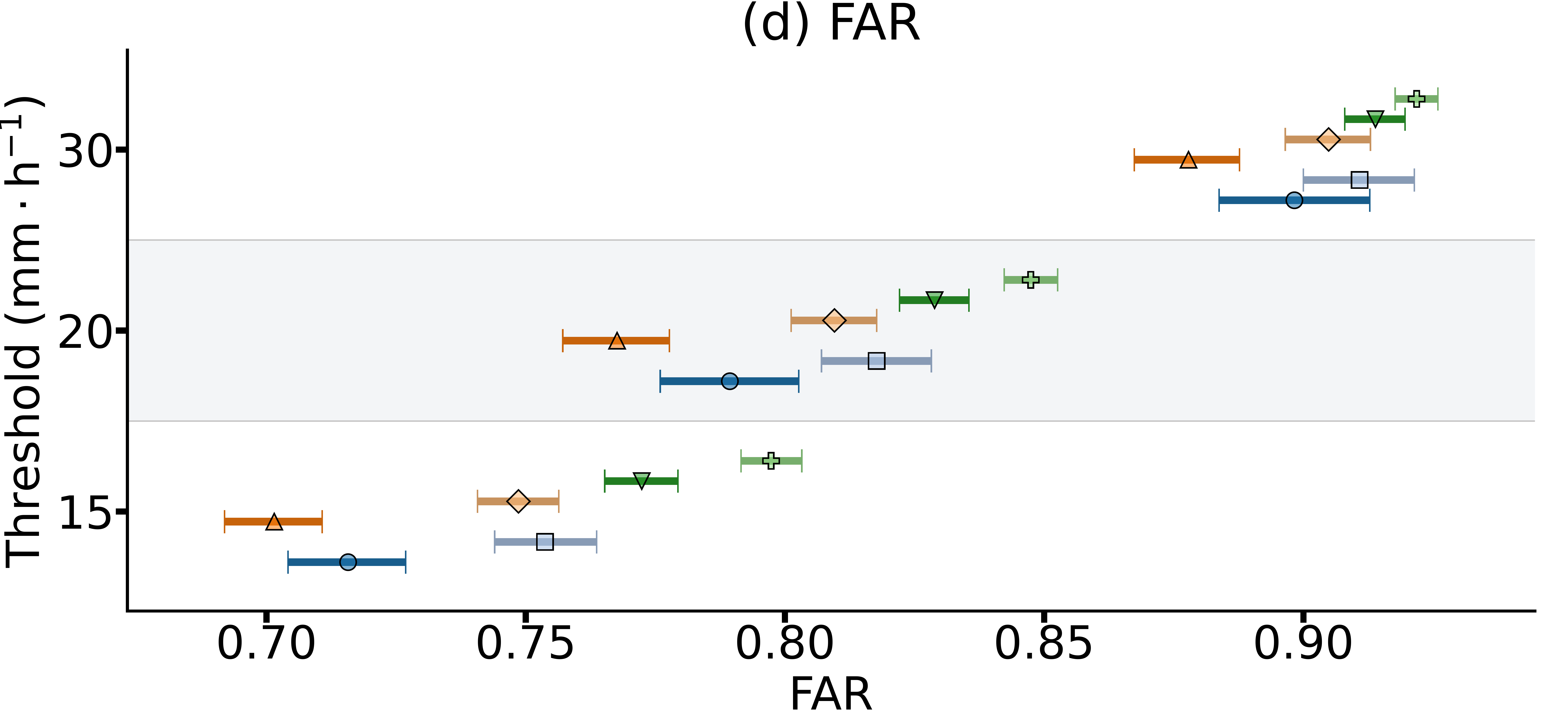}
    \par\vspace{1.5ex}
    \includegraphics[width=0.46\textwidth]{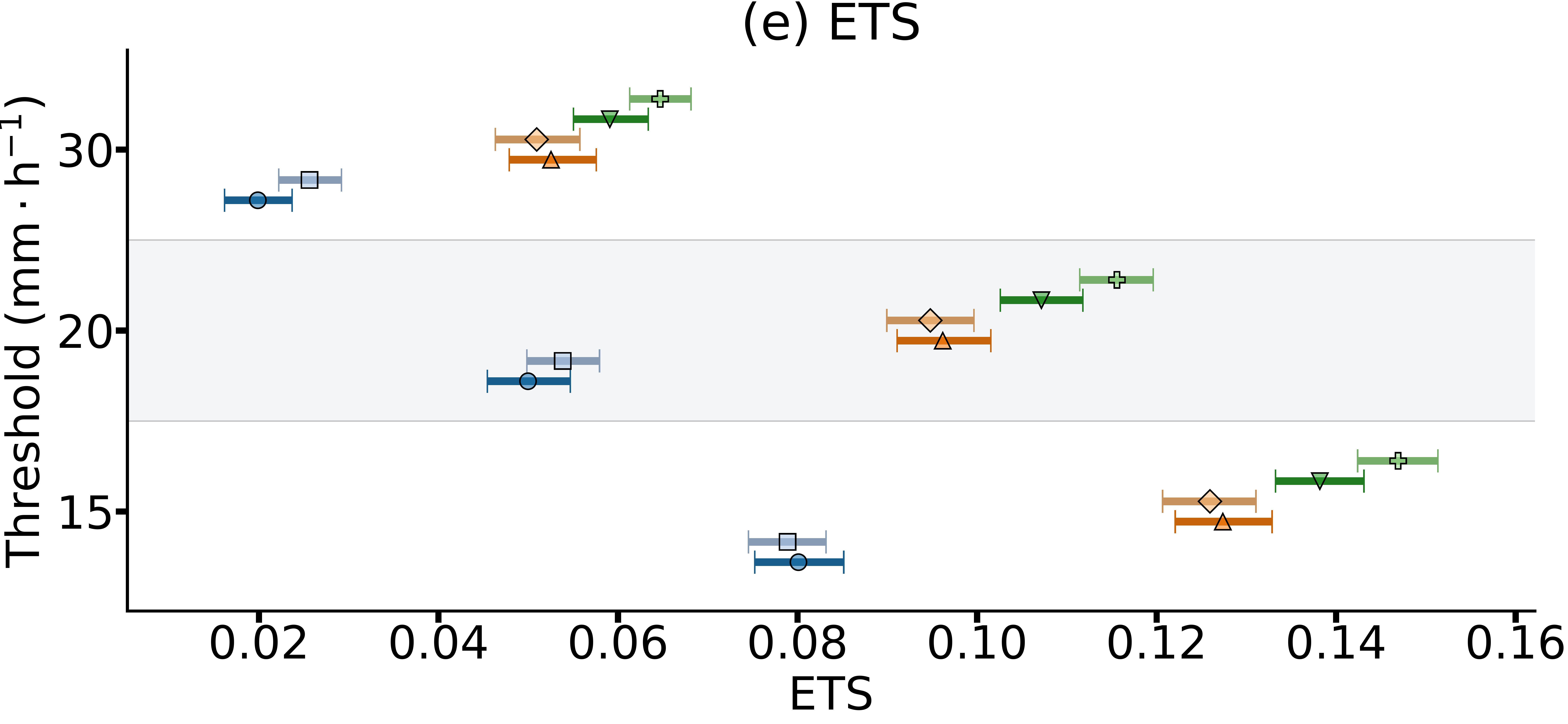}
    \includegraphics[width=0.46\textwidth]{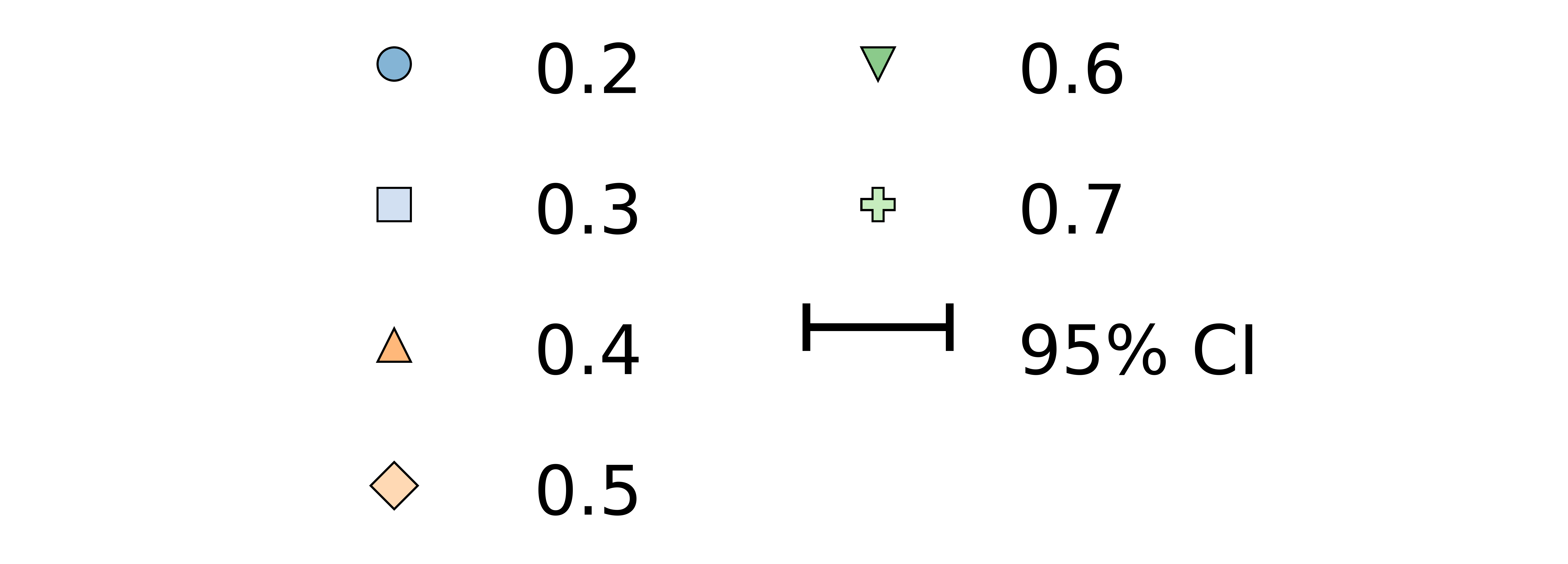}
    \caption{Performance obtained with different $\sigma$ values at thresholds of 15, 20, and 30 $\mathrm{mm}\cdot\mathrm{h}^{-1}$: (a) RMSE, (b) ME, (c) POD, (d) FAR, and (e) ETS. Error bars denote 95\% paired-bootstrap confidence intervals.}
    \label{fig:sigma_ci}
\end{figure*}

In the current Hurdle--RMIL framework, $\sigma$ is treated as a hyperparameter, the selection of which substantially influences model performance. Here, the sensitivity of $\sigma$ across rainfall thresholds is quantitatively evaluated. Fig.~\ref{fig:sigma_ana} summarizes the responses of RMSE, ME, POD, FAR, and ETS to variations in $\sigma$. At thresholds of 0.1--2 $\mathrm{mm}\cdot\mathrm{h}^{-1}$, RMSE generally increases with $\sigma$ (Fig.~\ref{fig:sigma_ana}a). At thresholds of 15--30 $\mathrm{mm}\cdot\mathrm{h}^{-1}$, the lowest RMSE values are obtained when $\sigma$ ranges from 0.4 to 0.6. Increasing $\sigma$ also reduces the negative ME while increasing both POD and FAR (Figs.~\ref{fig:sigma_ana}b--d). When $\sigma$ is set to 0.2, the ME at the 30 $\mathrm{mm}\cdot\mathrm{h}^{-1}$ threshold is $-30.20~\mathrm{mm}\cdot\mathrm{h}^{-1}$, indicating pronounced underestimation. Larger $\sigma$ values generally yield higher ETS at the 15, 20, and 30 $\mathrm{mm}\cdot\mathrm{h}^{-1}$ thresholds, but also increase FAR and, when $\sigma$ is set to 0.7, substantially increase RMSE. The 95\% confidence intervals in Fig.~\ref{fig:sigma_ci} do not affect the interpretation of the trends indicated by the above point estimates, while the overlap among several RMSE and ETS estimates for $\sigma$ values of 0.4--0.6 indicates that none of these intermediate settings consistently outperforms the others across all metrics. Therefore, rather than being determined by an isolated optimum, the selection of $\sigma$ should reflect the trade-off among error magnitude, bias, and detection performance. A value of 0.5 is used for $\sigma$ in the subsequent experiments.


\subsection{Effectiveness Analysis of RMIL} \label{section:RMIL_ana}

\begin{figure*}[!t]
    \centering
        \includegraphics[width=0.46\textwidth]{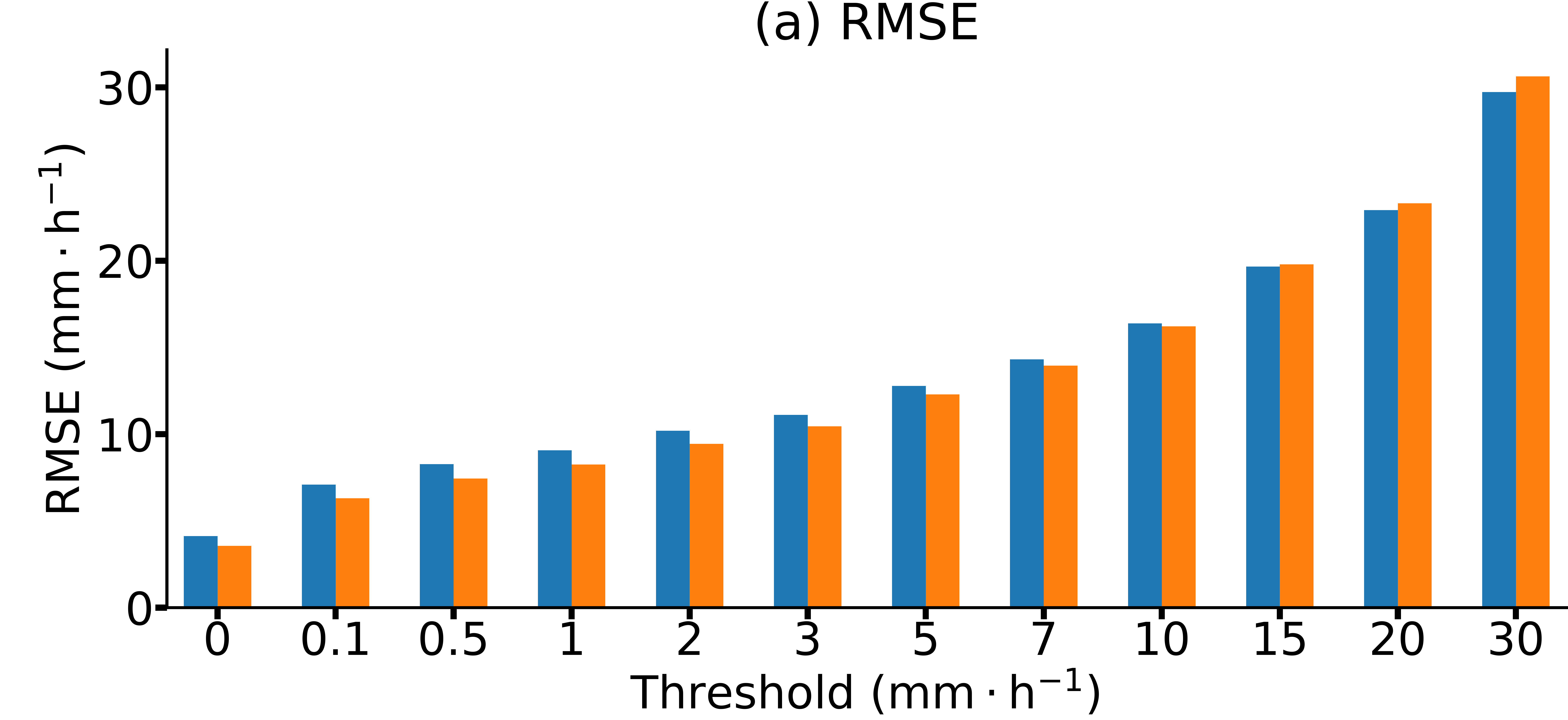}
        \includegraphics[width=0.46\textwidth]{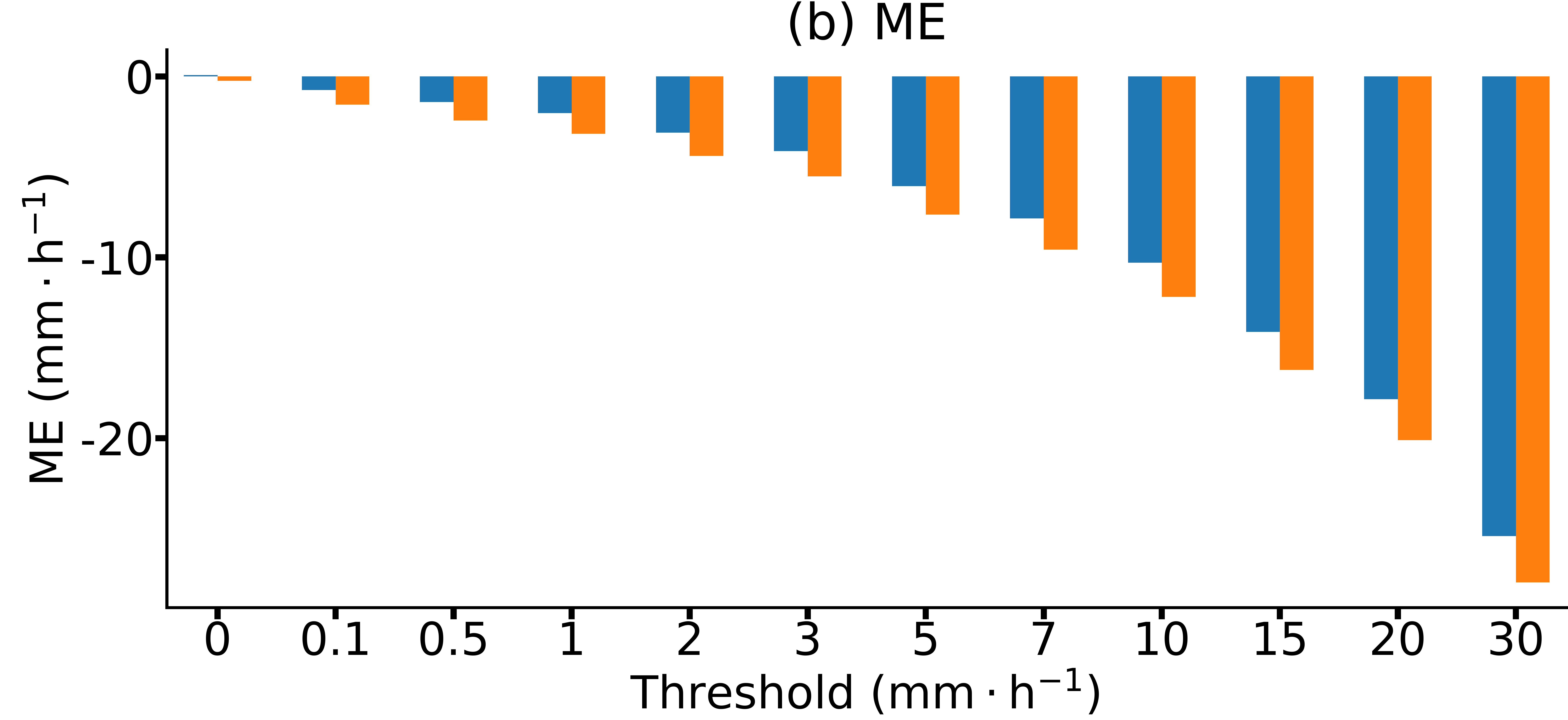}
        \par\vspace{1.5ex}
        \includegraphics[width=0.46\textwidth]{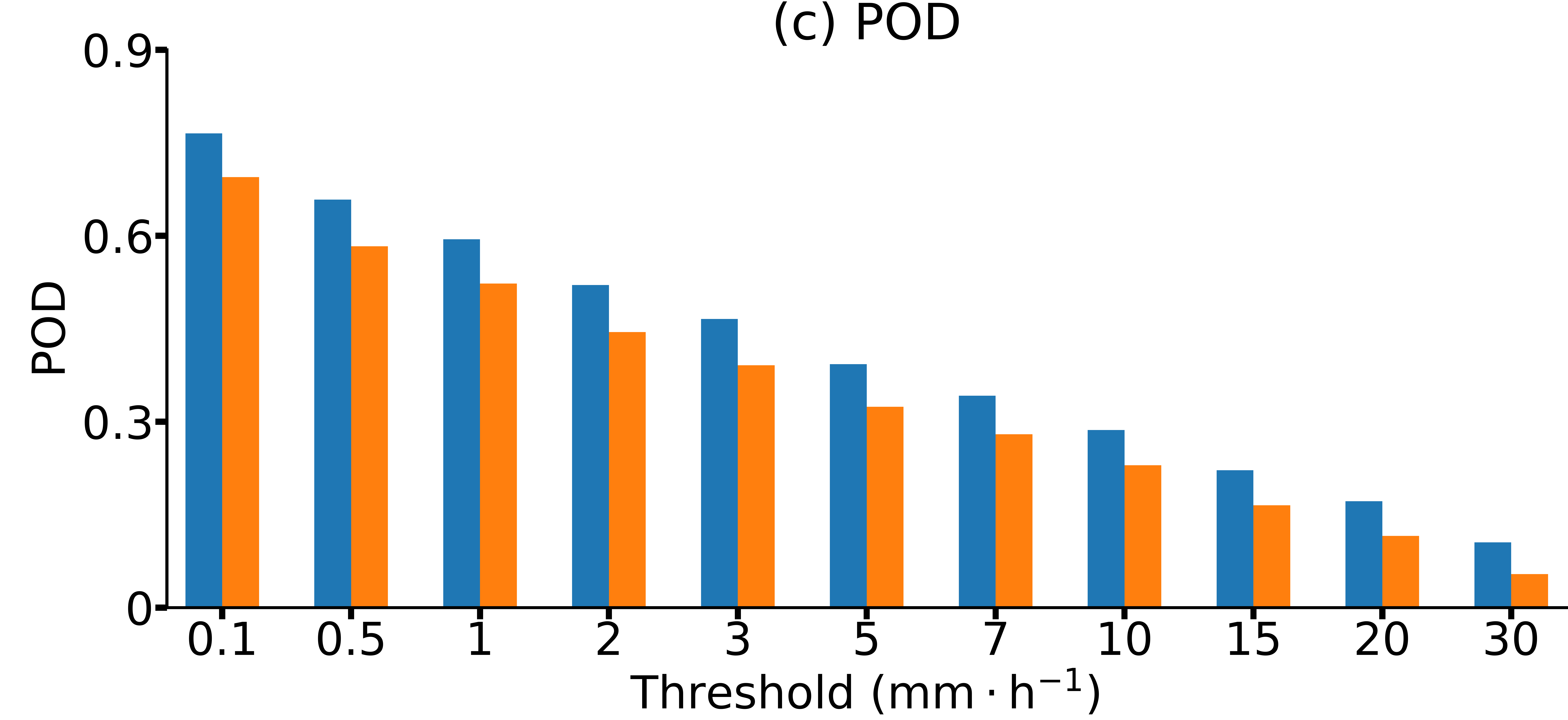}
        \includegraphics[width=0.46\textwidth]{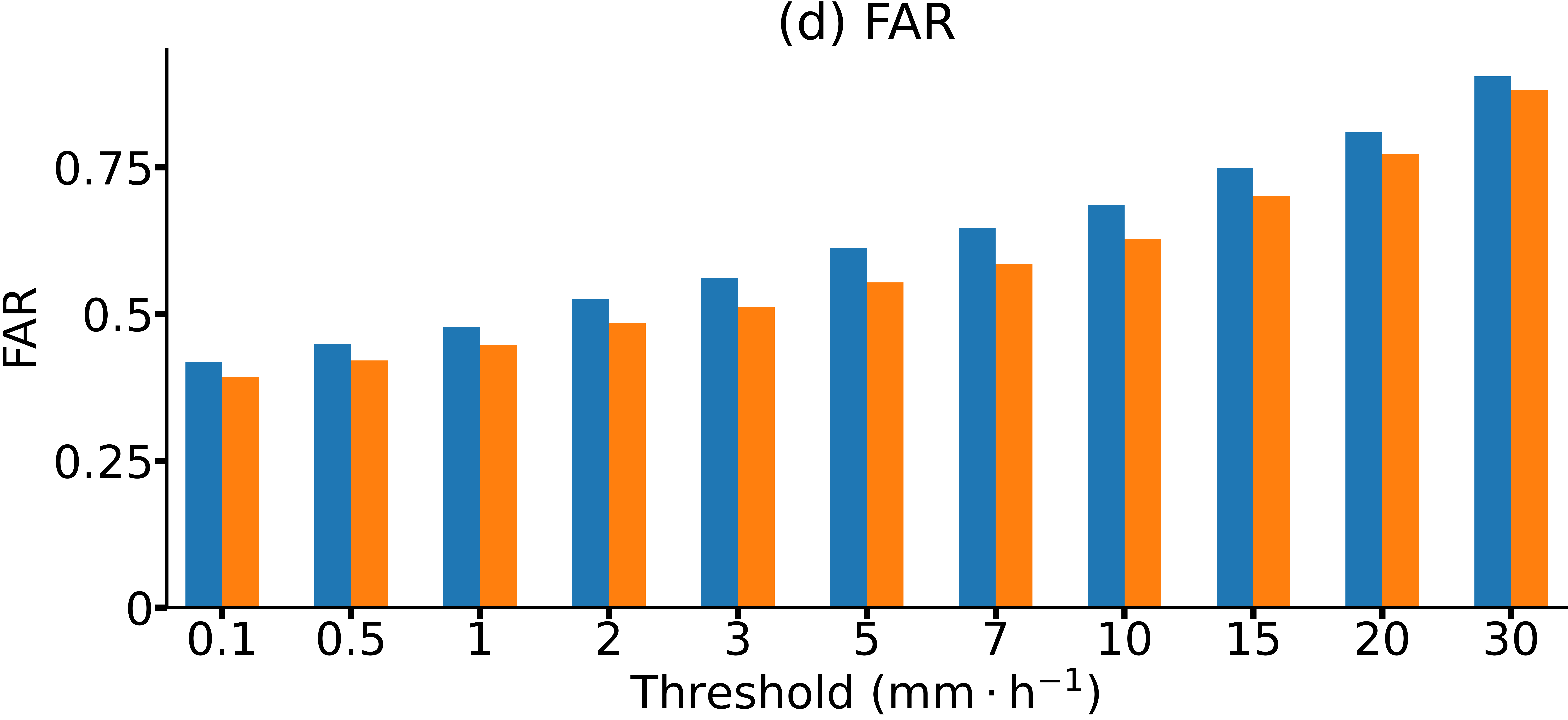}
        \par\vspace{1.5ex}
        \includegraphics[width=0.46\textwidth]{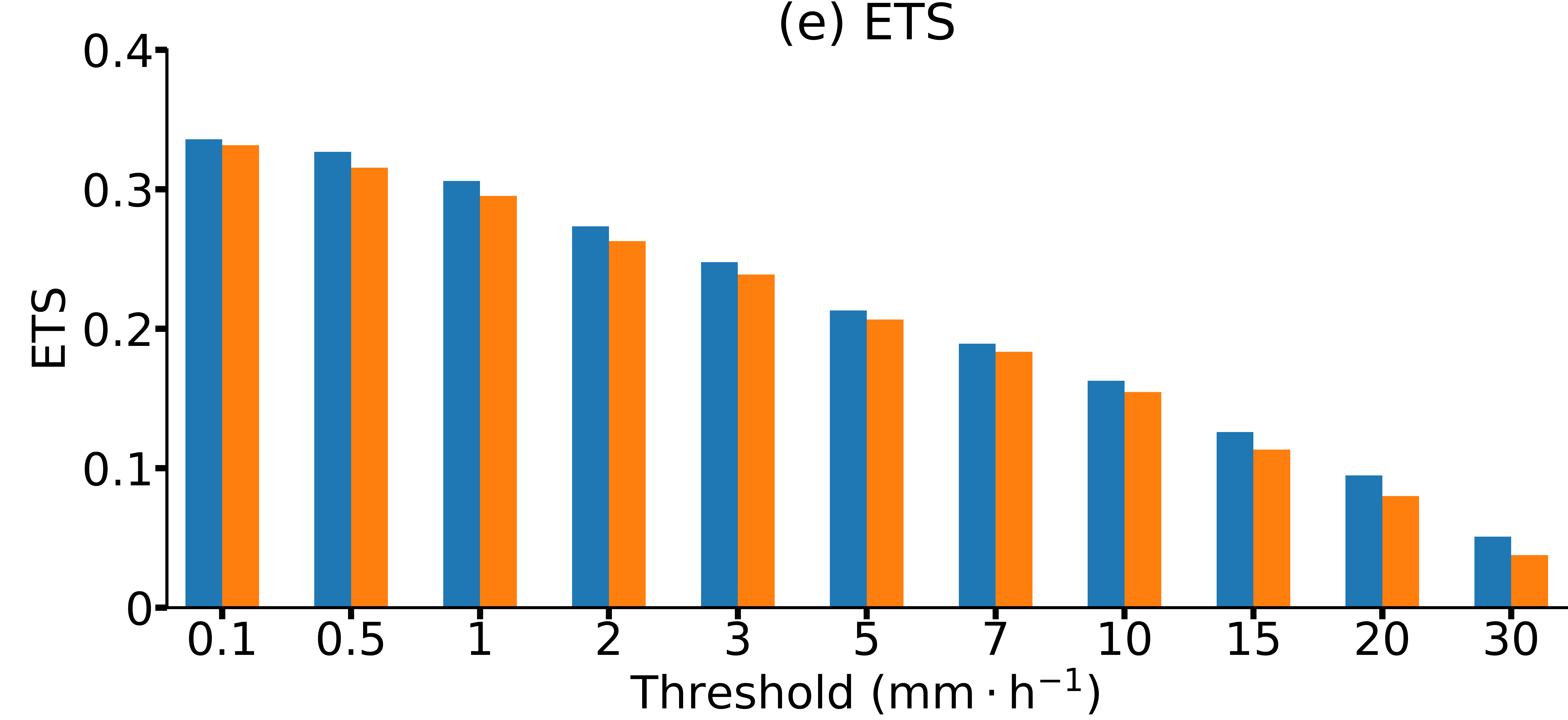}
        \includegraphics[width=0.46\textwidth]{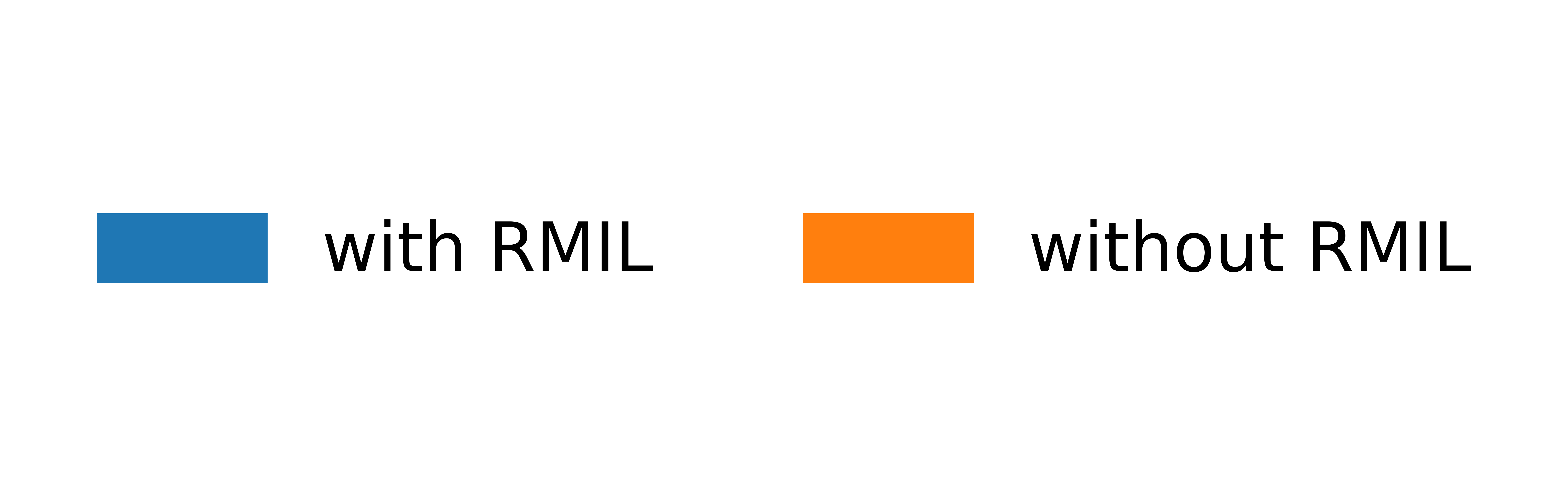}
    \caption{Ablation results with and without RMIL over the pooled test set: (a) RMSE, (b) ME, (c) POD, (d) FAR, and (e) ETS.}
    \label{fig:RMIL_ana}
\end{figure*}

\begin{figure*}[!t]
    \centering
    \includegraphics[width=0.46\textwidth]{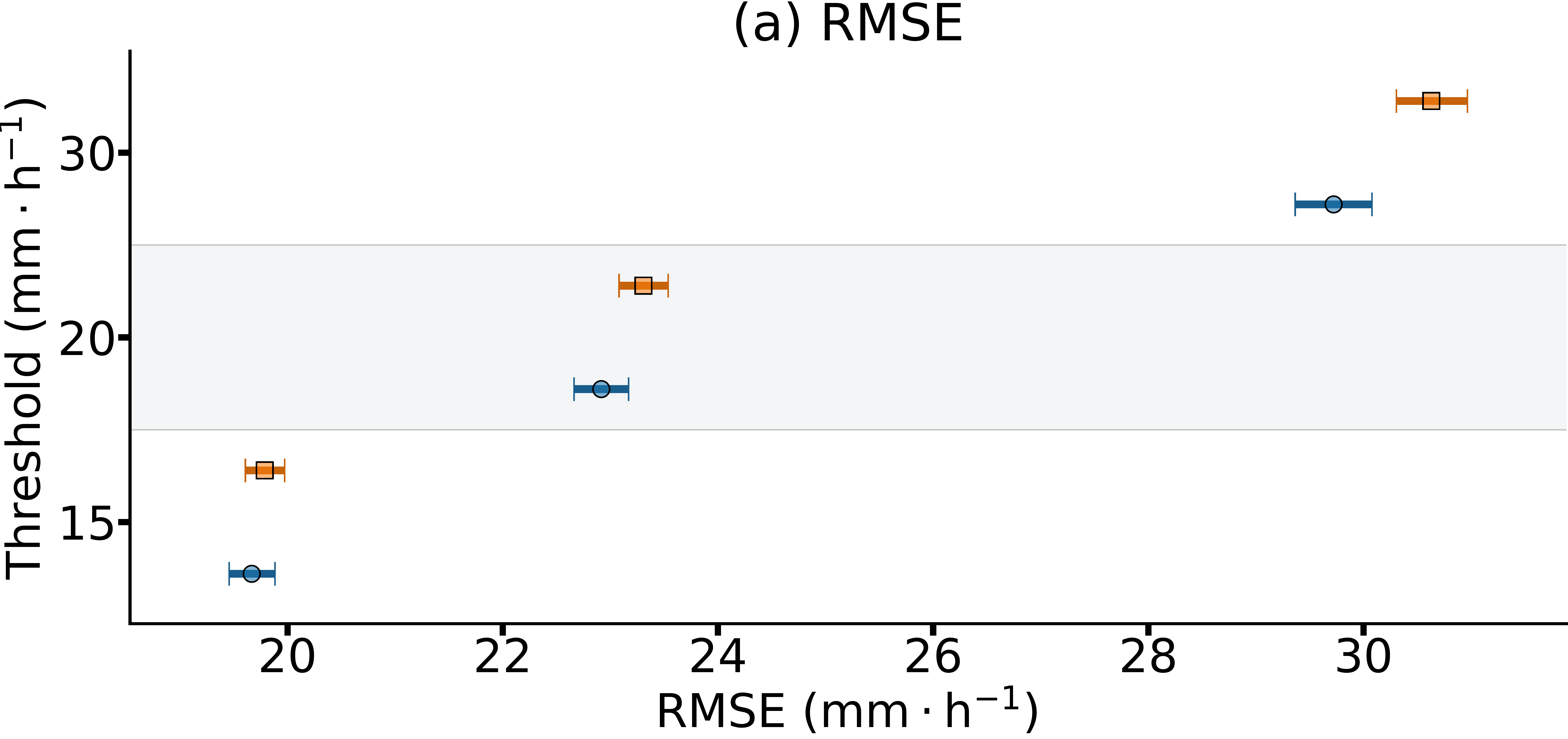}
    \includegraphics[width=0.46\textwidth]{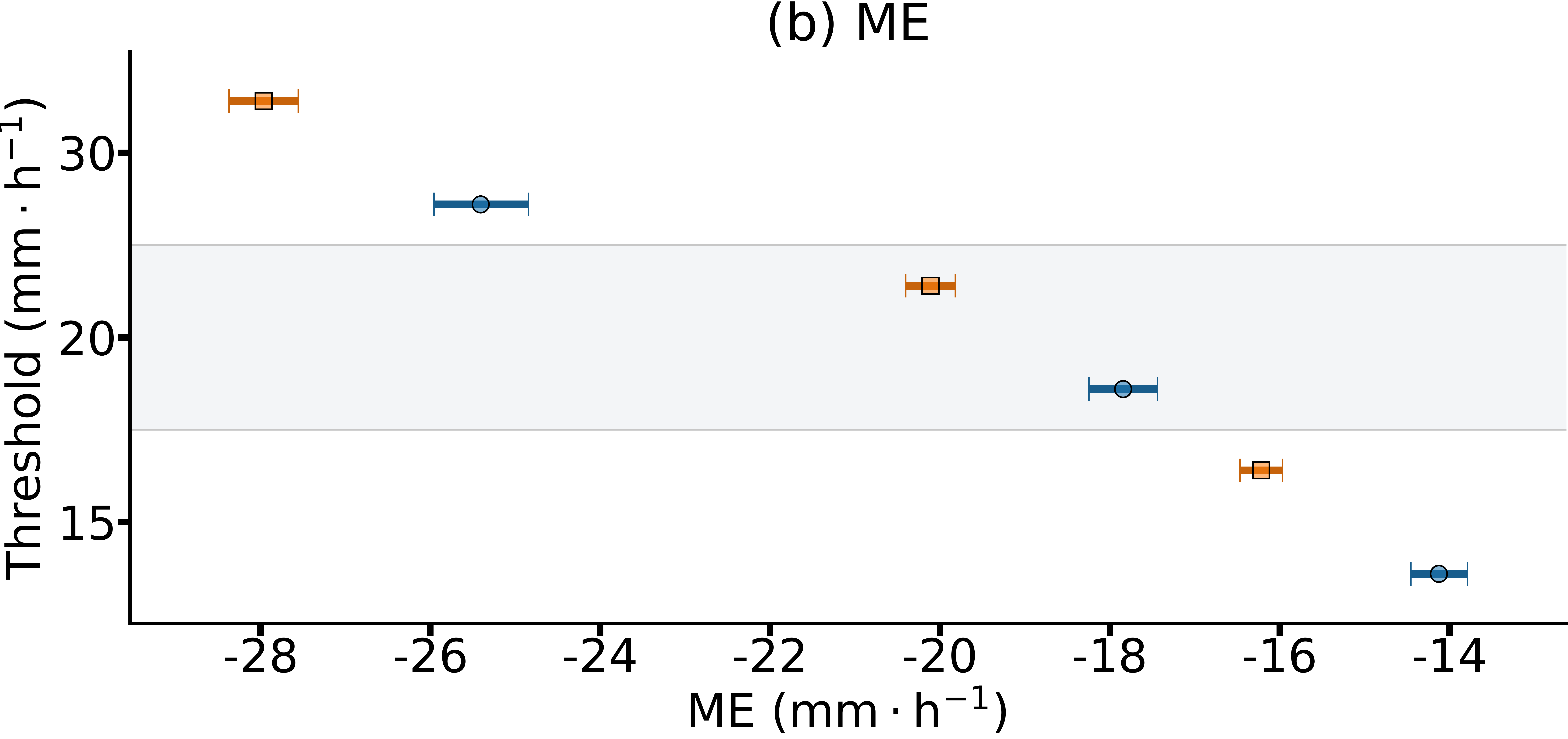}
    \par\vspace{1.5ex}
    \includegraphics[width=0.46\textwidth]{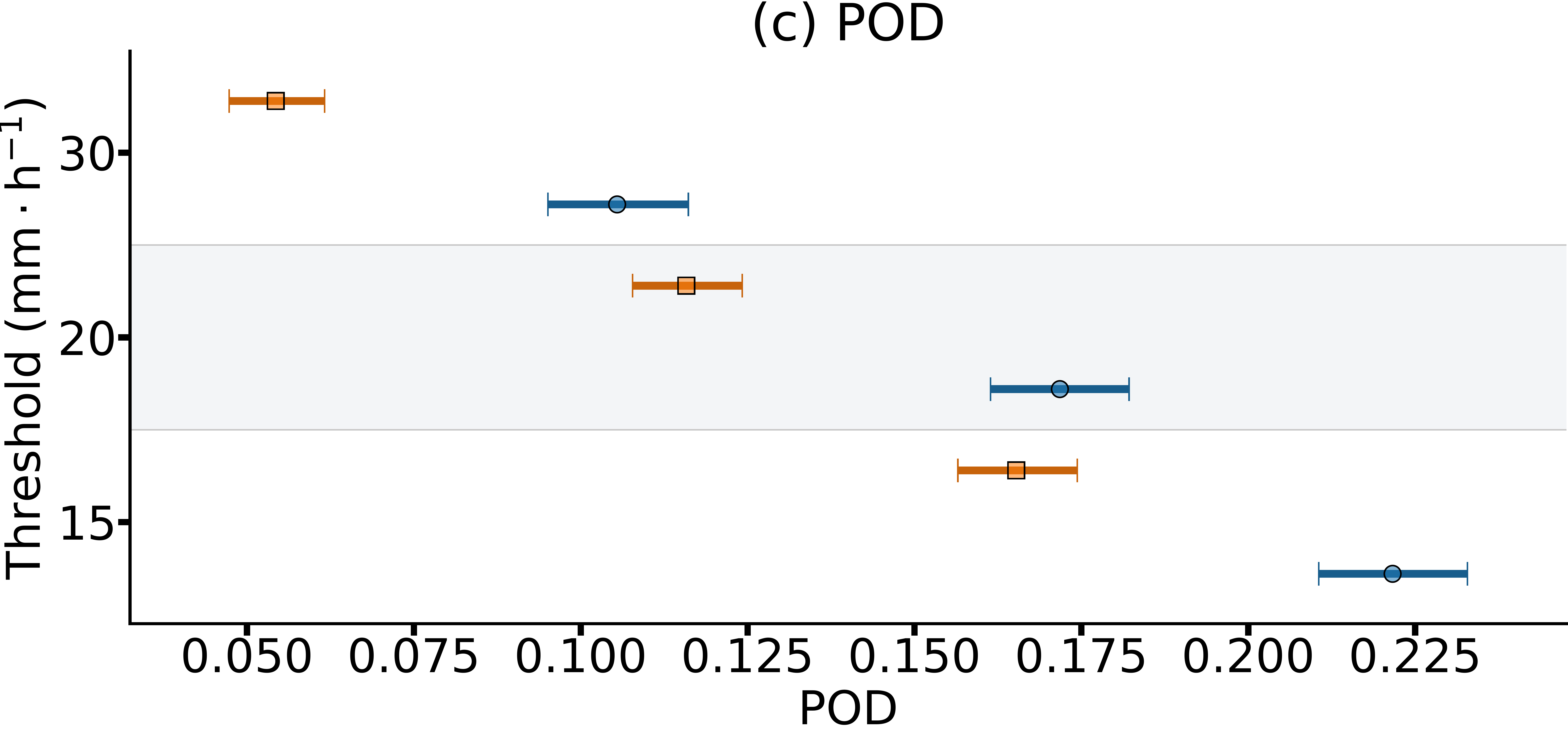}
    \includegraphics[width=0.46\textwidth]{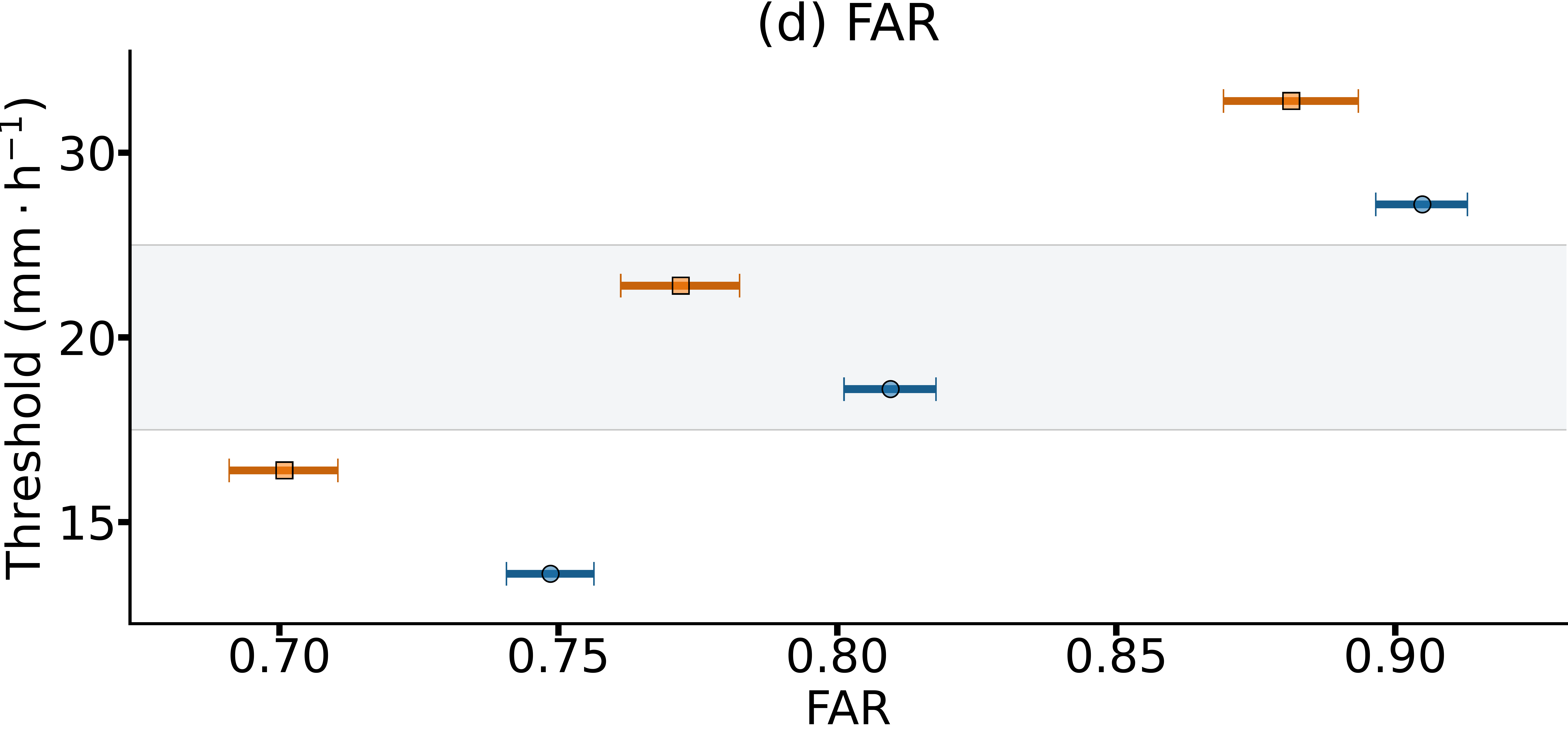}
    \par\vspace{1.5ex}
    \includegraphics[width=0.46\textwidth]{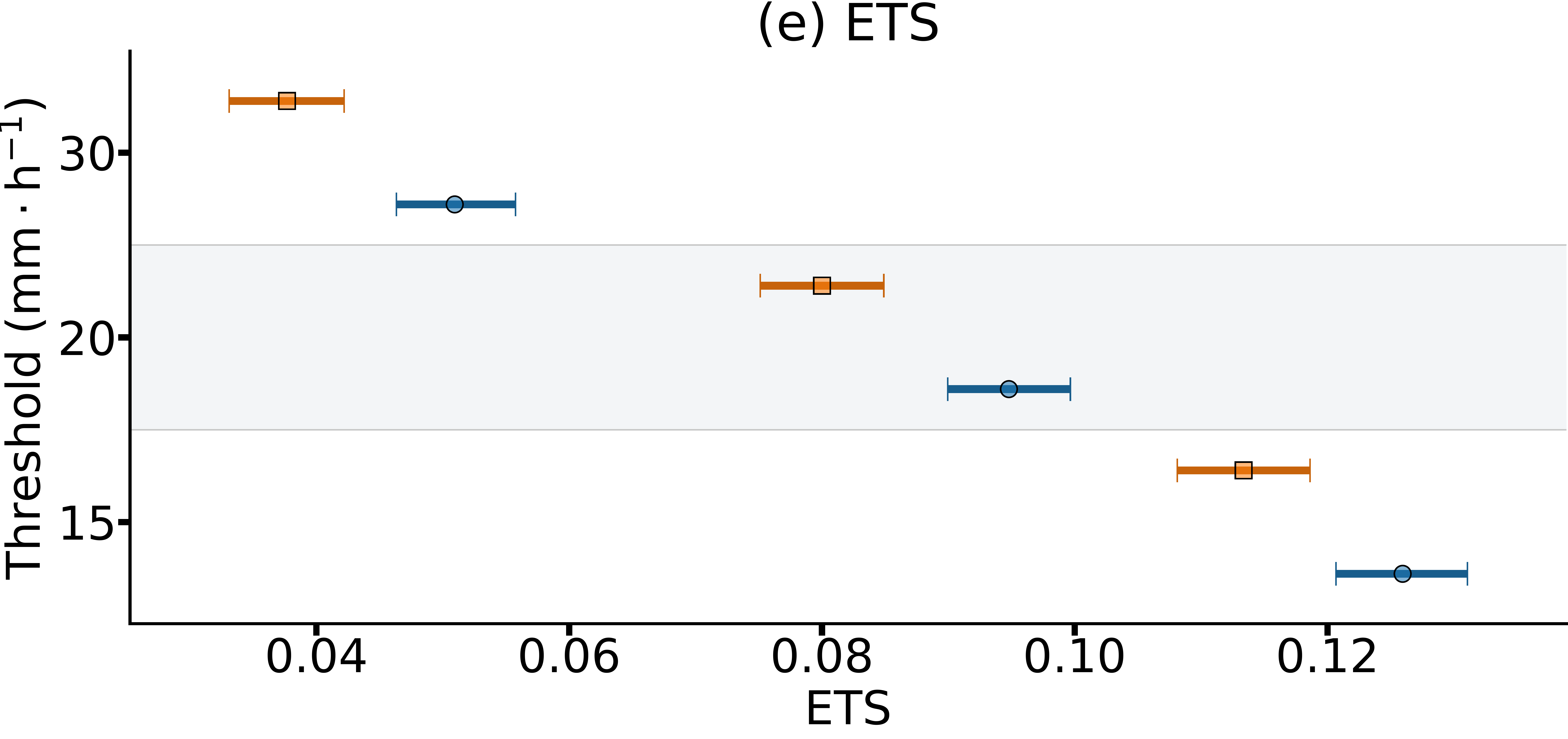}
    \includegraphics[width=0.46\textwidth]{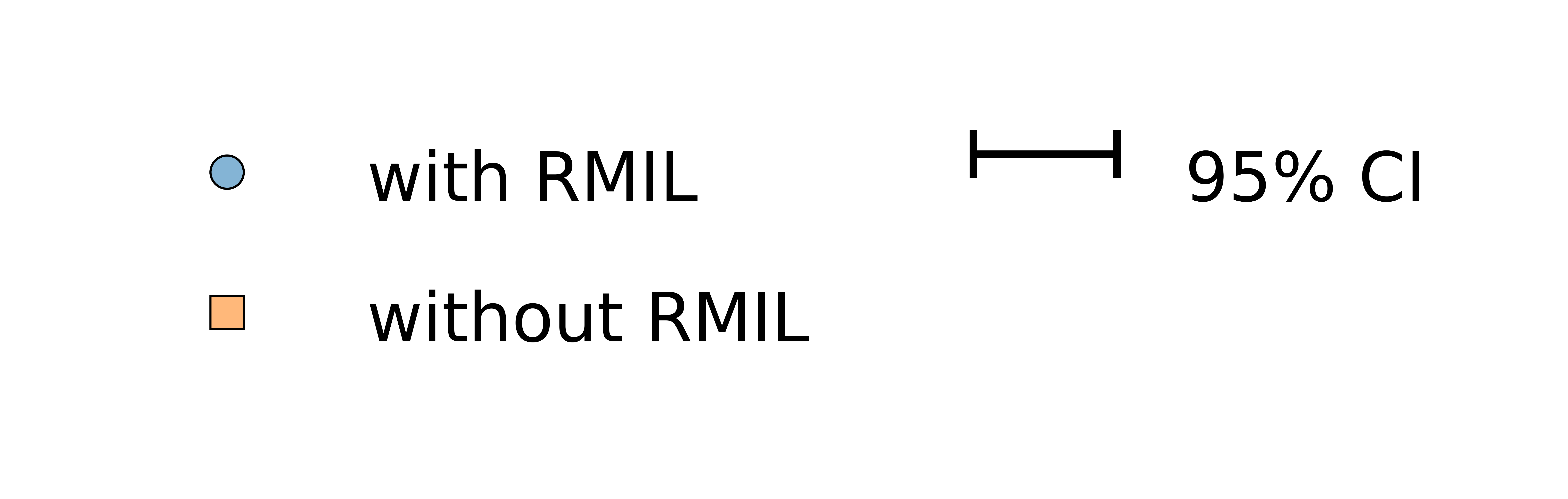}
    \caption{Ablation results with and without RMIL at thresholds of 15, 20, and 30 $\mathrm{mm}\cdot\mathrm{h}^{-1}$: (a) RMSE, (b) ME, (c) POD, (d) FAR, and (e) ETS. Error bars denote 95\% paired-bootstrap confidence intervals.}
    \label{fig:RMIL_ci}
\end{figure*}

RMIL uses the probability-density relationship between the retrieval models associated with long-tailed and balanced rainfall distributions to estimate the latter by maximum likelihood directly from naturally collected long-tailed samples. Its effectiveness is evaluated through the ablation analysis in Figs.~\ref{fig:RMIL_ana} and \ref{fig:RMIL_ci}, which compares RMSE, ME, POD, FAR, and ETS obtained with and without RMIL. To isolate the contribution of RMIL, both configurations use the same hurdle occurrence component, lognormal conditional model, joint parameter-estimation scheme, and fixed $\sigma$ of 0.5, differing only in whether RMIL is applied. As shown in Fig.~\ref{fig:RMIL_ana}a, the model with RMIL yields slightly higher RMSE at thresholds up to 10 $\mathrm{mm}\cdot\mathrm{h}^{-1}$. At thresholds of 15--30 $\mathrm{mm}\cdot\mathrm{h}^{-1}$, however, RMIL reduces RMSE; at 30 $\mathrm{mm}\cdot\mathrm{h}^{-1}$, RMSE decreases from 30.63 to 29.72 $\mathrm{mm}\cdot\mathrm{h}^{-1}$. Regarding ME, both models exhibit systematic underestimation, whereas RMIL consistently reduces the magnitude of the negative bias (Fig.~\ref{fig:RMIL_ana}b). RMIL also yields higher POD and FAR point estimates across all evaluated thresholds (Figs.~\ref{fig:RMIL_ana}c and \ref{fig:RMIL_ana}d). Finally, RMIL achieves higher ETS point estimates across all evaluated thresholds (Fig.~\ref{fig:RMIL_ana}e). At the 30 $\mathrm{mm}\cdot\mathrm{h}^{-1}$ threshold, ETS increases from 0.038 to 0.051. As shown in Fig.~\ref{fig:RMIL_ci}, the RMSE intervals overlap at 15 and 20 $\mathrm{mm}\cdot\mathrm{h}^{-1}$, indicating that the corresponding reductions in RMSE are not clearly distinguishable from sampling uncertainty; their separation at 30 $\mathrm{mm}\cdot\mathrm{h}^{-1}$ provides stronger support for the reduction in RMSE at this threshold. The nonoverlapping intervals for ME, POD, FAR, and ETS at all three thresholds indicate that the corresponding differences between the two models are less affected by sampling uncertainty. These results demonstrate that RMIL effectively mitigates underestimation and improves rainfall detection at high thresholds.

\subsection{Comparison of Parameter-Estimation Approaches} \label{section:single_model_ana}

\begin{figure*}[!t]
    \centering
        \includegraphics[width=0.46\textwidth]{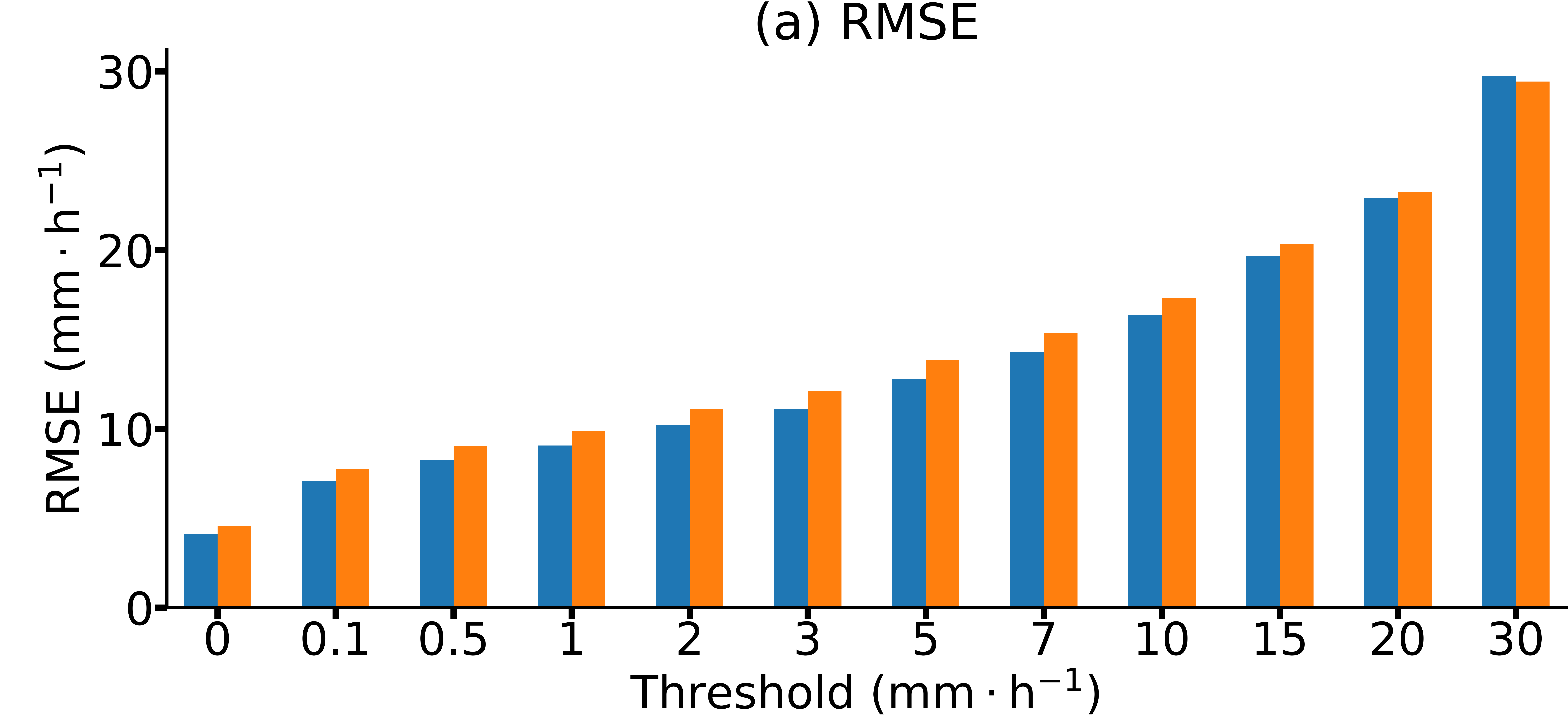}
        \includegraphics[width=0.46\textwidth]{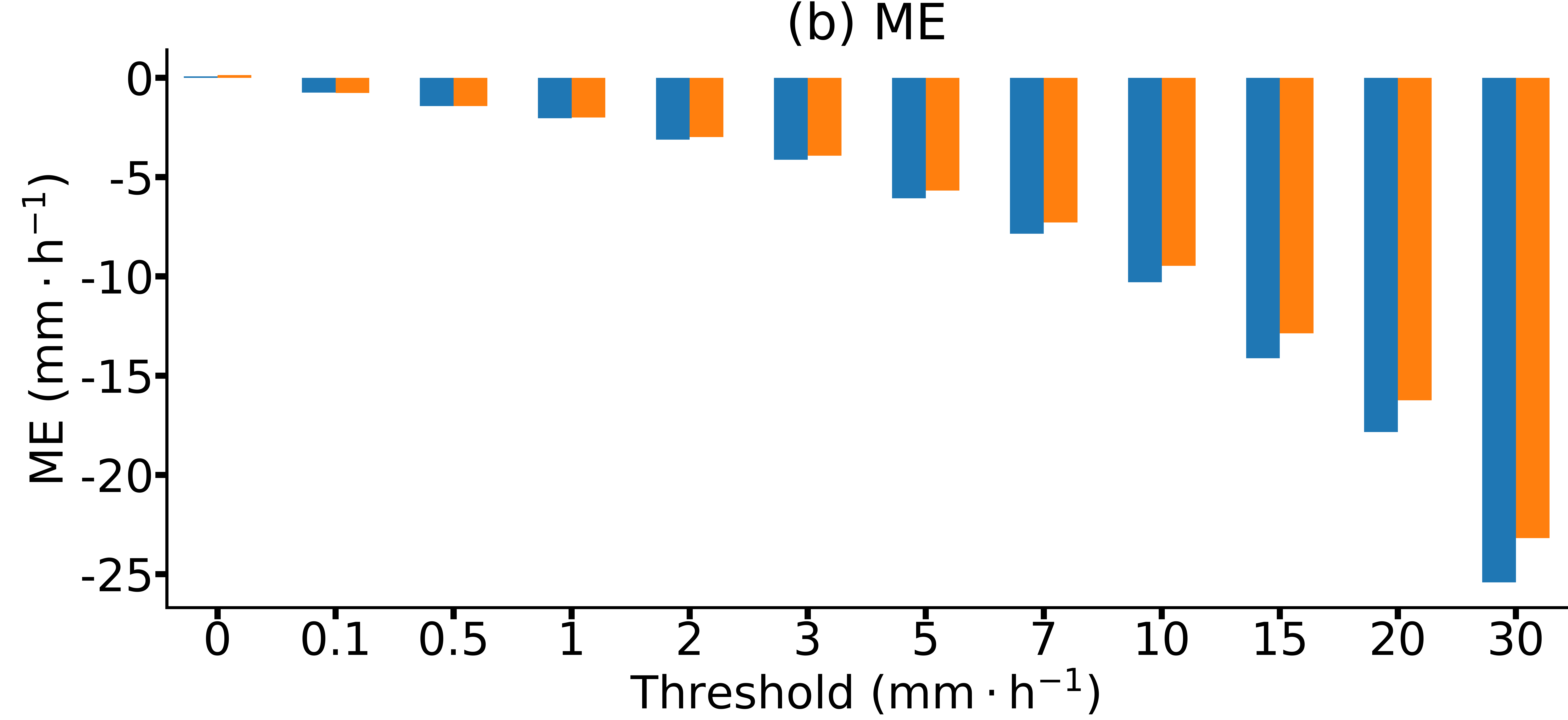}
        \par\vspace{1.5ex}
        \includegraphics[width=0.46\textwidth]{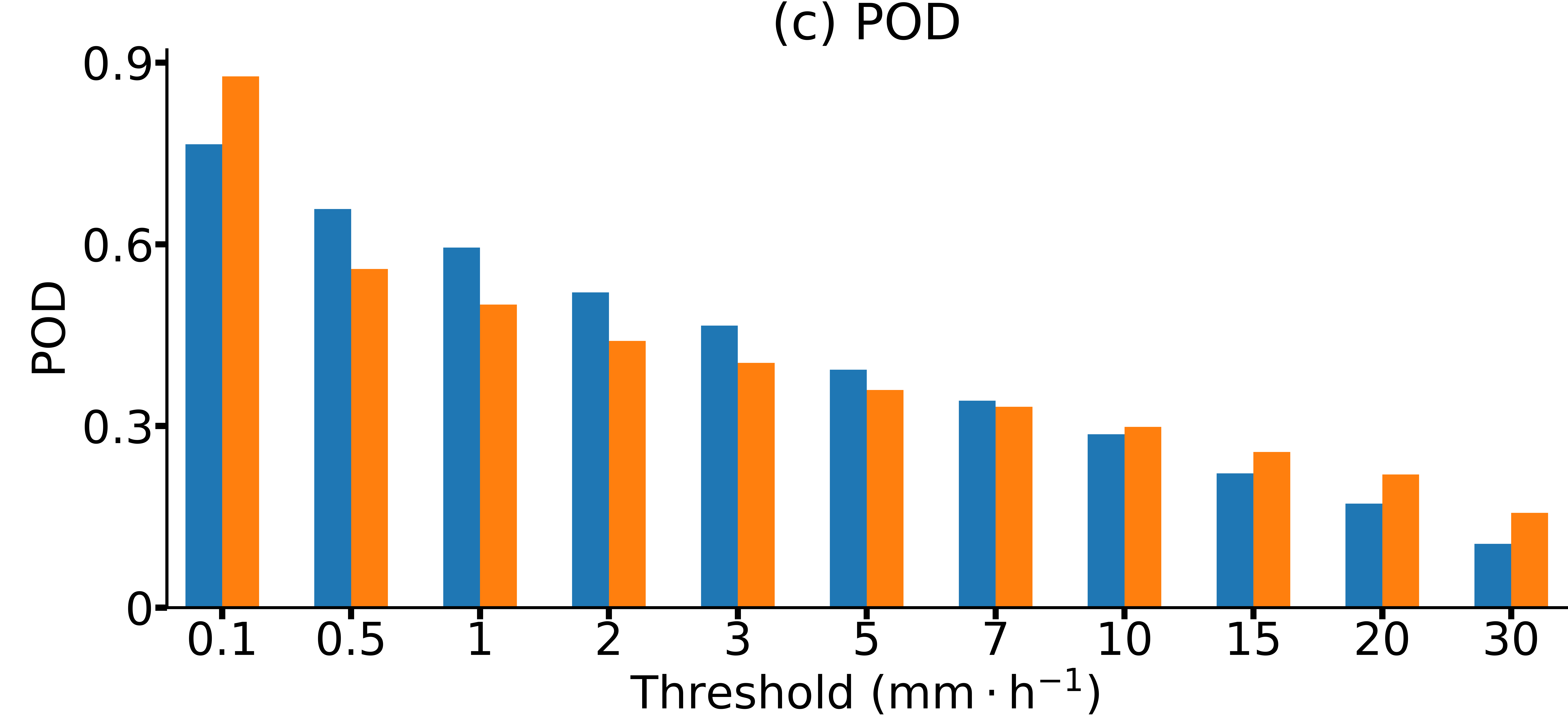}
        \includegraphics[width=0.46\textwidth]{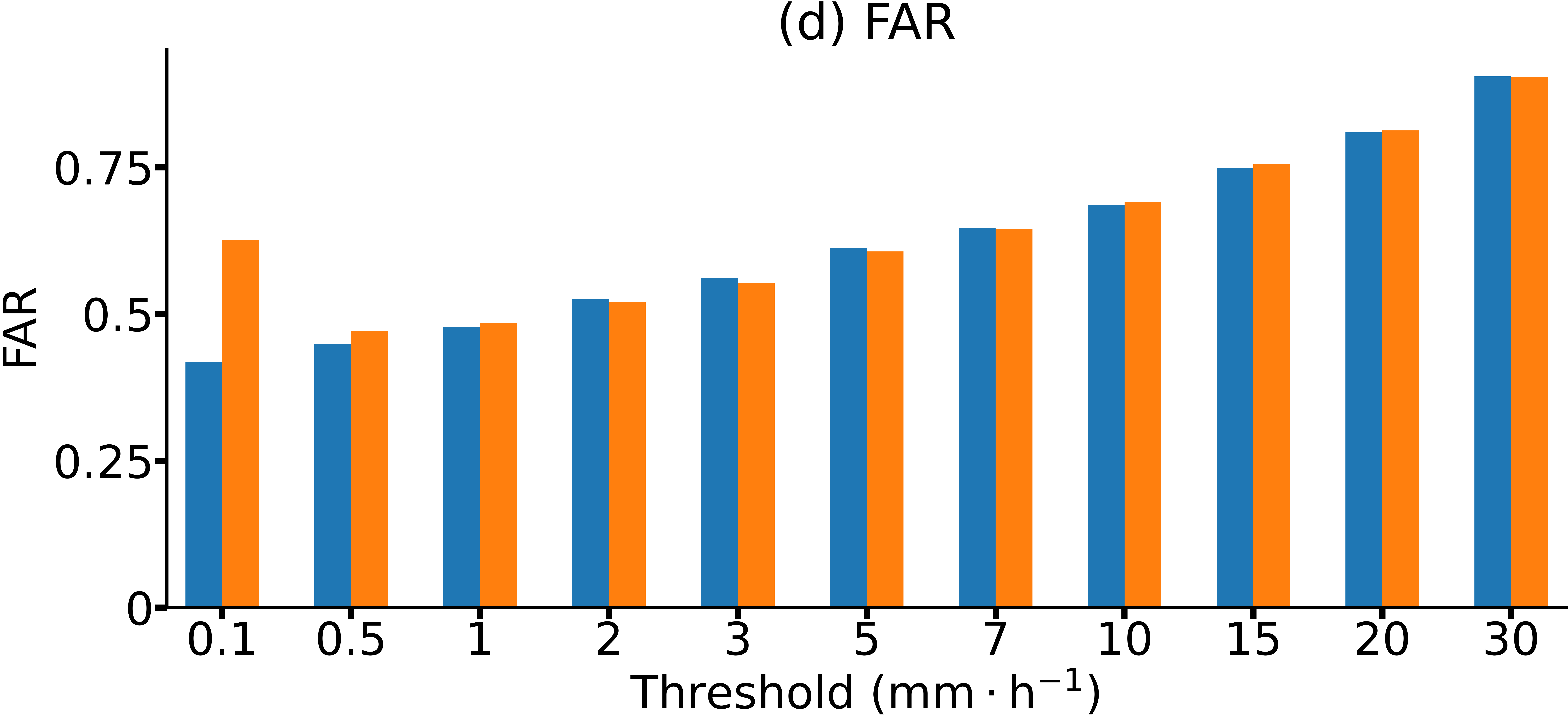}
        \par\vspace{1.5ex}
        \includegraphics[width=0.46\textwidth]{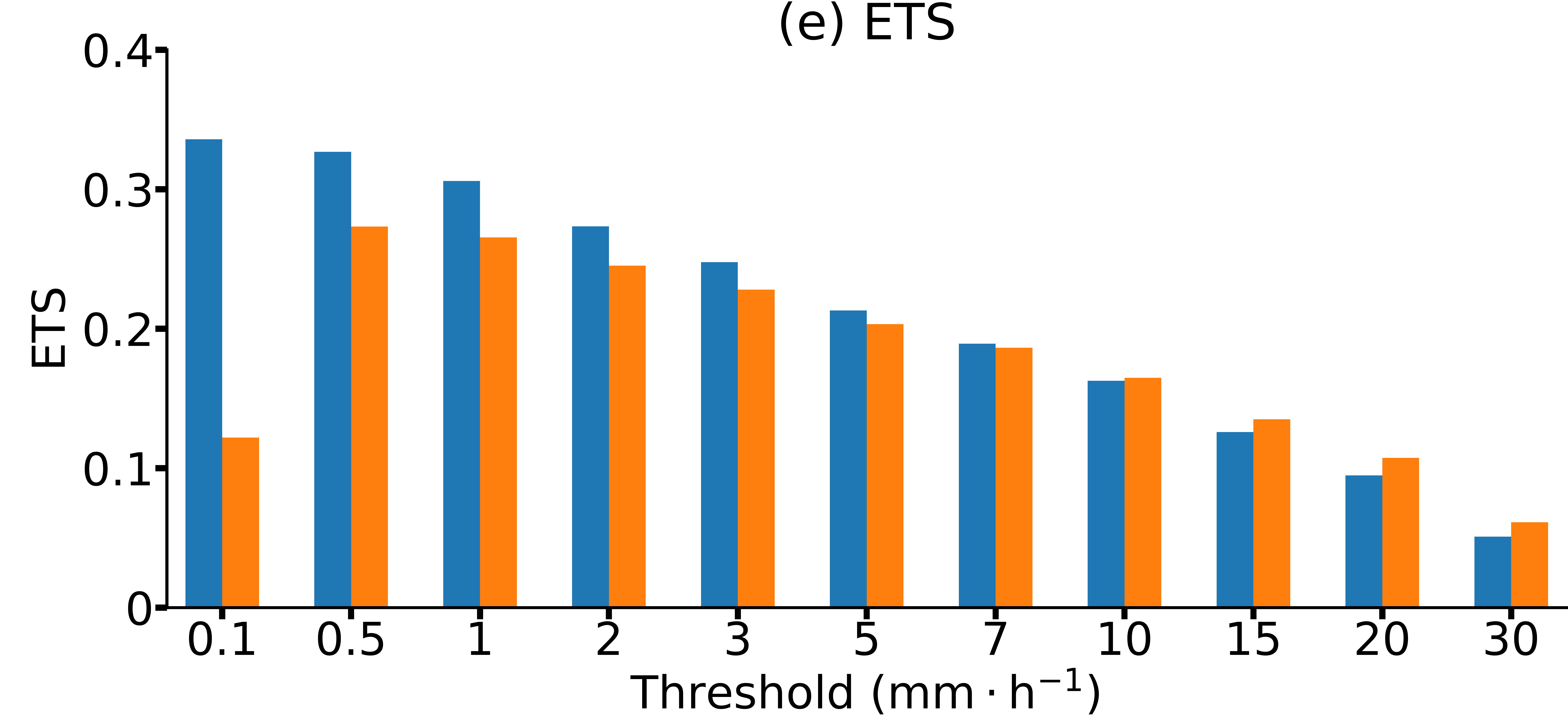}
        \includegraphics[width=0.46\textwidth]{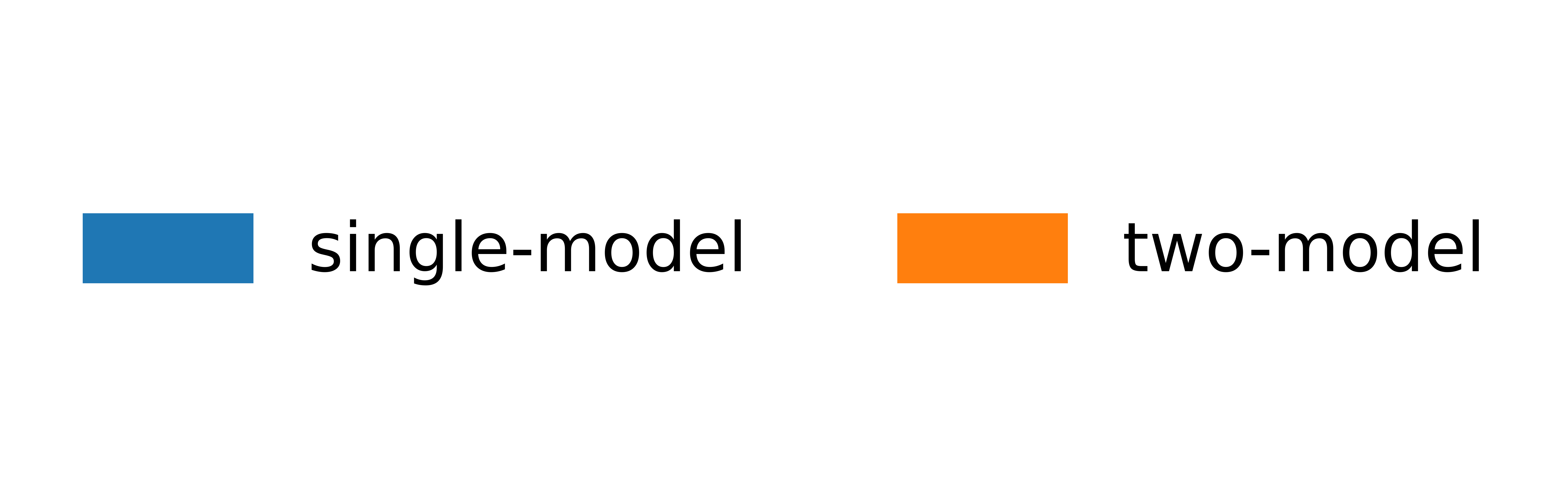}
    \caption{Comparison of parameter-estimation approaches over the pooled test set: (a) RMSE, (b) ME, (c) POD, (d) FAR, and (e) ETS. The single-model approach estimates $p$ and $\mu$ jointly, whereas the two-model approach estimates them sequentially.}
    \label{fig:single_ana}
\end{figure*}

\begin{figure*}[!t]
    \centering
    \includegraphics[width=0.46\textwidth]{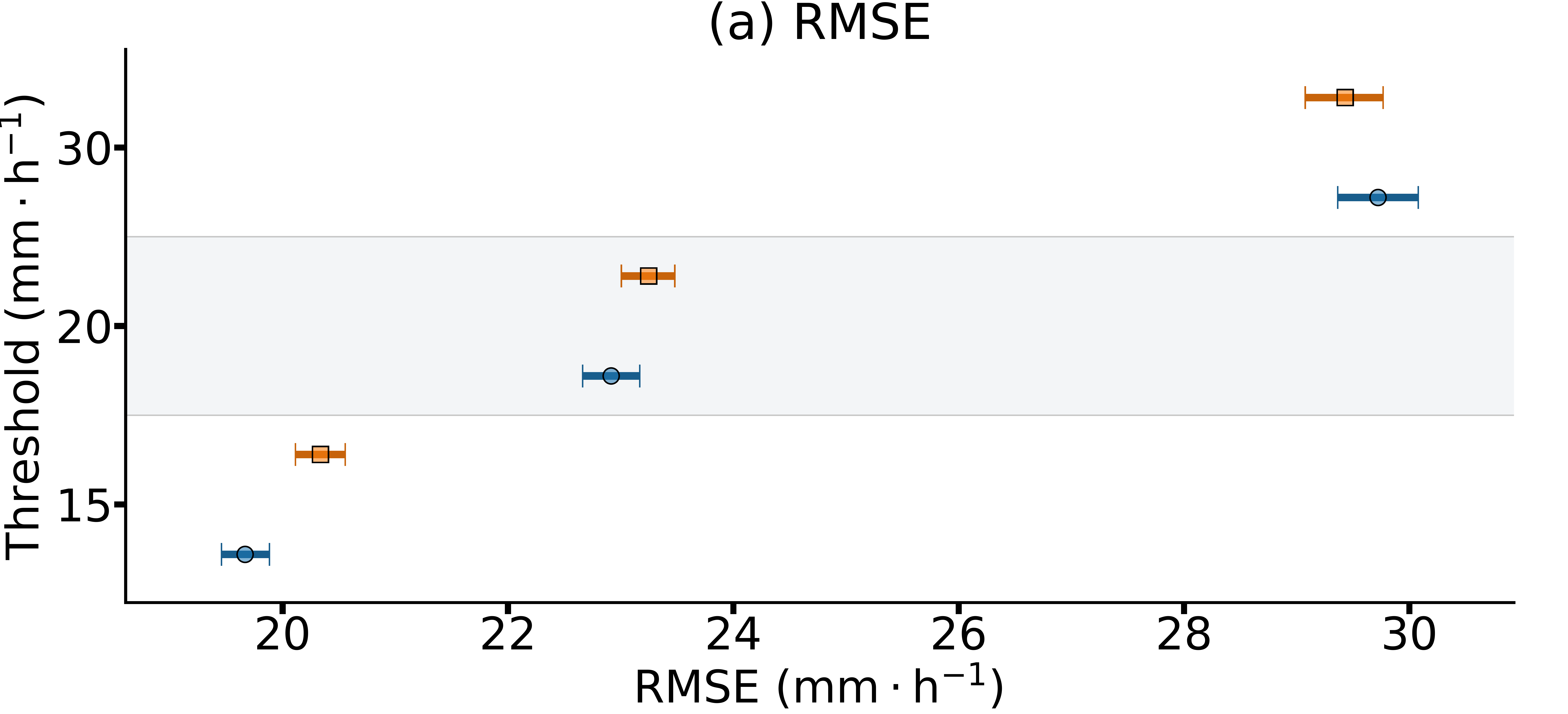}
    \includegraphics[width=0.46\textwidth]{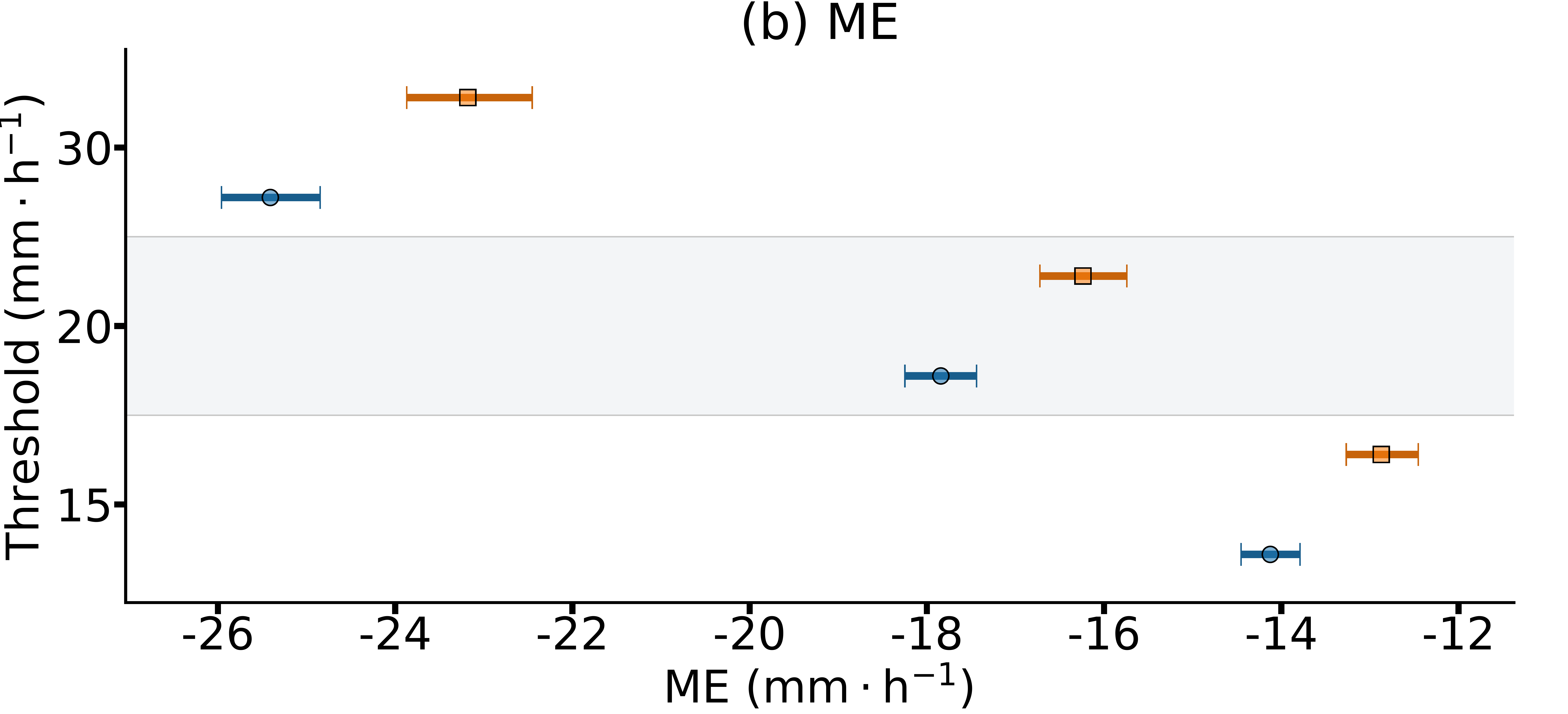}
    \par\vspace{1.5ex}
    \includegraphics[width=0.46\textwidth]{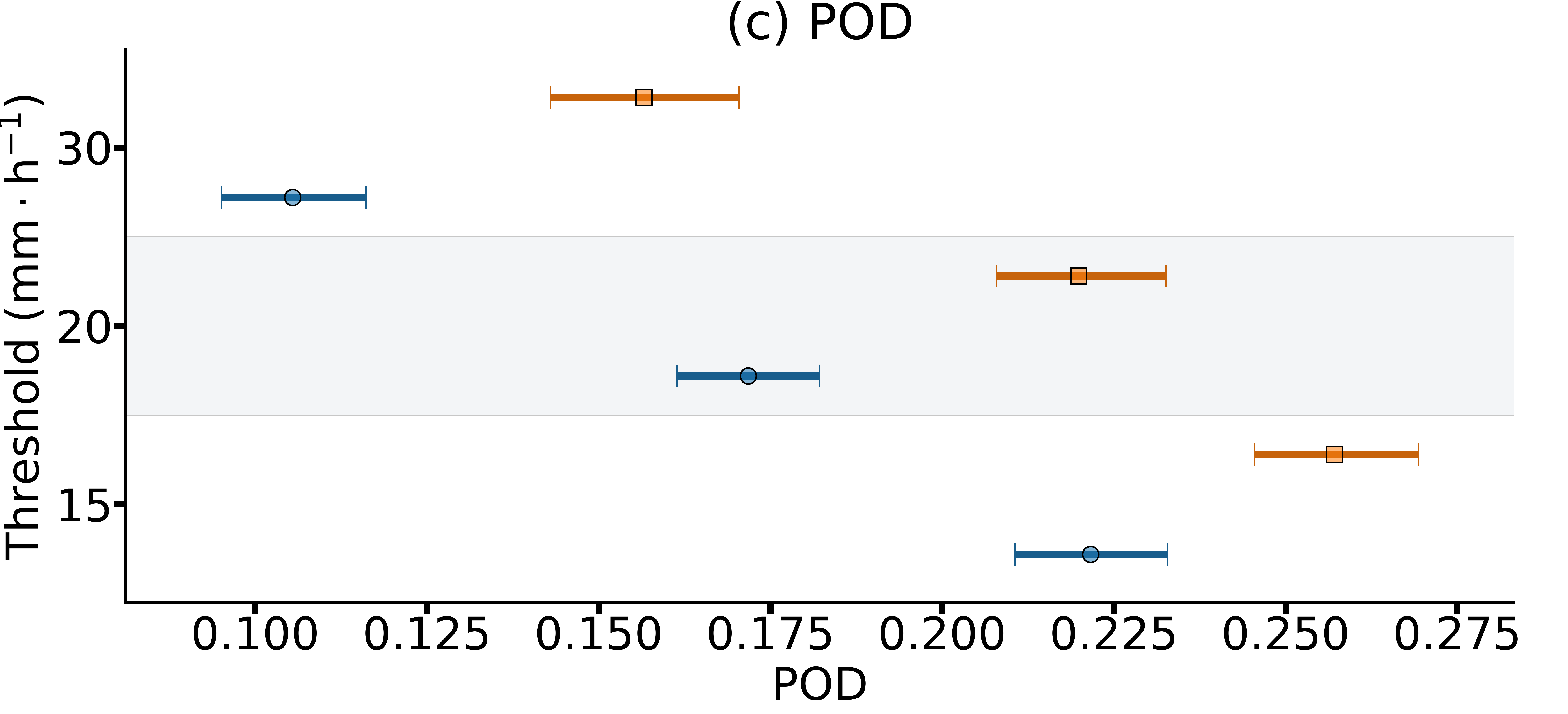}
    \includegraphics[width=0.46\textwidth]{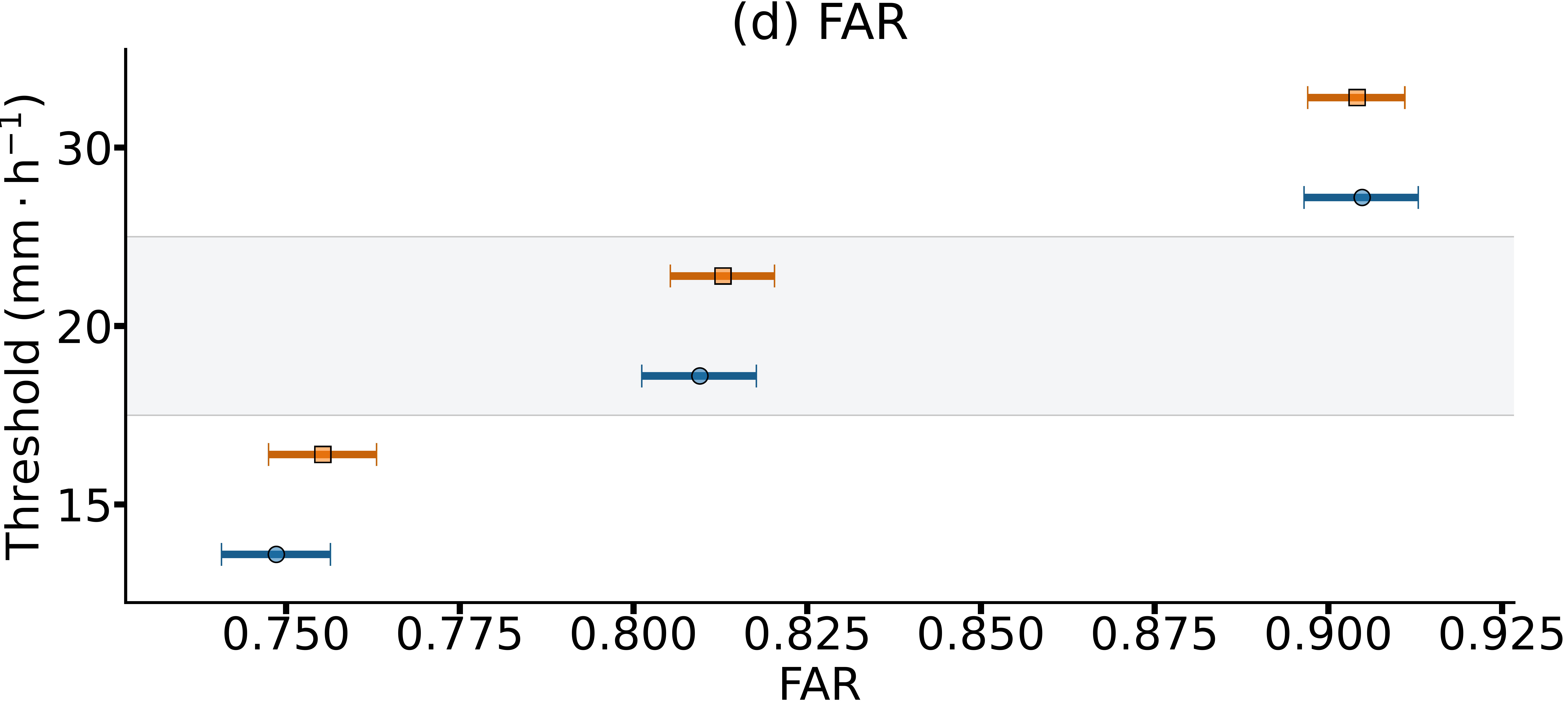}
    \par\vspace{1.5ex}
    \includegraphics[width=0.46\textwidth]{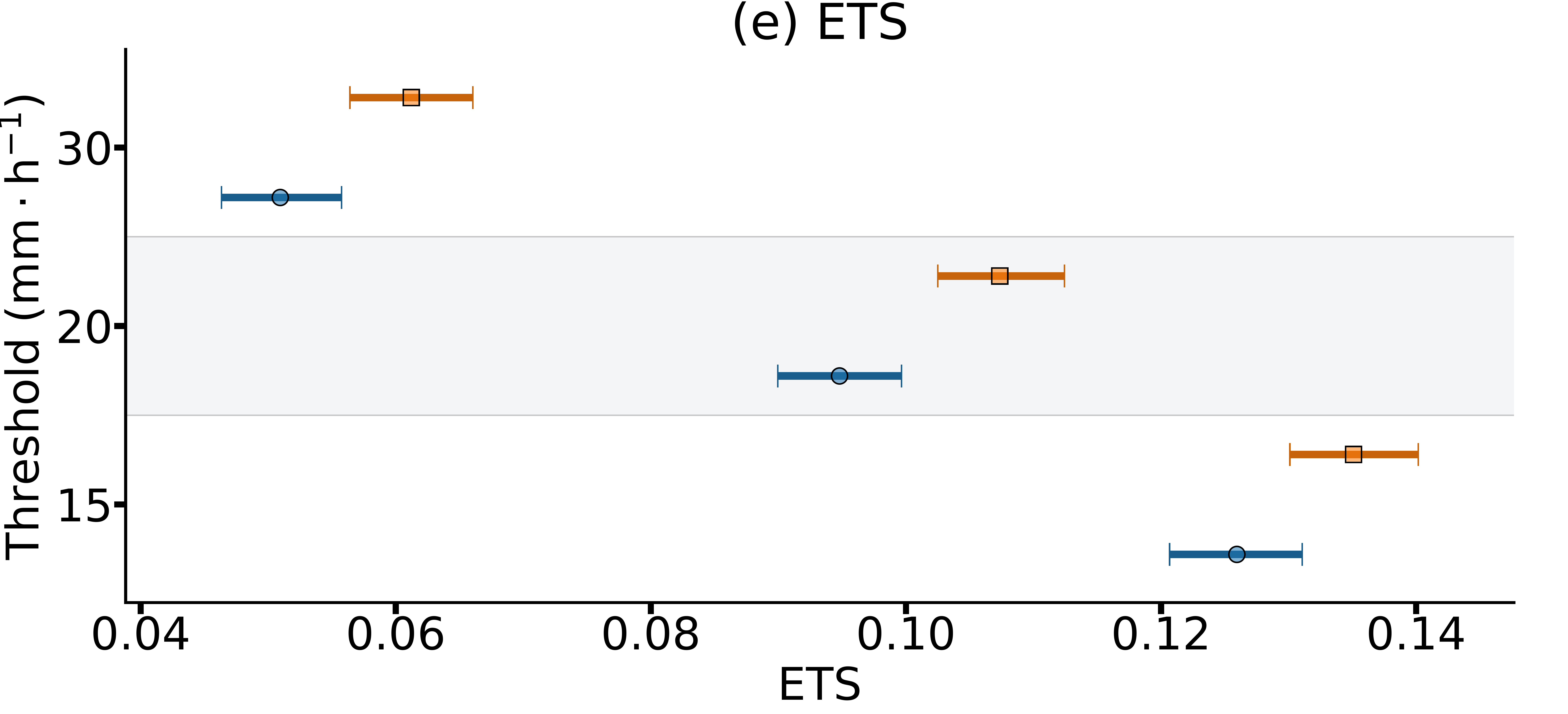}
    \includegraphics[width=0.46\textwidth]{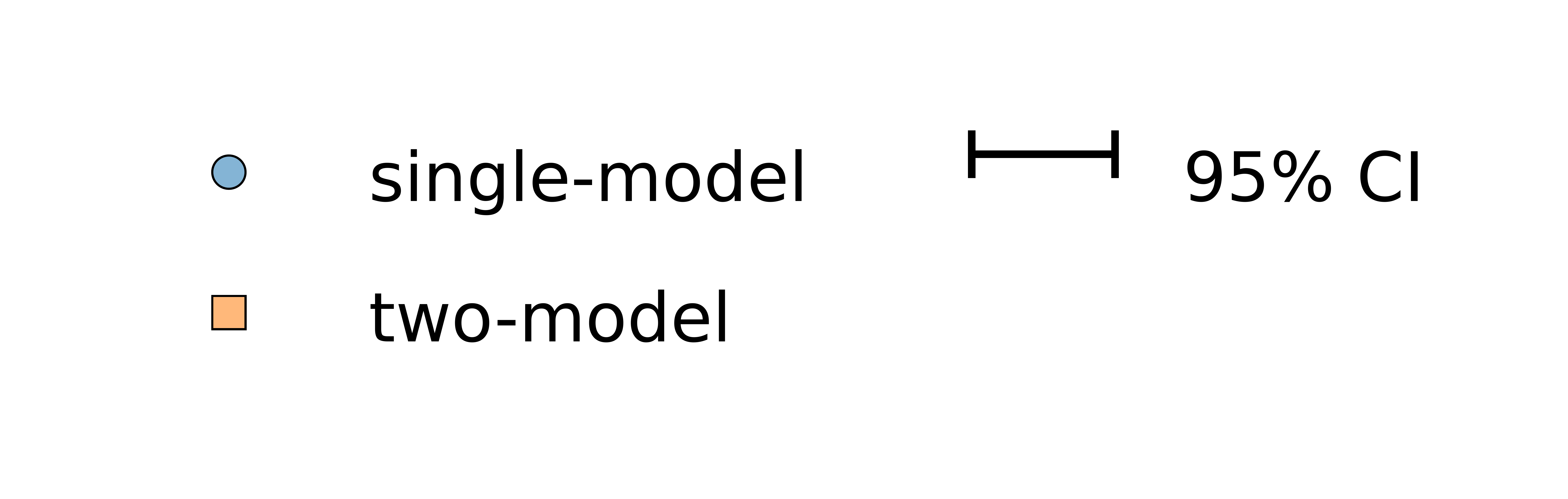}
    \caption{Comparison of parameter-estimation approaches at thresholds of 15, 20, and 30 $\mathrm{mm}\cdot\mathrm{h}^{-1}$: (a) RMSE, (b) ME, (c) POD, (d) FAR, and (e) ETS. Error bars denote 95\% paired-bootstrap confidence intervals.}
    \label{fig:single_ci}
\end{figure*}

Figure~\ref{fig:single_ana} compares the joint estimation of $p$ and $\mu$ using a single model (single-model approach) with their sequential estimation using two separate models (two-model approach), with $\sigma = 0.5$. As shown in Fig.~\ref{fig:single_ana}a, the single-model approach yields lower RMSE at thresholds of 0.1--20 $\mathrm{mm}\cdot\mathrm{h}^{-1}$; at 10 $\mathrm{mm}\cdot\mathrm{h}^{-1}$, for example, RMSE is 16.39, compared with 17.32 $\mathrm{mm}\cdot\mathrm{h}^{-1}$ for the two-model approach. At 30 $\mathrm{mm}\cdot\mathrm{h}^{-1}$, the two approaches yield similar RMSE values of 29.72 and 29.43 $\mathrm{mm}\cdot\mathrm{h}^{-1}$, respectively. Regarding ME, the two-model approach exhibits less pronounced underestimation at thresholds of 15--30 $\mathrm{mm}\cdot\mathrm{h}^{-1}$ (Fig.~\ref{fig:single_ana}b). It also yields higher POD at these thresholds, whereas the FAR values obtained with the two approaches are similar (Figs.~\ref{fig:single_ana}c and \ref{fig:single_ana}d). Regarding ETS, the single-model approach performs substantially better at the 0.1 $\mathrm{mm}\cdot\mathrm{h}^{-1}$ threshold, achieving 0.336 compared with 0.122 for the two-model approach (Fig.~\ref{fig:single_ana}e). This marked reduction indicates that estimating $p$ and $\mu$ separately substantially degrades the overall discrimination between rain and no-rain samples. Although the two-model approach yields higher ETS at thresholds of 15--30 $\mathrm{mm}\cdot\mathrm{h}^{-1}$, this improvement does not compensate for its pronounced deterioration at the 0.1 $\mathrm{mm}\cdot\mathrm{h}^{-1}$ threshold. As shown in Fig.~\ref{fig:single_ci}, the separated intervals for ME, POD, and ETS at 15, 20, and 30 $\mathrm{mm}\cdot\mathrm{h}^{-1}$ support the point-estimate result that the two-model approach reduces underestimation and improves detection skill at these thresholds. For RMSE, the point estimates consistently favor the single-model approach from 0.1 to 20 $\mathrm{mm}\cdot\mathrm{h}^{-1}$; although the intervals overlap at 20 $\mathrm{mm}\cdot\mathrm{h}^{-1}$, the overall threshold-dependent pattern still indicates an RMSE advantage for the single-model approach. At 30 $\mathrm{mm}\cdot\mathrm{h}^{-1}$, the small difference and overlapping intervals indicate no clear RMSE advantage for either approach. The overlapping FAR intervals at all three thresholds are also consistent with their comparable false-alarm performance. Overall, the single-model approach is more effective because it yields lower RMSE across most thresholds and preserves substantially better discrimination between rain and no-rain samples.

\subsection{Comparative Analysis with Baselines} \label{section:compare}
\subsubsection{Case Studies} \label{section:case_ana}

\begin{figure*}[!t]
    \centering
    \includegraphics[width=1\linewidth]{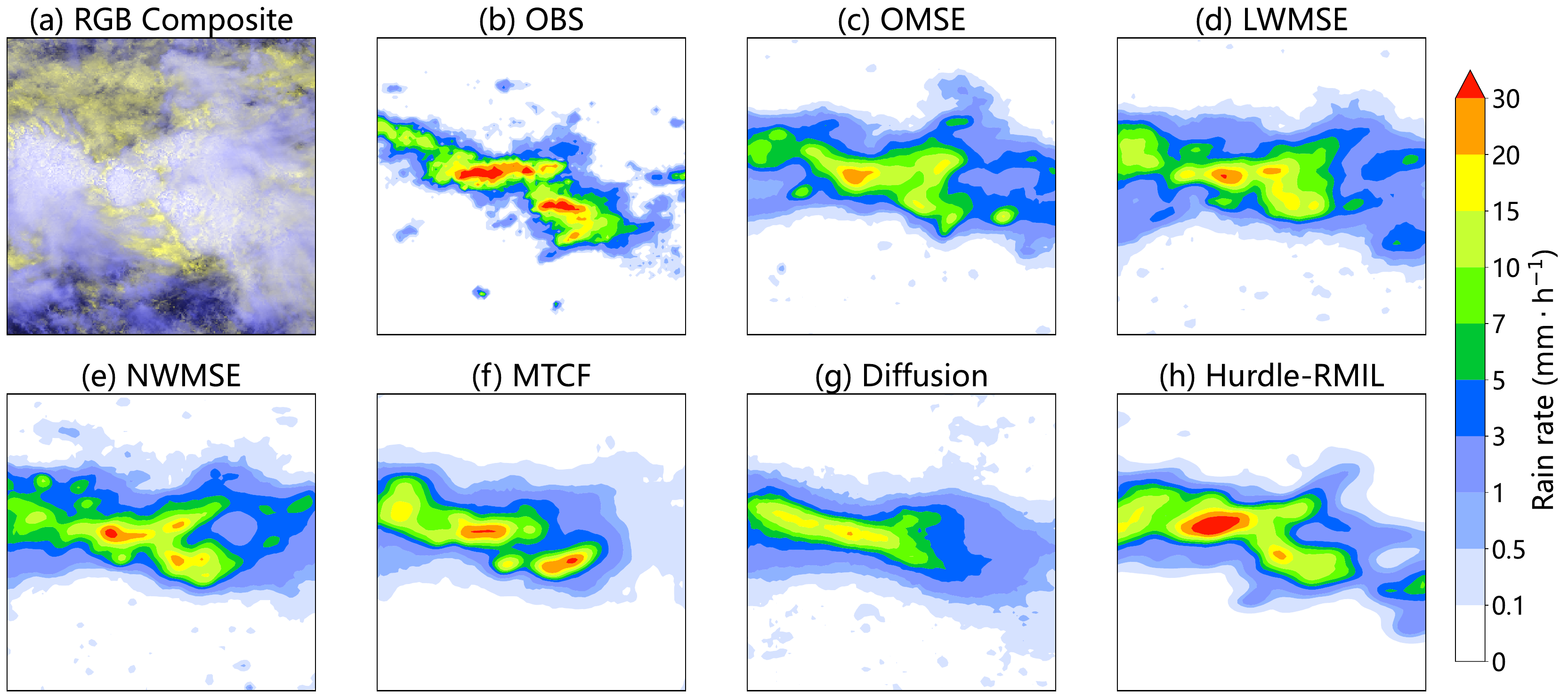}
    \caption{(a) AHI RGB composite image (red, 0.64 $\mathrm{\mu m}$; green, 0.64 $\mathrm{\mu m}$; blue, 11.2 $\mathrm{\mu m}$ reversed), (b) rain gauge measurements within the $\pm$ 0.5 h window ($\mathrm{mm}$), and (c)–(h) hourly rain rate retrievals ($\mathrm{mm} \cdot \mathrm{h}^{-1}$) from OMSE, LWMSE, NWMSE, MTCF, Diffusion, and Hurdle--RMIL for a Meiyu front rainfall case at 04:00 UTC on July 2, 2021.}
    \label{fig:case1}
\end{figure*}

This section presents two representative cases from the test set to evaluate the advantages of Hurdle--RMIL, with $\sigma$ set to 0.5. The first case occurred at 04:00 UTC on July 2, 2021 (Fig.~\ref{fig:case1}), when the Meiyu front induced continuous rain over East China. The rainband exhibited a zonal distribution with peak intensity of 48.9 $\mathrm{mm} \cdot \mathrm{h}^{-1}$. Both OMSE and Diffusion underestimated the peak rainfall intensity; in particular, Diffusion reproduced the general orientation of the rainband but failed to capture its high-intensity cores. Although LWMSE and NWMSE produced values above 30 $\mathrm{mm} \cdot \mathrm{h}^{-1}$, the spatial extent above this intensity was severely underestimated. In contrast, the OMSE, LWMSE, NWMSE, and Diffusion substantially overestimated the areal extent of rain above 1 $\mathrm{mm} \cdot \mathrm{h}^{-1}$, assigning rainfall above this threshold to many locations where the observed intensity remained below it. Overall, Hurdle--RMIL provided the most reasonable estimates, capturing both the spatial extent of the rainband and the intensity of the rain.

\begin{figure*}[!t]
    \centering
    \includegraphics[width=1\linewidth]{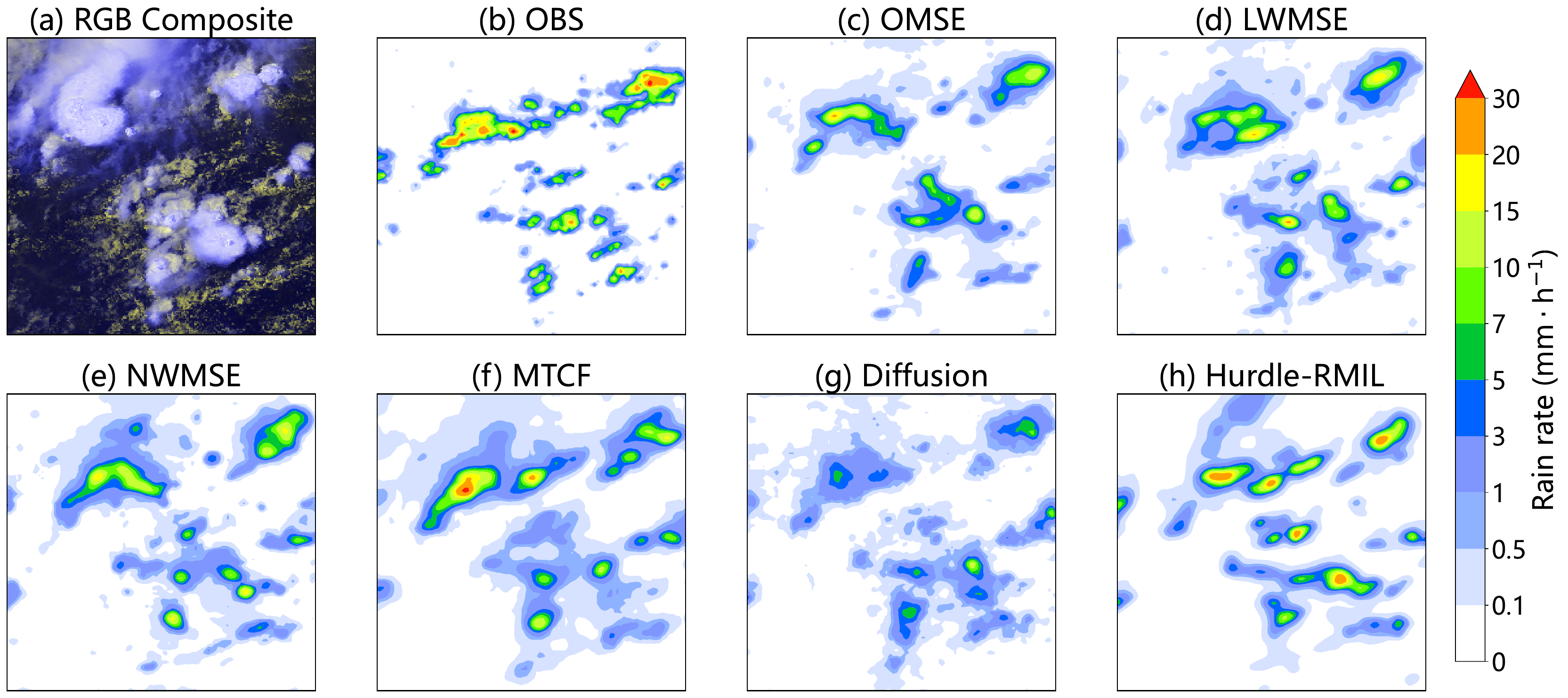}
    \caption{As in Fig.~\ref{fig:case1}, but for a convective rainfall case at 06:00 UTC on July 7, 2021.}
    \label{fig:case2}
\end{figure*}

The second case occurred at 06:00 UTC on July 7, 2021 (Fig.~\ref{fig:case2}), after the Meiyu front had dissipated. Under hot and humid conditions, convective cells developed in multiple locations and produced scattered rain with local intensity of up to 30 $\mathrm{mm} \cdot \mathrm{h}^{-1}$. In this case, LWMSE, NWMSE, MTCF, and Diffusion overestimated the spatial extent of rainfall, whereas OMSE, LWMSE, NWMSE, and Diffusion underestimated the rainfall intensity. For Diffusion, the intensity underestimation was accompanied by an overly diffuse spatial distribution, preventing the observed convective cells from being accurately reproduced. Again, Hurdle--RMIL was closest to the observations in terms of both spatial extent and intensity.

It should be noted that the current algorithm directly estimates hourly rain from a single satellite snapshot at the half-hour mark, without considering cloud motion and evolution within the intervening 30 minutes. This inevitably introduces errors in both the location and the magnitude of rain. A more accurate strategy would be to exploit the 10-min observation frequency of Himawari-8, that is, first retrieving 10-min accumulations and then summing six consecutive estimates to derive hourly totals \cite{zhugeRainfallRetrievalNowcasting2011a,yuRainfallRetrievalNowcasting2011}, which might yield results more consistent with the ground truth.

\subsubsection{Overall Performance} \label{section:stat_ana}

\begin{figure*}[!t]
    \centering
    \includegraphics[width=0.46\textwidth]{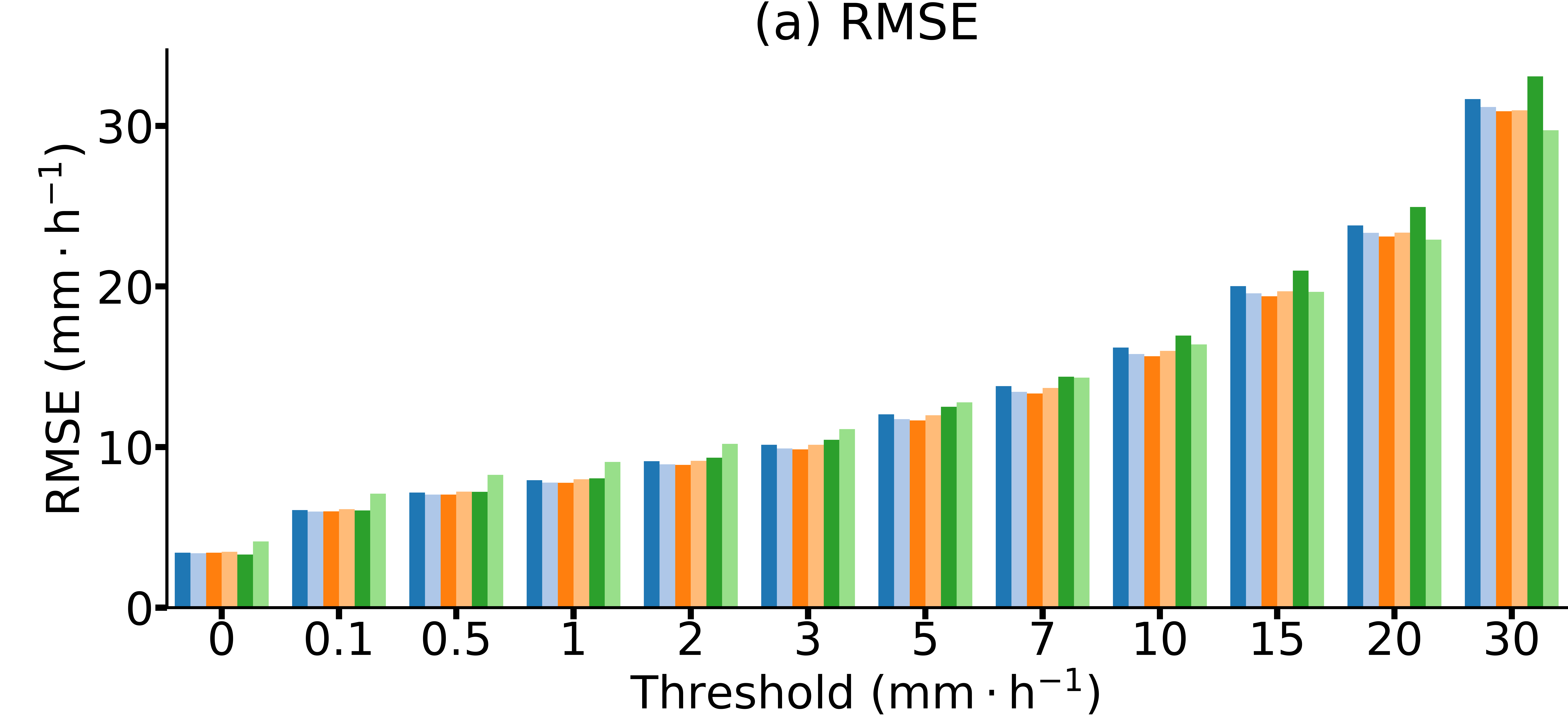}
    \includegraphics[width=0.46\textwidth]{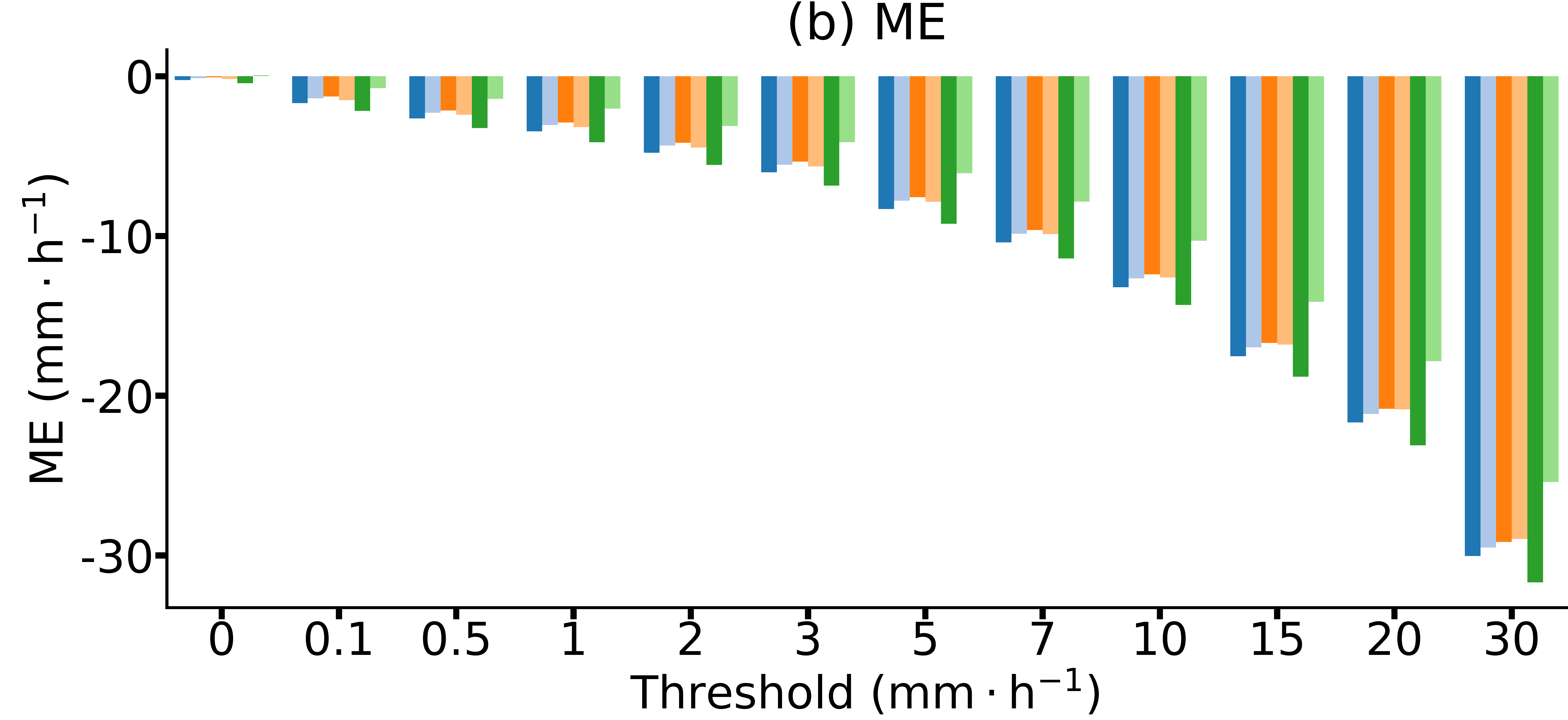}
    \par\vspace{1.5ex}
    \includegraphics[width=0.46\textwidth]{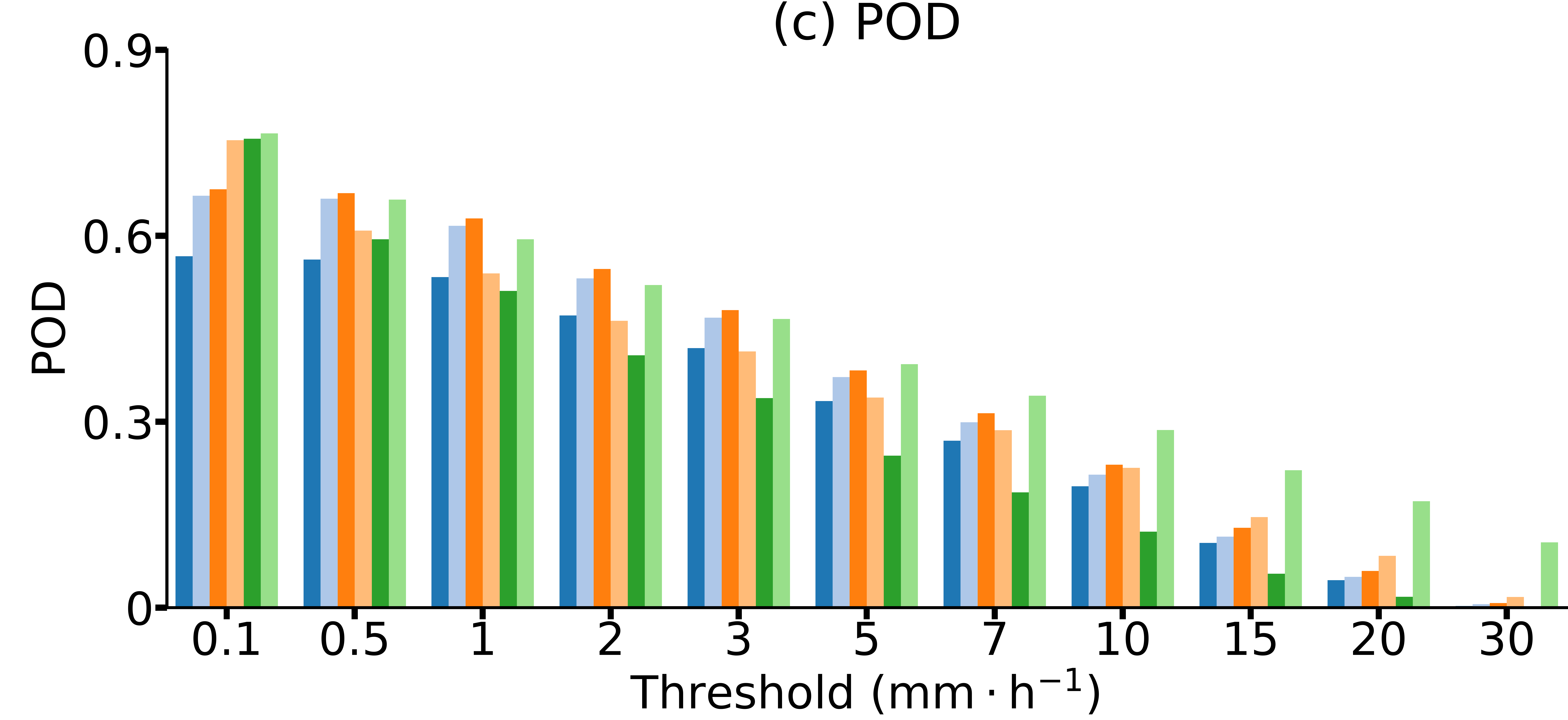}
    \includegraphics[width=0.46\textwidth]{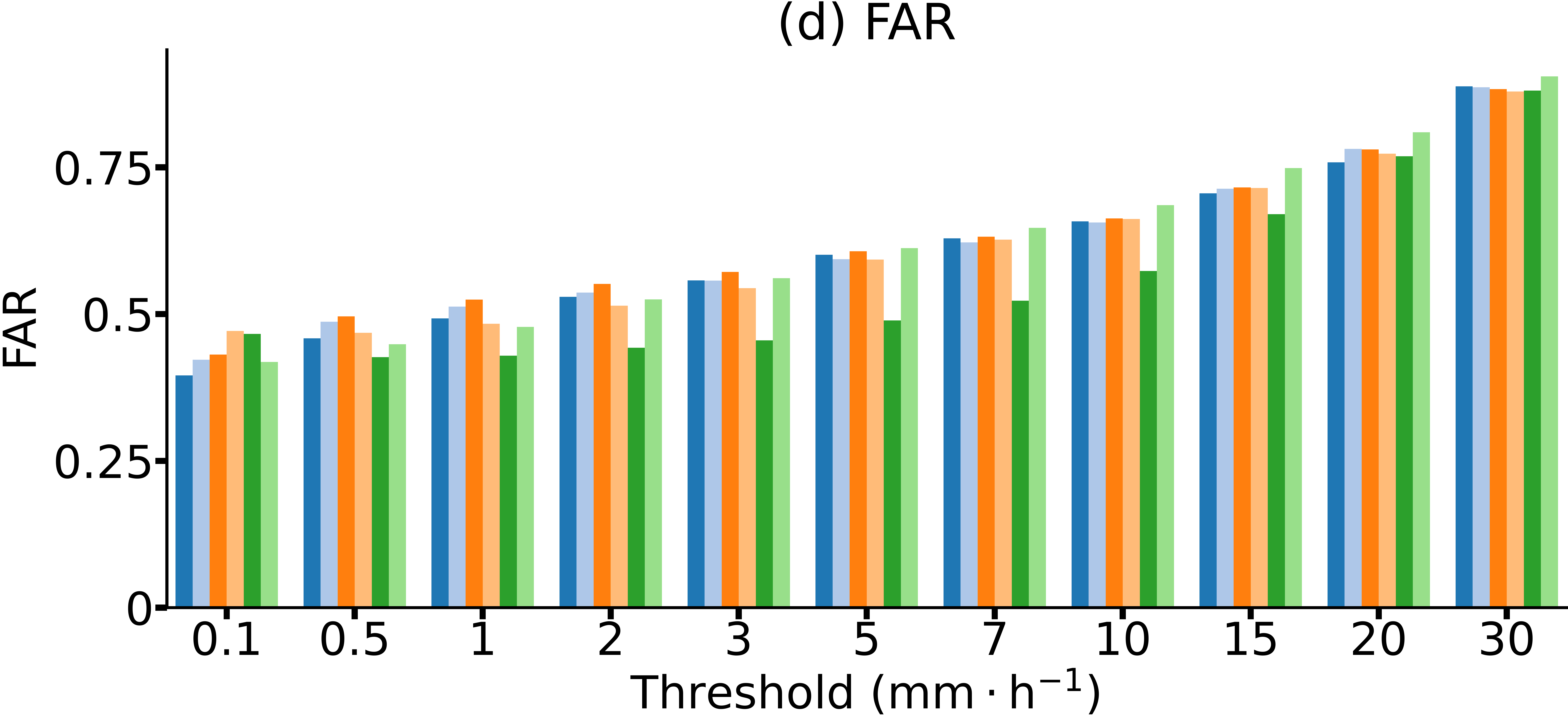}
    \par\vspace{1.5ex}
    \includegraphics[width=0.46\textwidth]{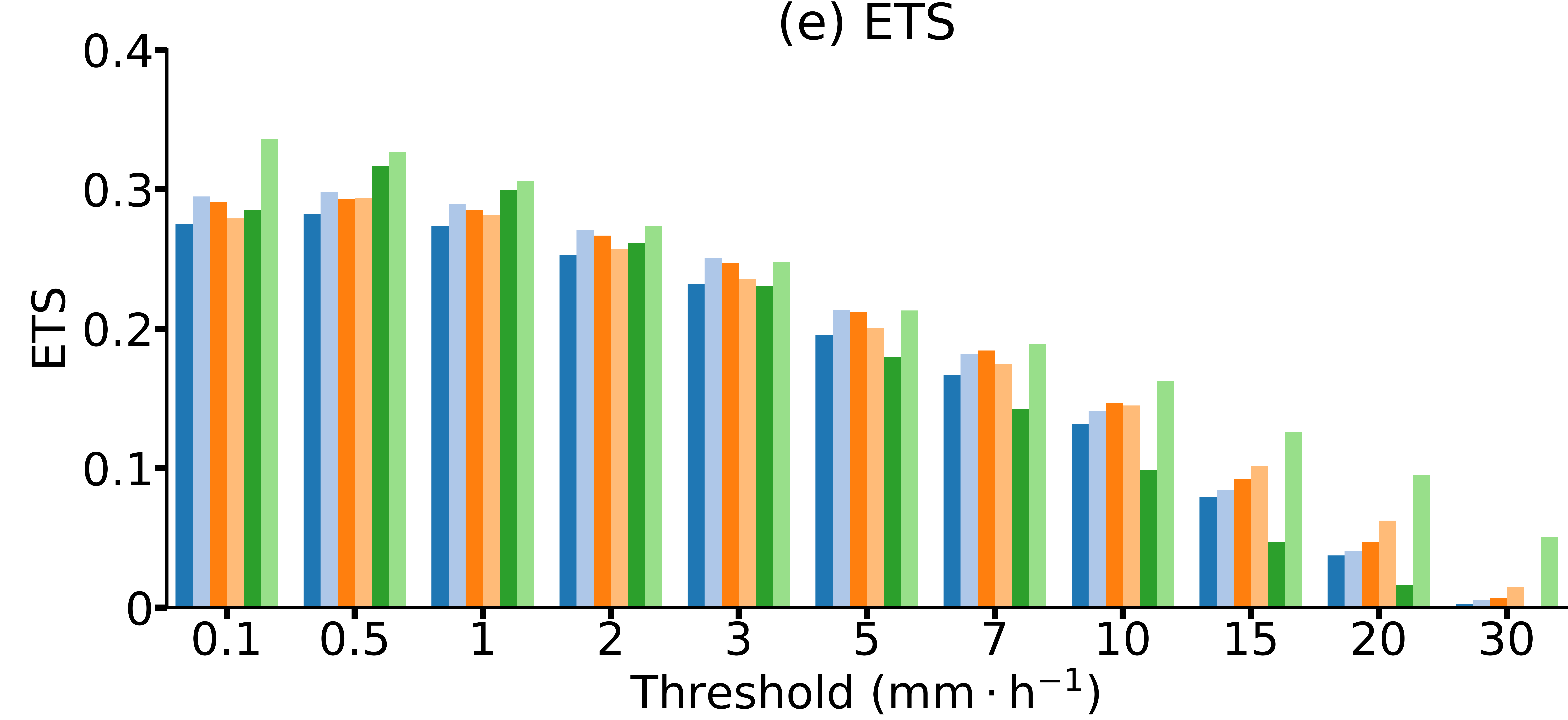}
    \includegraphics[width=0.46\textwidth]{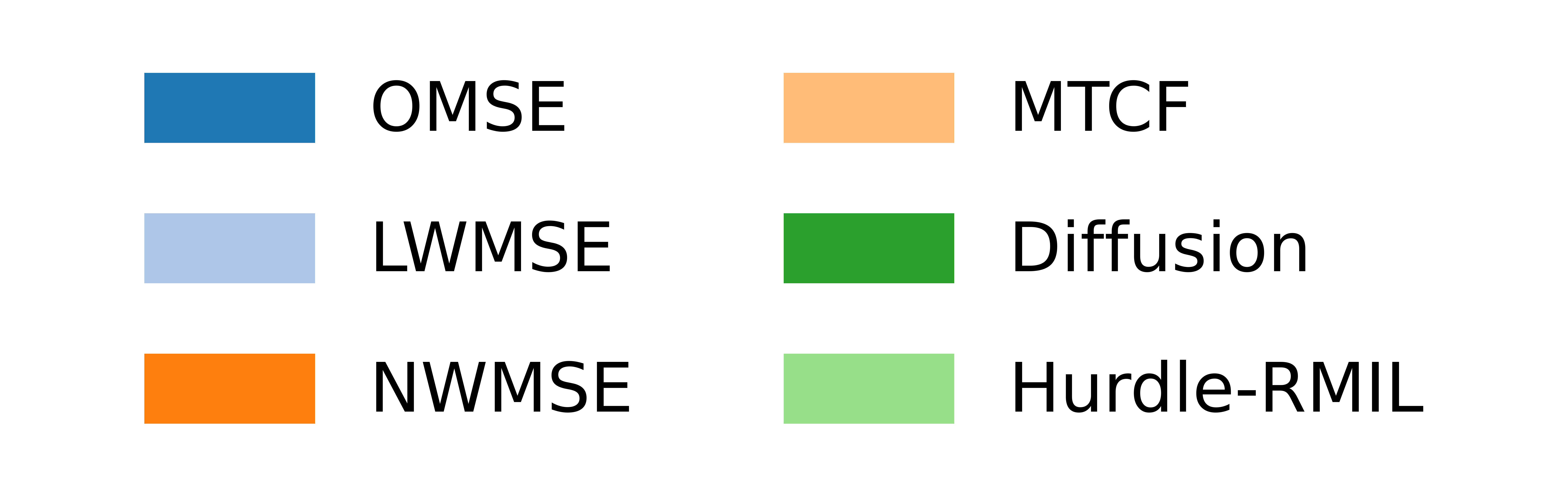}
    \caption{Performance of Hurdle--RMIL and five baselines over the pooled test set: (a) RMSE, (b) ME, (c) POD, (d) FAR, and (e) ETS.}
    \label{fig:compare}
\end{figure*}

\begin{figure*}[!t]
    \centering
    \includegraphics[width=0.46\textwidth]{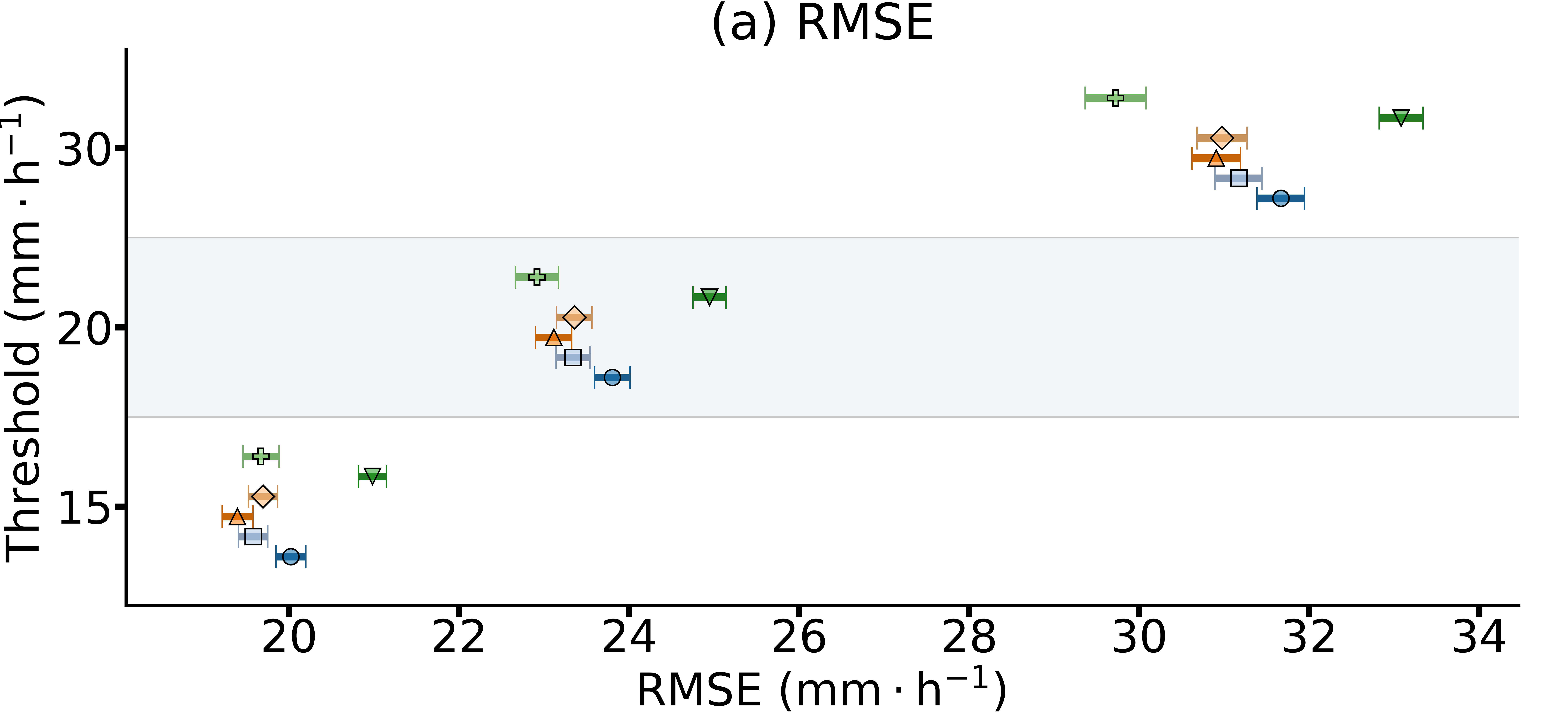}
    \includegraphics[width=0.46\textwidth]{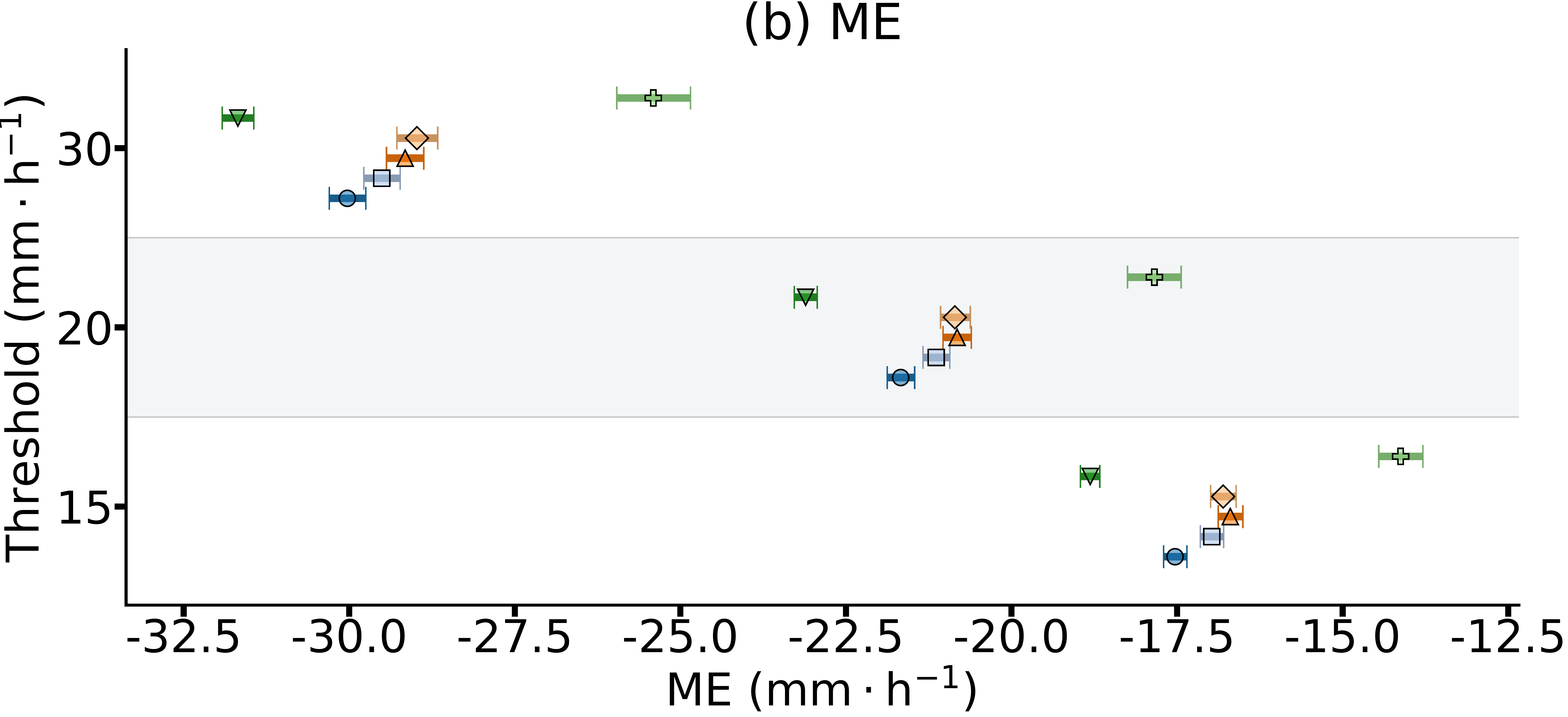}
    \par\vspace{1.5ex}
    \includegraphics[width=0.46\textwidth]{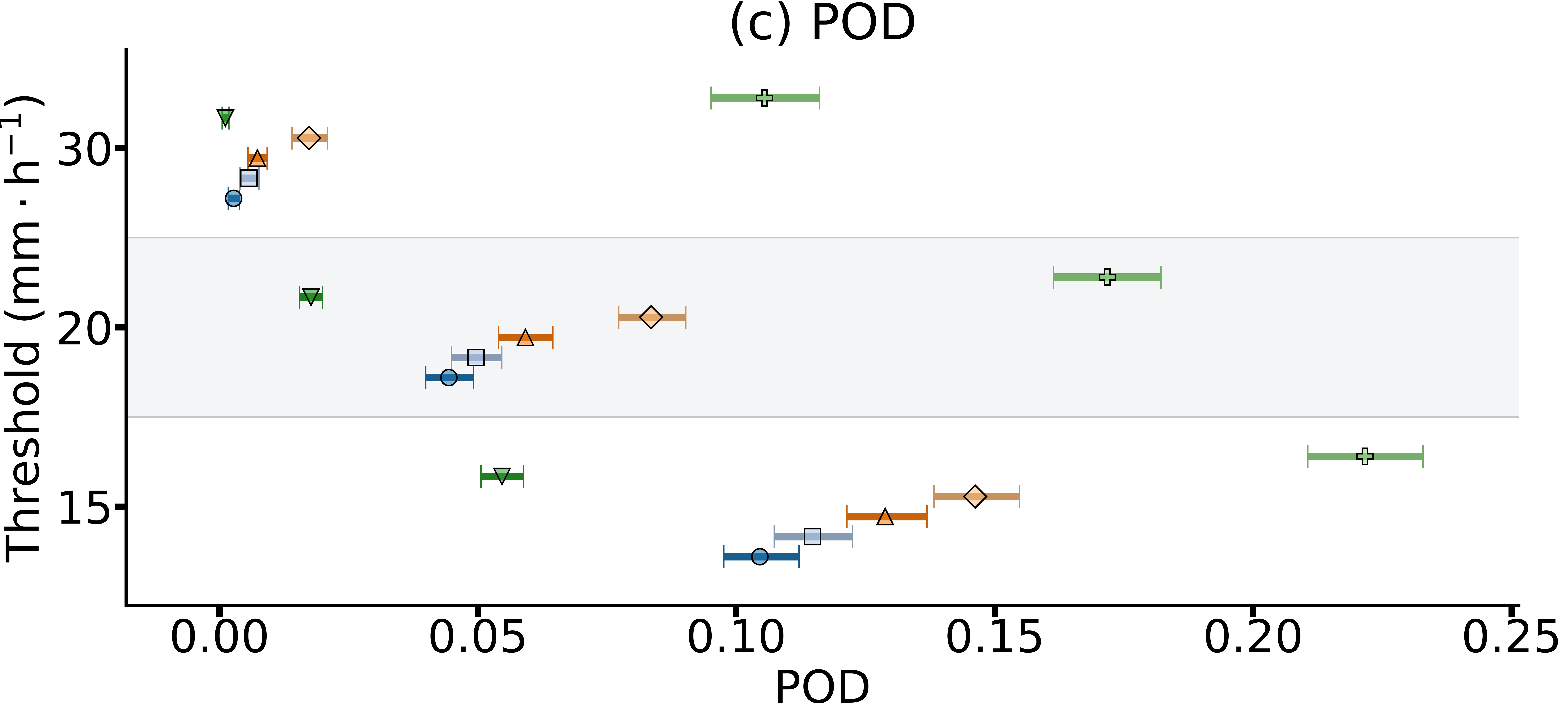}
    \includegraphics[width=0.46\textwidth]{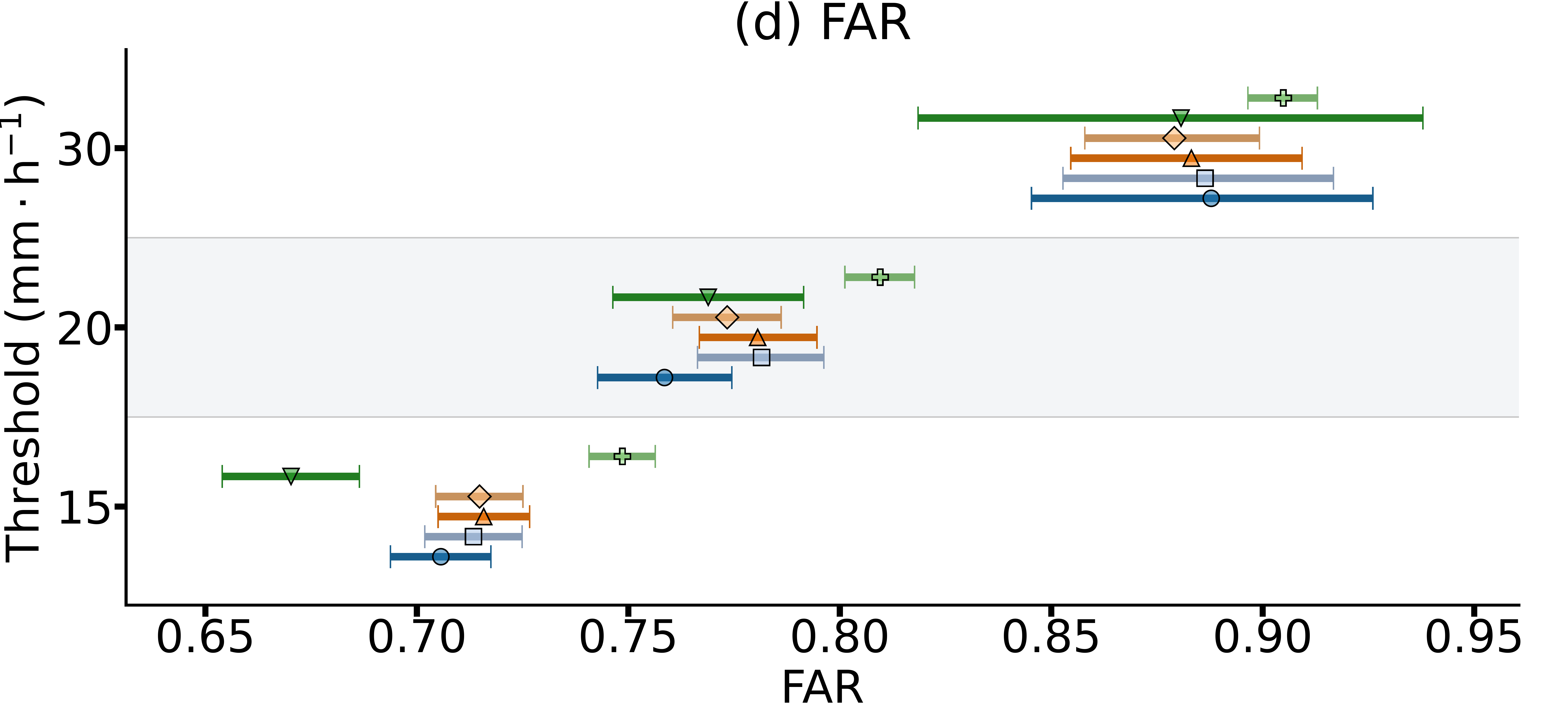}
    \par\vspace{1.5ex}
    \includegraphics[width=0.46\textwidth]{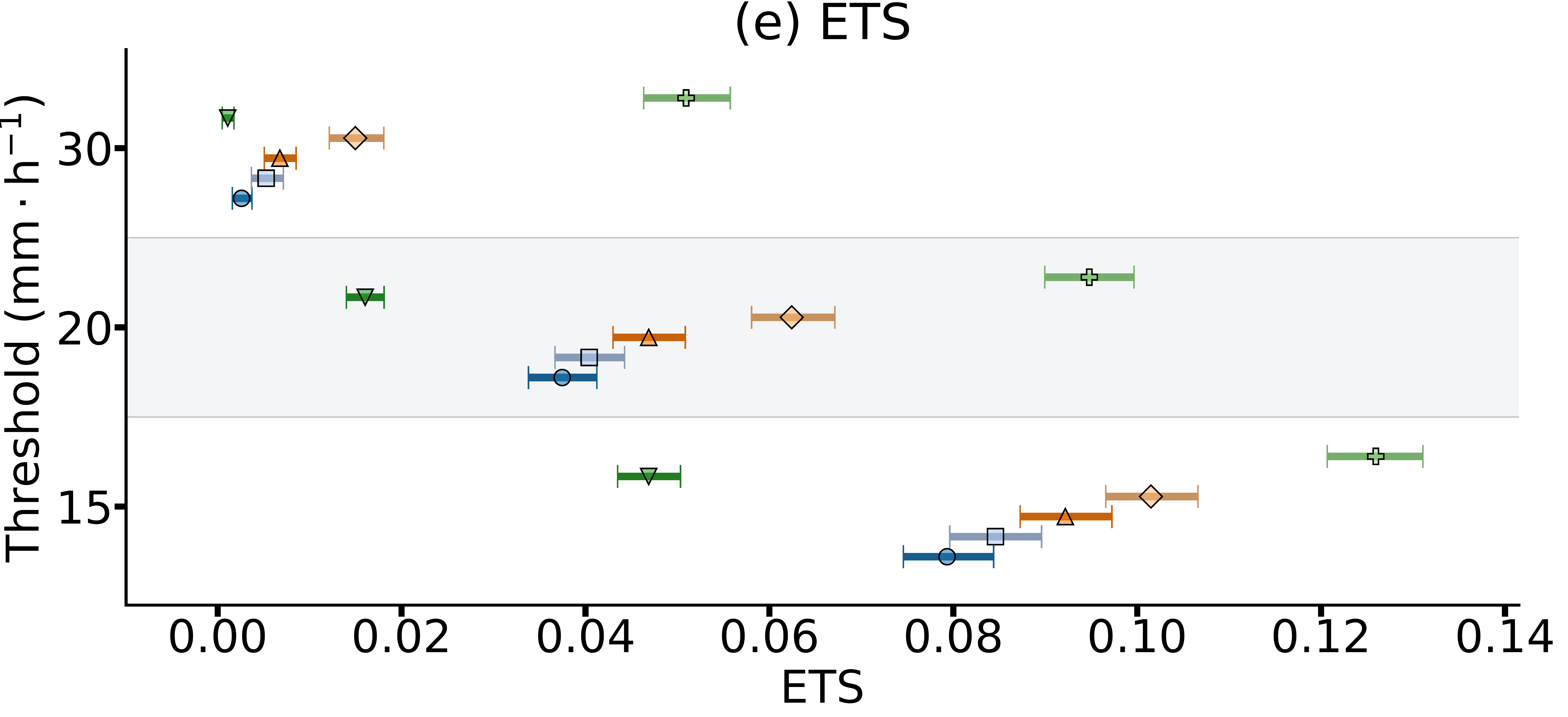}
    \includegraphics[width=0.46\textwidth]{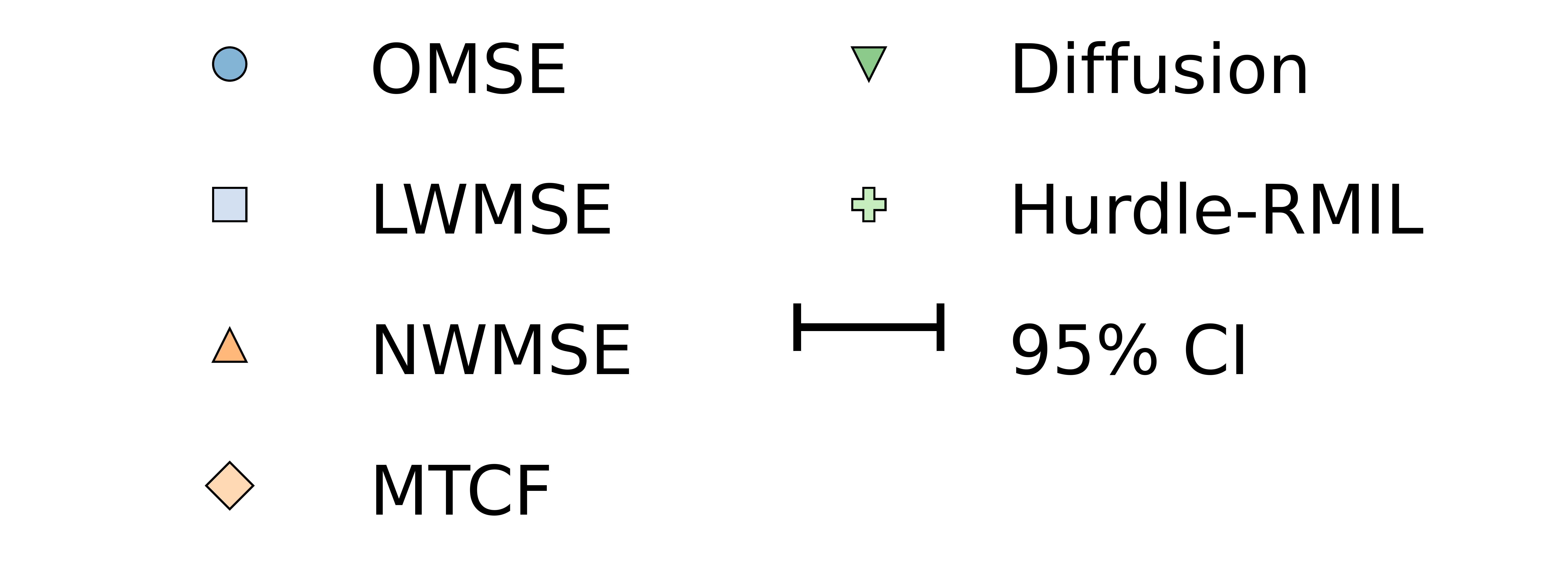}
    \caption{Performance of Hurdle--RMIL and five baselines at thresholds of 15, 20, and 30 $\mathrm{mm}\cdot\mathrm{h}^{-1}$: (a) RMSE, (b) ME, (c) POD, (d) FAR, and (e) ETS. Error bars denote 95\% paired-bootstrap confidence intervals.}
    \label{fig:compare_ci}
\end{figure*}

Based on the entire test set, a quantitative comparison between Hurdle--RMIL and the five baselines was conducted (Figs.~\ref{fig:compare} and \ref{fig:compare_ci}). As shown in Fig.~\ref{fig:compare}a, Hurdle--RMIL does not achieve the lowest RMSE at thresholds of 0.1--15 $\mathrm{mm}\cdot\mathrm{h}^{-1}$. Its relative performance improves as the threshold increases, and it yields the lowest RMSE at 20 and 30 $\mathrm{mm}\cdot\mathrm{h}^{-1}$, with values of 22.92 and 29.72 $\mathrm{mm}\cdot\mathrm{h}^{-1}$, respectively. Regarding ME, all methods exhibit negative values (Fig.~\ref{fig:compare}b), indicating increasingly pronounced underestimation as the threshold increases. Hurdle--RMIL consistently produces the smallest magnitude of ME. At the 30 $\mathrm{mm}\cdot\mathrm{h}^{-1}$ threshold, its ME is $-25.41$ $\mathrm{mm}\cdot\mathrm{h}^{-1}$, compared with $-28.98$ $\mathrm{mm}\cdot\mathrm{h}^{-1}$ for MTCF, which performs best among the baselines. Hurdle--RMIL achieves the highest POD point estimate at 0.1 $\mathrm{mm}\cdot\mathrm{h}^{-1}$ and at all evaluated thresholds from 5 to 30 $\mathrm{mm}\cdot\mathrm{h}^{-1}$ (Fig.~\ref{fig:compare}c), with its advantage over the baselines becoming pronounced at high thresholds. At thresholds of 15--30 $\mathrm{mm}\cdot\mathrm{h}^{-1}$, its FAR point estimates are also higher than those of the baselines (Fig.~\ref{fig:compare}d). For ETS, Hurdle--RMIL achieves the highest point estimate at nine of the eleven evaluated thresholds (Fig.~\ref{fig:compare}e); at 3 and 5 $\mathrm{mm}\cdot\mathrm{h}^{-1}$, LWMSE is marginally higher. Above 5 $\mathrm{mm}\cdot\mathrm{h}^{-1}$, the ETS advantage of Hurdle--RMIL becomes increasingly pronounced at higher thresholds. At 30 $\mathrm{mm}\cdot\mathrm{h}^{-1}$, Hurdle--RMIL obtains a POD of 0.105 and an ETS of 0.051, whereas the highest POD and ETS among the baselines are 0.017 and 0.015, respectively. As shown in Fig.~\ref{fig:compare_ci}, the intervals for ME, POD, and ETS remain separated from those of the baselines at 15, 20, and 30 $\mathrm{mm}\cdot\mathrm{h}^{-1}$, supporting the point-estimate result that Hurdle--RMIL reduces underestimation and improves detection skill at these thresholds. For RMSE, the point estimates favor Hurdle--RMIL at 20 and 30 $\mathrm{mm}\cdot\mathrm{h}^{-1}$; the overlapping intervals at 20 $\mathrm{mm}\cdot\mathrm{h}^{-1}$ indicate that this advantage is modest, whereas the separated intervals at 30 $\mathrm{mm}\cdot\mathrm{h}^{-1}$ provide stronger support for the reduction in RMSE. The FAR intervals indicate a clear increase at 15 and 20 $\mathrm{mm}\cdot\mathrm{h}^{-1}$, while the distinction is less evident at 30 $\mathrm{mm}\cdot\mathrm{h}^{-1}$. Overall, these results demonstrate that Hurdle--RMIL effectively alleviates rainfall underestimation and improves detection at high rainfall thresholds.


\section{Discussion}
\label{discussion}
The comparative evaluation against the baselines shows that the performance gains of Hurdle--RMIL are concentrated at high rainfall thresholds. Across the evaluated thresholds, Hurdle--RMIL produces the smallest magnitude of ME, whereas it yields the lowest RMSE only at 20 and 30 $\mathrm{mm}\cdot\mathrm{h}^{-1}$. At the lower rainfall thresholds of 0.1--10 $\mathrm{mm}\cdot\mathrm{h}^{-1}$, it does not achieve the lowest RMSE, but the differences from the best-performing baselines are small. Its improved performance at high thresholds is therefore achieved without a substantial deterioration in retrieval error for lower-intensity rainfall. The most pronounced advantage is observed in ETS: Hurdle--RMIL achieves the highest point estimate at nine of the eleven evaluated thresholds; at 3 and 5 $\mathrm{mm}\cdot\mathrm{h}^{-1}$, LWMSE is marginally higher. Above 5 $\mathrm{mm}\cdot\mathrm{h}^{-1}$, the ETS advantage of Hurdle--RMIL becomes increasingly pronounced at higher thresholds. At 30 $\mathrm{mm}\cdot\mathrm{h}^{-1}$, for example, its ETS reaches 0.051, compared with the highest baseline value of 0.015. At 15, 20, and 30 $\mathrm{mm}\cdot\mathrm{h}^{-1}$, the separated confidence intervals for ME, POD, and ETS support the reductions in underestimation and the improvements in detection indicated by the point estimates. The RMSE advantage is modest at 20 $\mathrm{mm}\cdot\mathrm{h}^{-1}$ because the confidence intervals overlap, but is more clearly supported at 30 $\mathrm{mm}\cdot\mathrm{h}^{-1}$. The case studies further show that Hurdle--RMIL better captures both rainfall intensity and spatial extent. These results are obtained from a pooled test set comprising R23 and six additional regions, to which the trained model is applied without regional retraining or fine-tuning, indicating that the improvements persist across the study domain.
These advantages stem from three pillars: the widely adopted divide-and-conquer strategy, the hurdle model, and the innovative learning method (i.e., RMIL). 
Serving as the top-level design philosophy for the framework, the divide-and-conquer strategy disentangles the complex issue of imbalanced label distribution into manageable subproblems: zero inflation and long tail, streamlining the overall task.
The hurdle model serves as the probabilistic framework through which the divide-and-conquer principle is implemented. It represents rain occurrence through the zero-rain probability mass $p(s)$ and positive rain rate through the conditional density $\mathrm{f}_{R\mid S}$, thereby addressing zero inflation.
RMIL then exploits the invariance of the forward process to derive a Bayes-based probability-density relationship between the retrieval models associated with long-tailed and balanced rainfall distributions. This relationship enables the latter to be estimated by maximum likelihood directly from naturally collected long-tailed samples. The ablation experiments show that RMIL consistently reduces the magnitude of negative ME and increases the ETS point estimates. The uncertainty analysis provides the clearest support for the RMSE reduction at 30 $\mathrm{mm}\cdot\mathrm{h}^{-1}$, while the differences in ME, POD, and ETS at 15--30 $\mathrm{mm}\cdot\mathrm{h}^{-1}$ are less affected by sampling uncertainty.

The development of RMIL draws primary inspiration from “Balanced MSE”, proposed by \cite{renBalancedMseImbalanced2022} and originally devised for computer vision tasks. Balanced MSE establishes a probability-density relationship between models associated with imbalanced and balanced label distributions. Upon this foundation, two key and novel contributions are introduced. First, whereas Ren et al.~\cite{renBalancedMseImbalanced2022} regarded the invariance of the visual label-to-image mapping (analogous to the forward model in QRS) as a hypothetical assumption, RMIL relates this invariance to the rainfall-to-satellite forward process governed by radiative-transfer physics under fixed observation conditions. This provides a physical basis for adapting the transformation to infrared rainfall retrieval. Second, under the empirical lognormal formulation adopted for the marginal and conditional rain-rate densities, the corresponding negative log-likelihood admits a closed-form expression. The closed-form expression avoids numerical evaluation of the normalization integral during training and provides a tractable optimization objective.

The parameter-estimation experiment in Section~\ref{section:single_model_ana} further shows that the two implementation schemes exhibit distinct threshold-dependent behavior. Sequential estimation using two models reduces underestimation and yields higher POD and ETS at 15--30 $\mathrm{mm}\cdot\mathrm{h}^{-1}$, whereas joint estimation using a single model produces lower RMSE over most thresholds and preserves substantially better discrimination between rain and no-rain samples. In particular, separating the estimation of $p$ and $\mu$ reduces ETS from 0.336 to 0.122 at the 0.1 $\mathrm{mm}\cdot\mathrm{h}^{-1}$ threshold, indicating pronounced degradation in overall discrimination. The single-model approach is therefore more effective overall, although its advantage is not uniform across every metric at high rainfall thresholds. Because both components are estimated from shared features in the same network, information relevant to rain occurrence can also contribute to positive-rain-rate estimation, rather than being isolated in two independently trained models.

Several limitations merit attention. First, although the RMIL derivation is not restricted to a specific probability distribution, its implementation in this study adopts the lognormal distribution as an empirical choice. Performance may be constrained when the target variable or its conditional distribution deviates substantially from lognormality. Second, the current RMIL formulation cannot estimate $\sigma$ dynamically and differentially; $\sigma$ is therefore treated as a tunable hyperparameter. The sensitivity analysis shows that its effects differ across metrics and thresholds, and the overlapping RMSE and ETS intervals for several settings between 0.4 and 0.6 indicate uncertainty in their relative ordering. The value of 0.5 used here should therefore be regarded as an empirical setting for the present experiment rather than a universally optimal choice. Third, although testing across six regions outside R23 broadens the geographic evaluation, all data are from May to August, the model is trained in a single region, and only Himawari-8/AHI observations are used. Further evaluation across additional seasons, climatic regimes, training regions, and satellite sensors is required to establish broader robustness.

\section{Conclusion}
\label{conclusion}

Imbalanced label distributions constrain AI-based QRS by degrading retrieval performance for rare samples. Infrared rainfall retrieval exemplifies this problem because of its highly imbalanced label distribution. To address this problem, this study proposes Hurdle--RMIL. Following the divide-and-conquer principle, the framework decomposes rainfall imbalance into zero inflation and the long-tailed distribution of positive rain. The hurdle model serves as the probabilistic framework through which this principle is implemented: it represents the conditional rain-rate distribution as a probability mass at zero and a conditional density over positive values, thereby addressing zero inflation. For the long-tailed positive-rain distribution, the newly proposed RMIL exploits the invariance of the causal rainfall-to-satellite forward process under fixed observation conditions to derive a Bayes-based probability-density transformation that relates the retrieval models associated with long-tailed and balanced rainfall distributions. This transformation enables the retrieval model associated with a balanced rainfall distribution to be estimated by maximum likelihood directly from naturally collected long-tailed samples, without constructing a balanced training dataset. Under the empirical lognormal formulation adopted in this study, the corresponding negative log-likelihood admits a closed-form expression for AI model training.

Hurdle--RMIL was evaluated using multiregional test data spanning a broad area of China from 106.25$^{\circ}$--120.65$^{\circ}$E and 23.3$^{\circ}$--37.7$^{\circ}$N, without region-specific retraining or fine-tuning. The results show that Hurdle--RMIL mitigates systematic underestimation and substantially improves detection at high rainfall thresholds, without markedly increasing retrieval errors at lower rainfall thresholds. At thresholds of 0.1--10 $\mathrm{mm}\cdot\mathrm{h}^{-1}$, its RMSE differs only slightly from those of the best-performing baselines, while it achieves the highest ETS point estimate at nine of the eleven evaluated thresholds, with its advantage becoming more pronounced at high thresholds. At 30 $\mathrm{mm}\cdot\mathrm{h}^{-1}$, its ETS reaches 0.051, compared with the highest baseline value of 0.015; its ME is $-25.41$ $\mathrm{mm}\cdot\mathrm{h}^{-1}$, compared with $-28.98$ $\mathrm{mm}\cdot\mathrm{h}^{-1}$ for the best-performing baseline. The paired-bootstrap confidence intervals support the improvements in ME, POD, and ETS at 15, 20, and 30 $\mathrm{mm}\cdot\mathrm{h}^{-1}$. Together with the improved representation of rainfall intensity and spatial extent in the case studies, these findings demonstrate the effectiveness of Hurdle--RMIL in retrieving rare high-intensity rainfall across the study domain.
%

%

Hurdle--RMIL constitutes an effective framework for addressing imbalanced rainfall distributions. Because the RMIL derivation is not specific to infrared rainfall retrieval, it has the theoretical potential to be extended to other QRS tasks characterized by imbalanced target distributions, although this transferability remains to be evaluated. Future research could explore three key directions. First, integrating RMIL with alternative distributions, such as Gamma or Weibull, would reduce its reliance on a specific distributional assumption and enhance its flexibility. Second, further theoretical analysis is needed to understand the inability of the current RMIL formulation to estimate the shape parameter dynamically and differentially, thereby guiding the development of a more robust variant. Third, evaluation across additional seasons, climatic regimes, training regions, and satellite sensors, together with applications to other QRS tasks characterized by imbalanced target distributions, is needed to assess the robustness and transferability of RMIL.

\appendices
\section{Selection of MTCF's hyperparameter}
\label{app1}

\begin{figure*}[!t]
    \centering
    \includegraphics[width=0.46\textwidth]{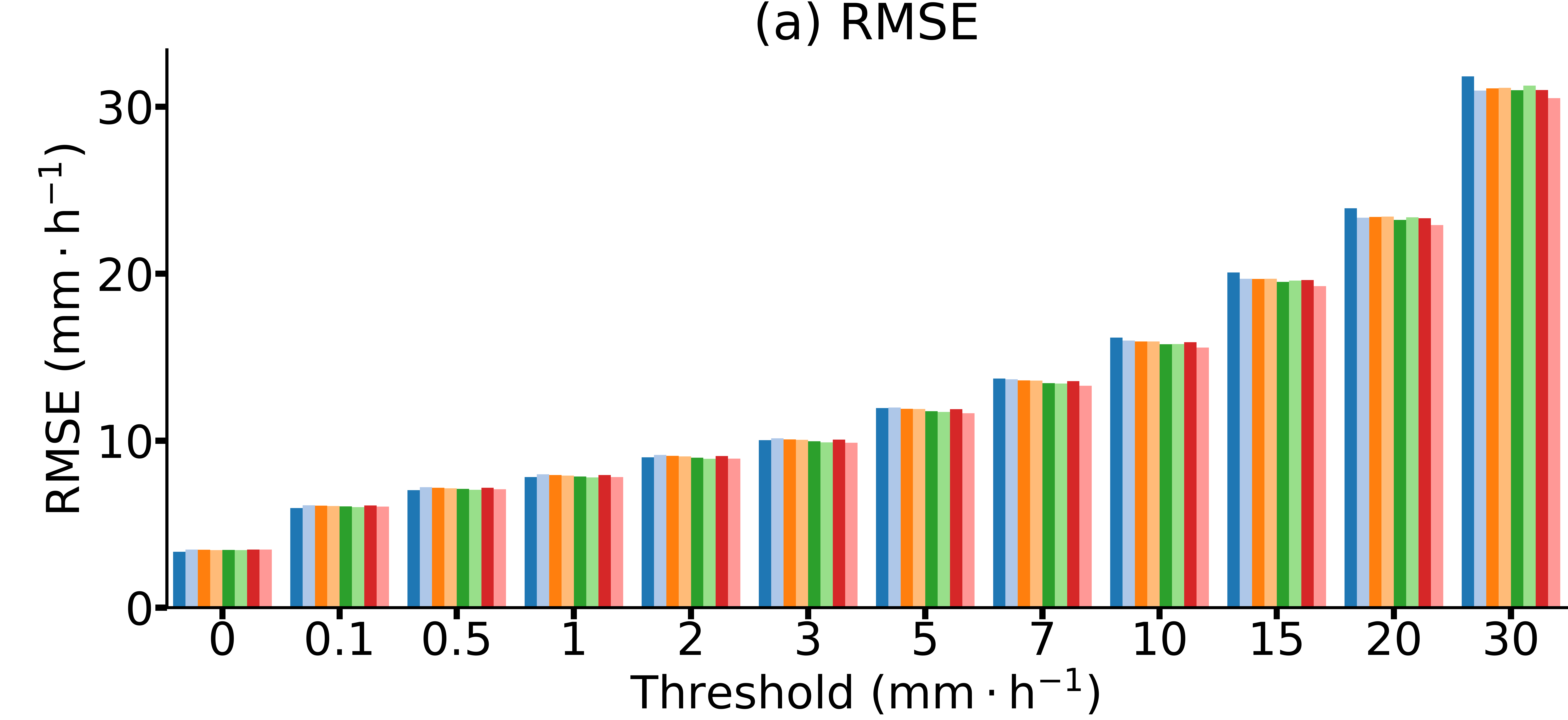}
    \includegraphics[width=0.46\textwidth]{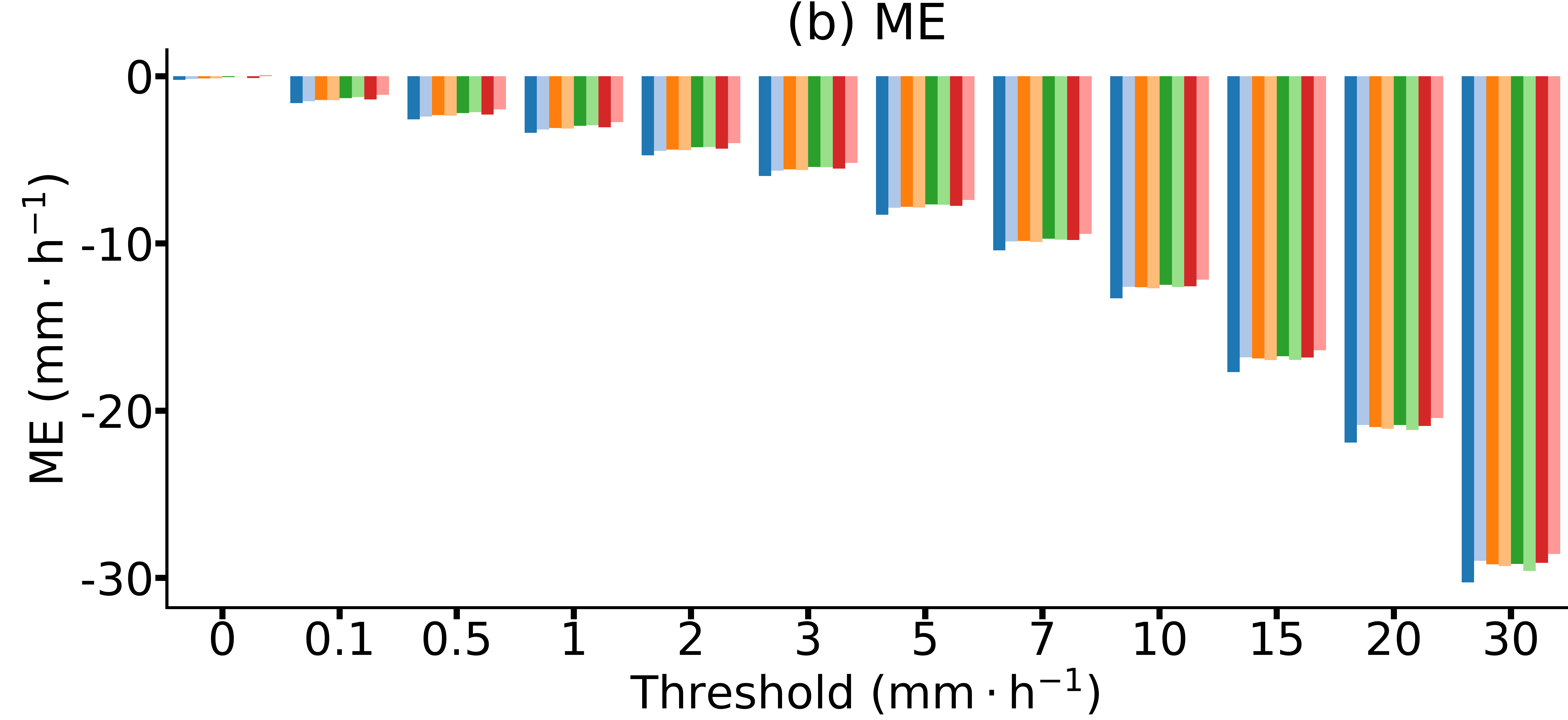}
    \par\vspace{1.5ex}
    \includegraphics[width=0.46\textwidth]{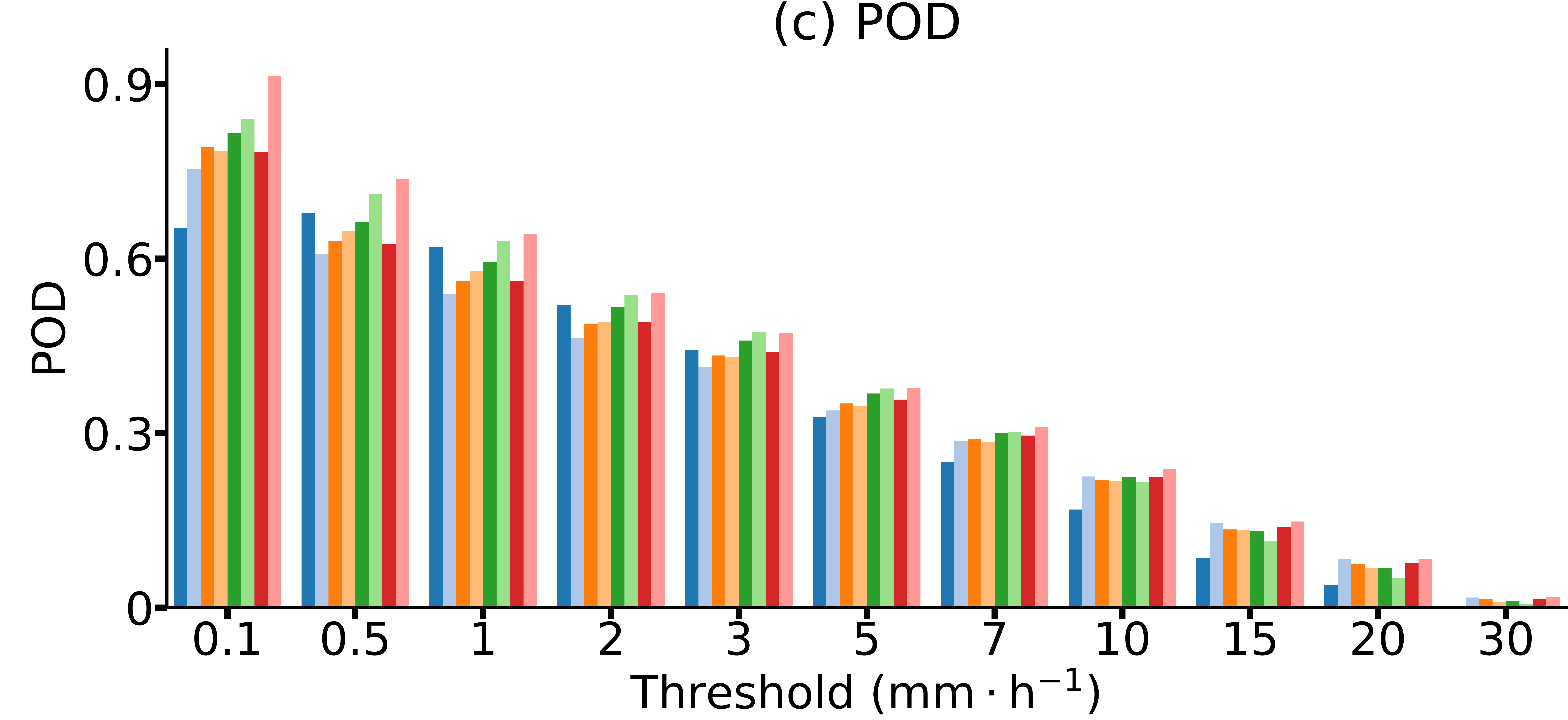}
    \includegraphics[width=0.46\textwidth]{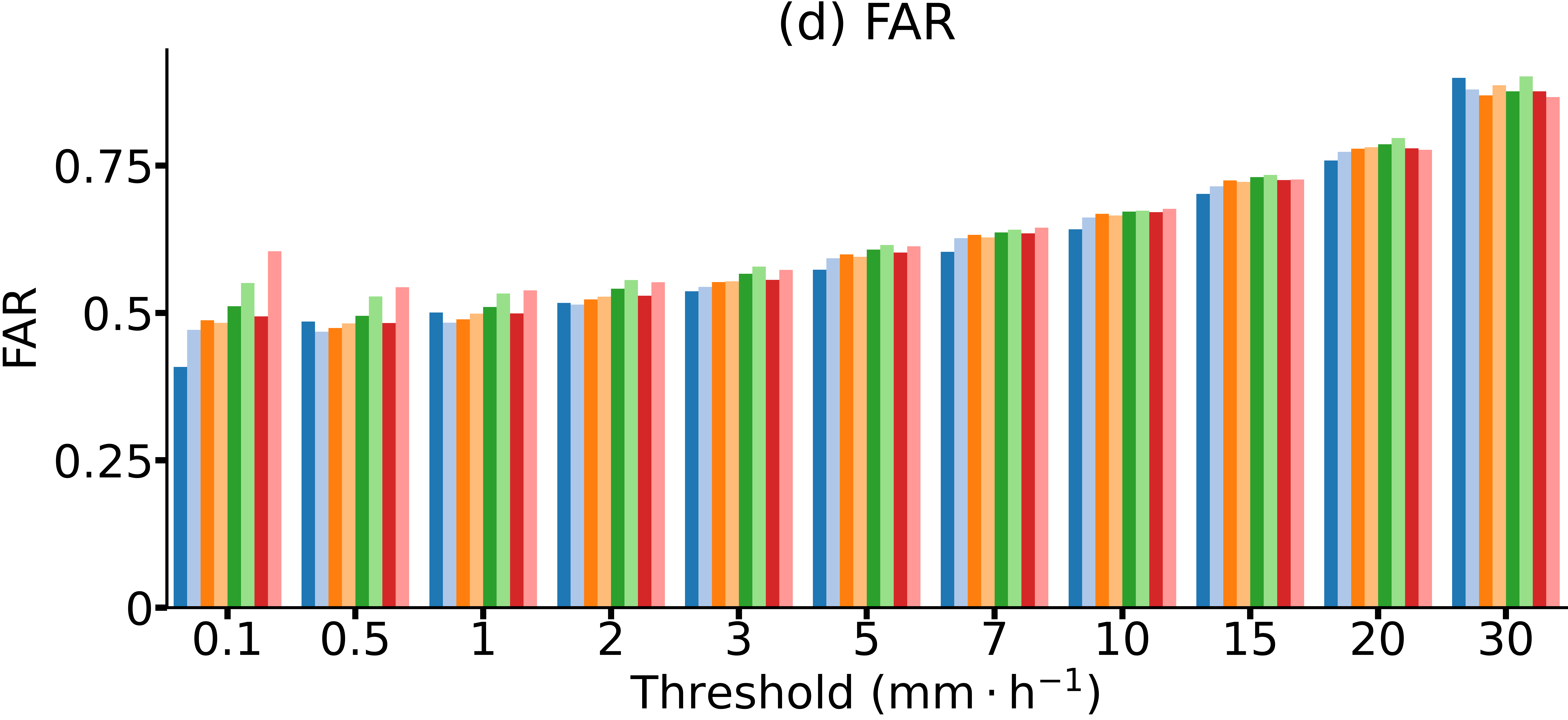}
    \par\vspace{1.5ex}
    \includegraphics[width=0.46\textwidth]{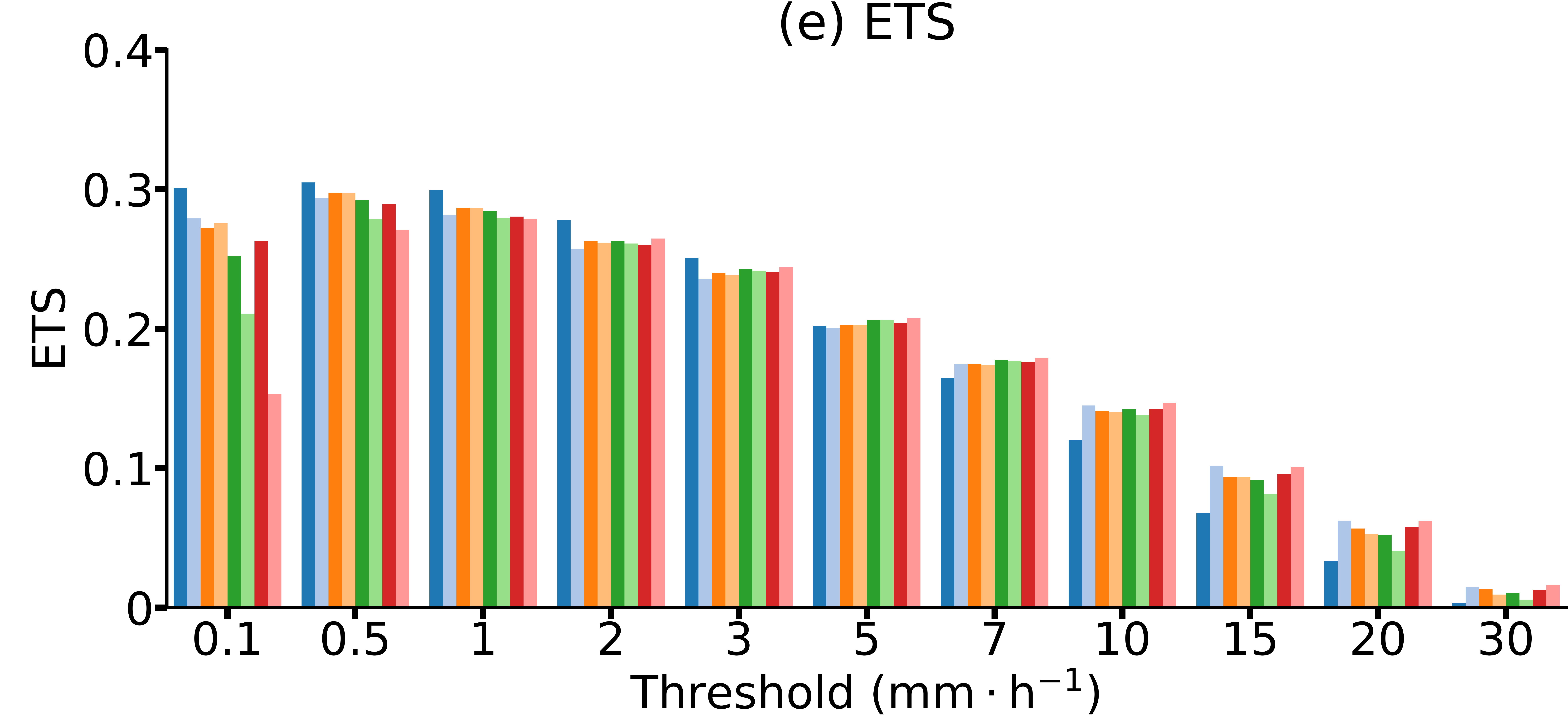}
    \includegraphics[width=0.46\textwidth]{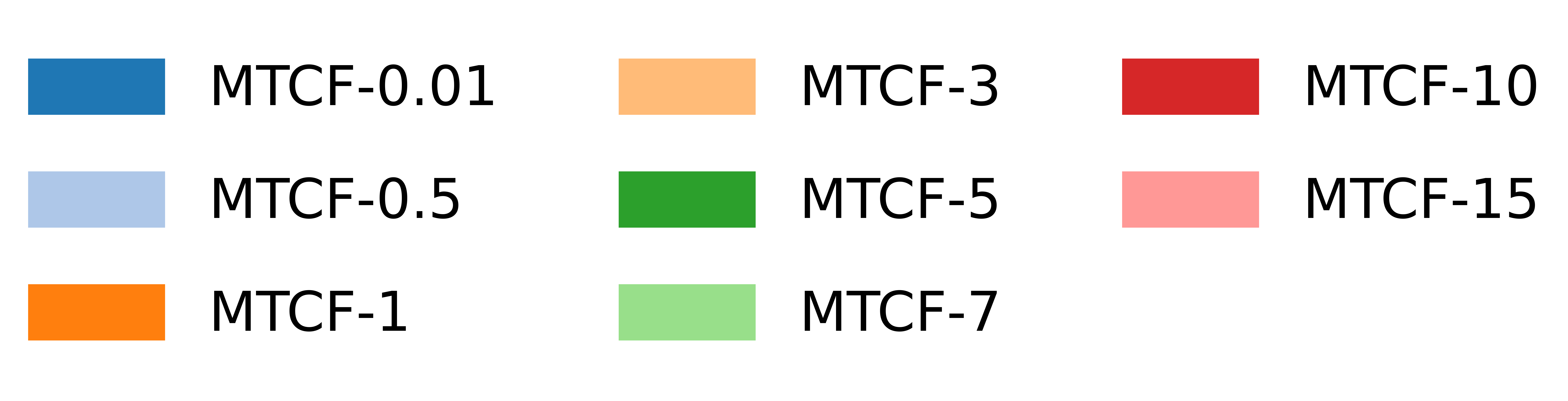}
    \caption{Performance of MTCF with different decision thresholds over the pooled test set: (a) RMSE, (b) ME, (c) POD, (d) FAR, and (e) ETS.}
    \label{fig:mtcf_ana}
\end{figure*}

\begin{figure*}[!t]
    \centering
    \includegraphics[width=0.46\textwidth]{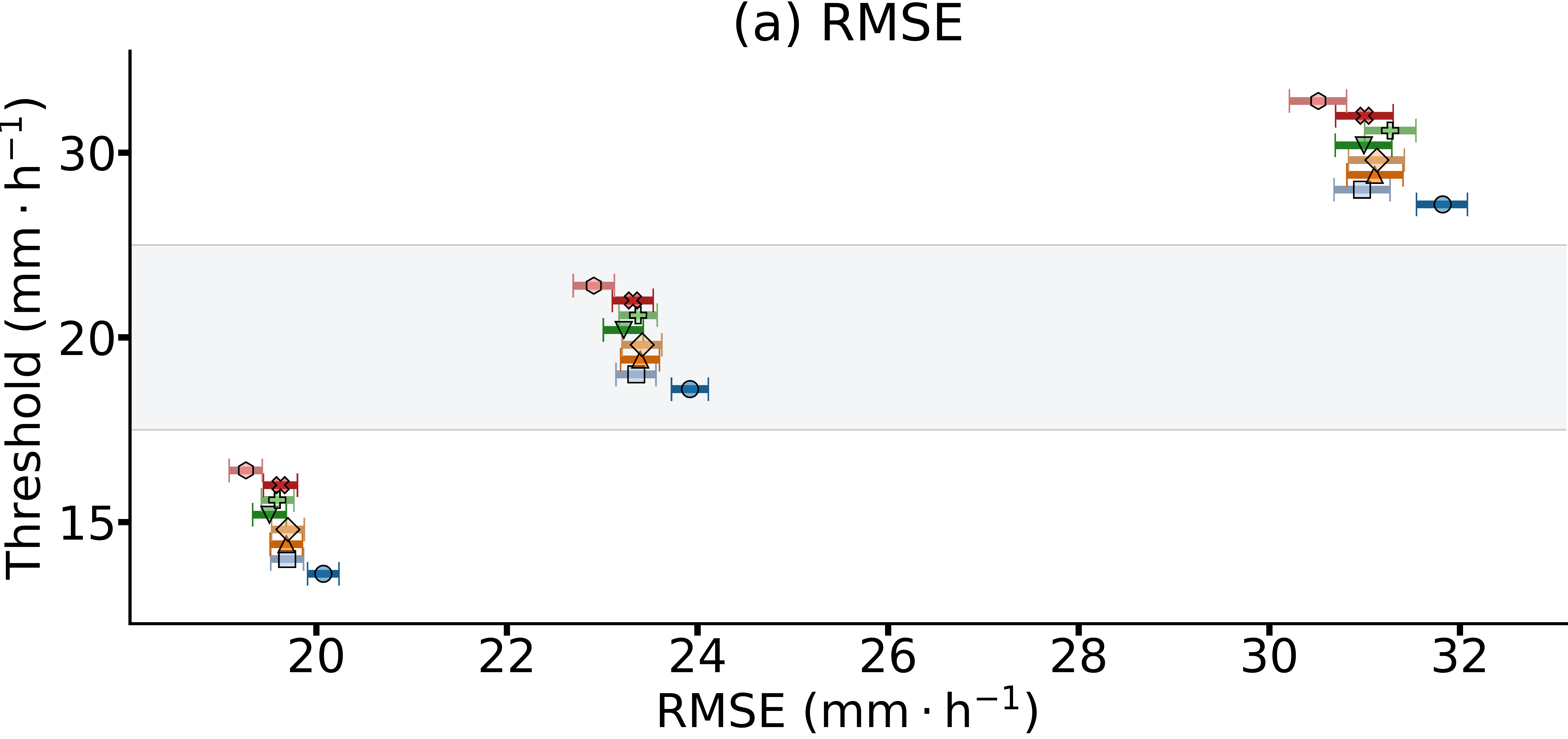}
    \includegraphics[width=0.46\textwidth]{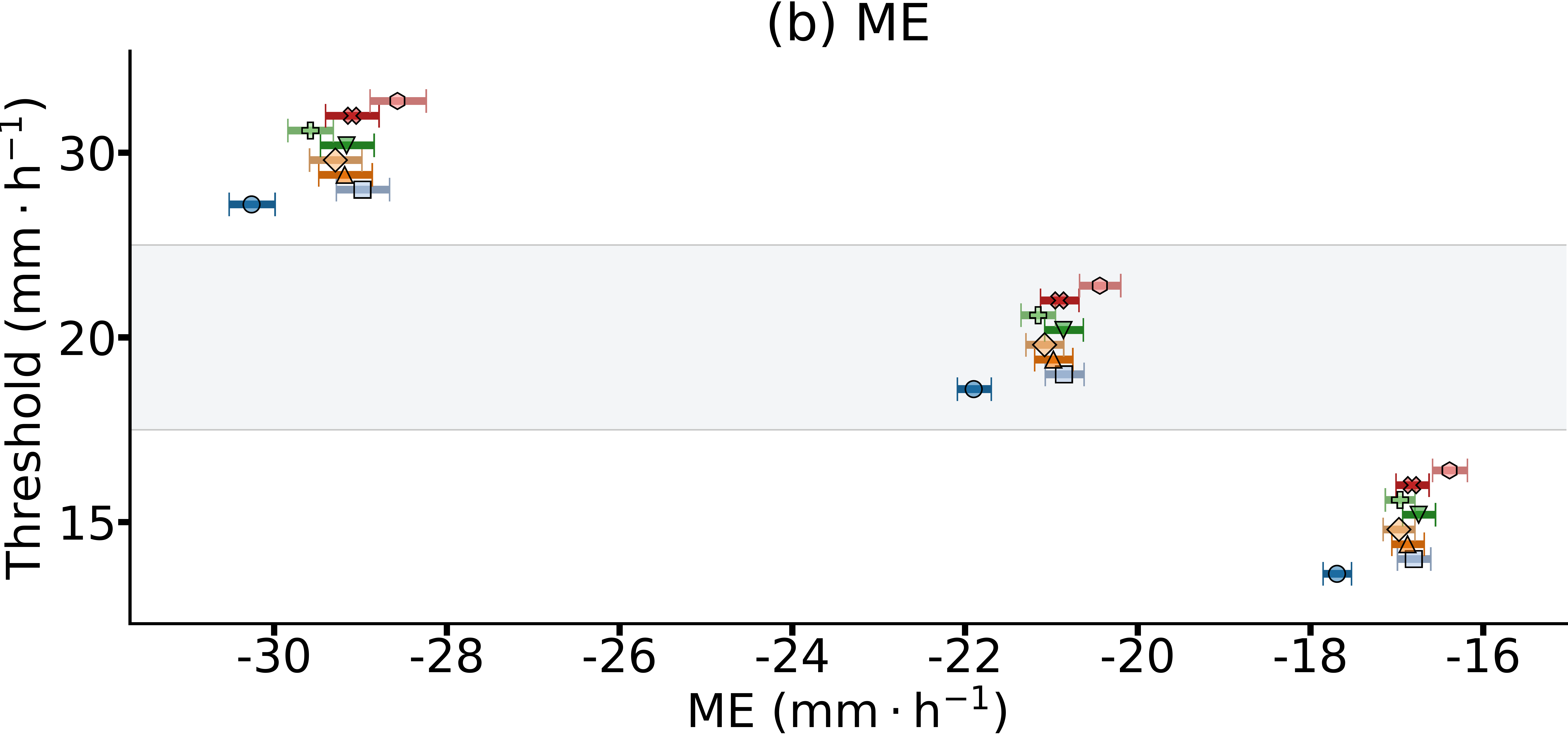}
    \par\vspace{1.5ex}
    \includegraphics[width=0.46\textwidth]{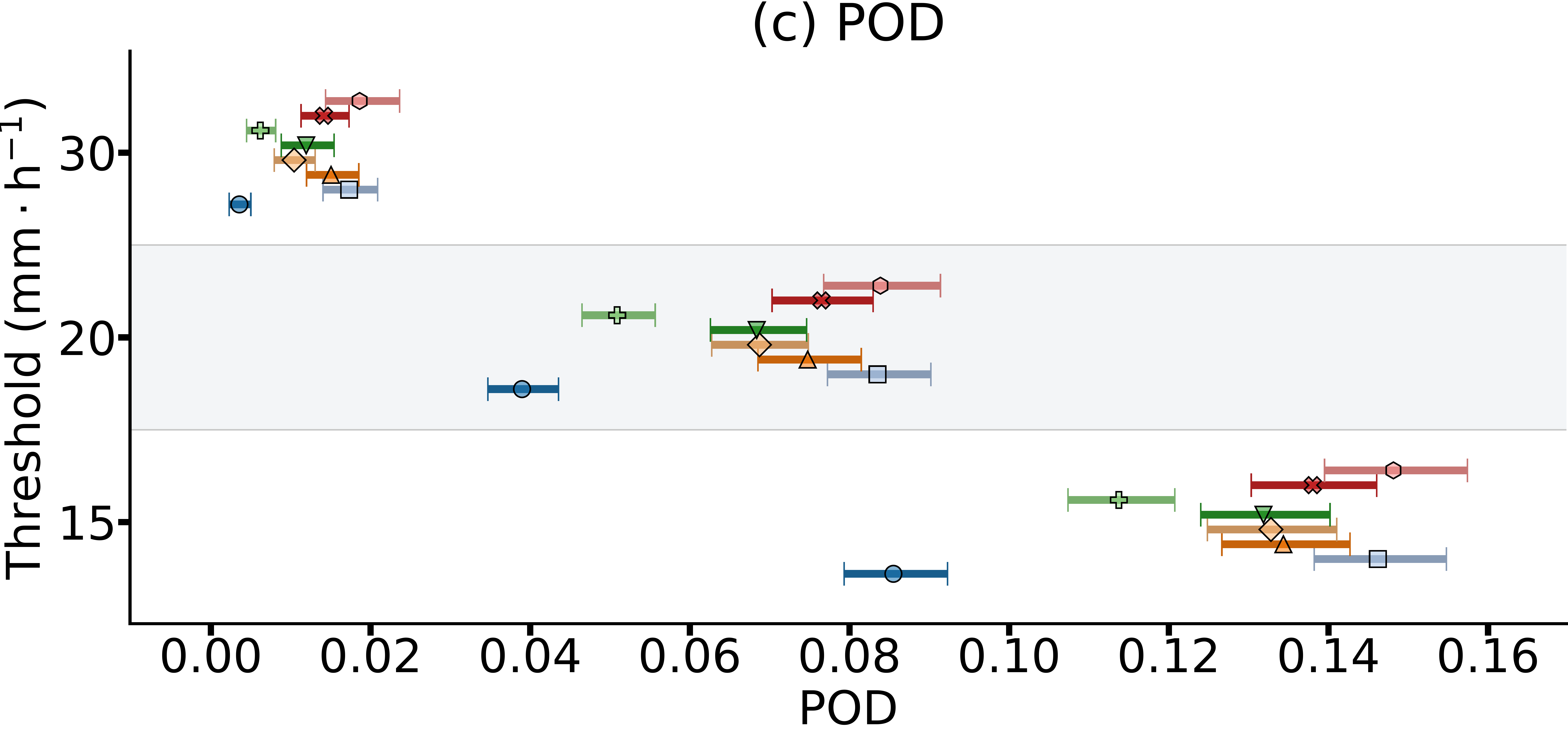}
    \includegraphics[width=0.46\textwidth]{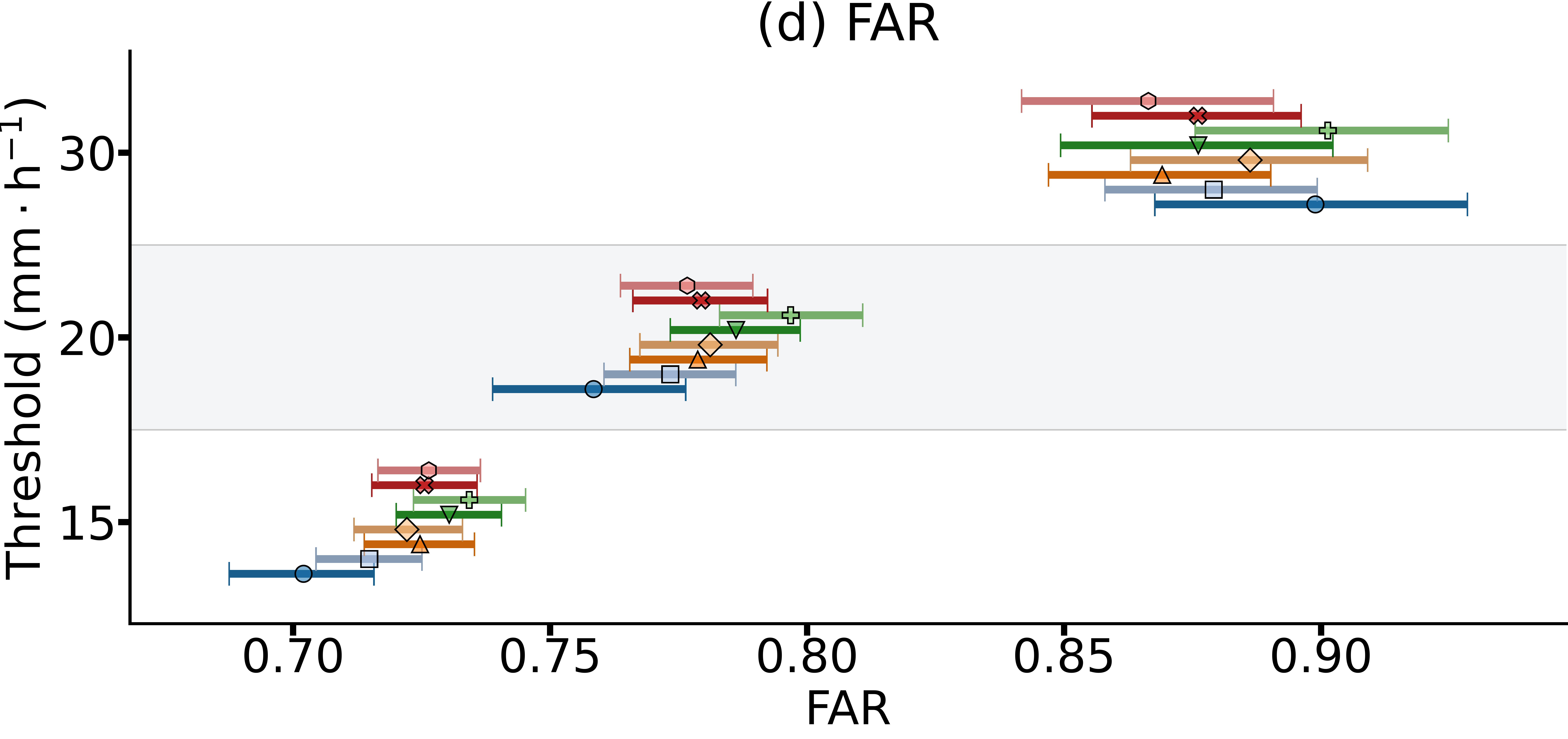}
    \par\vspace{1.5ex}
    \includegraphics[width=0.46\textwidth]{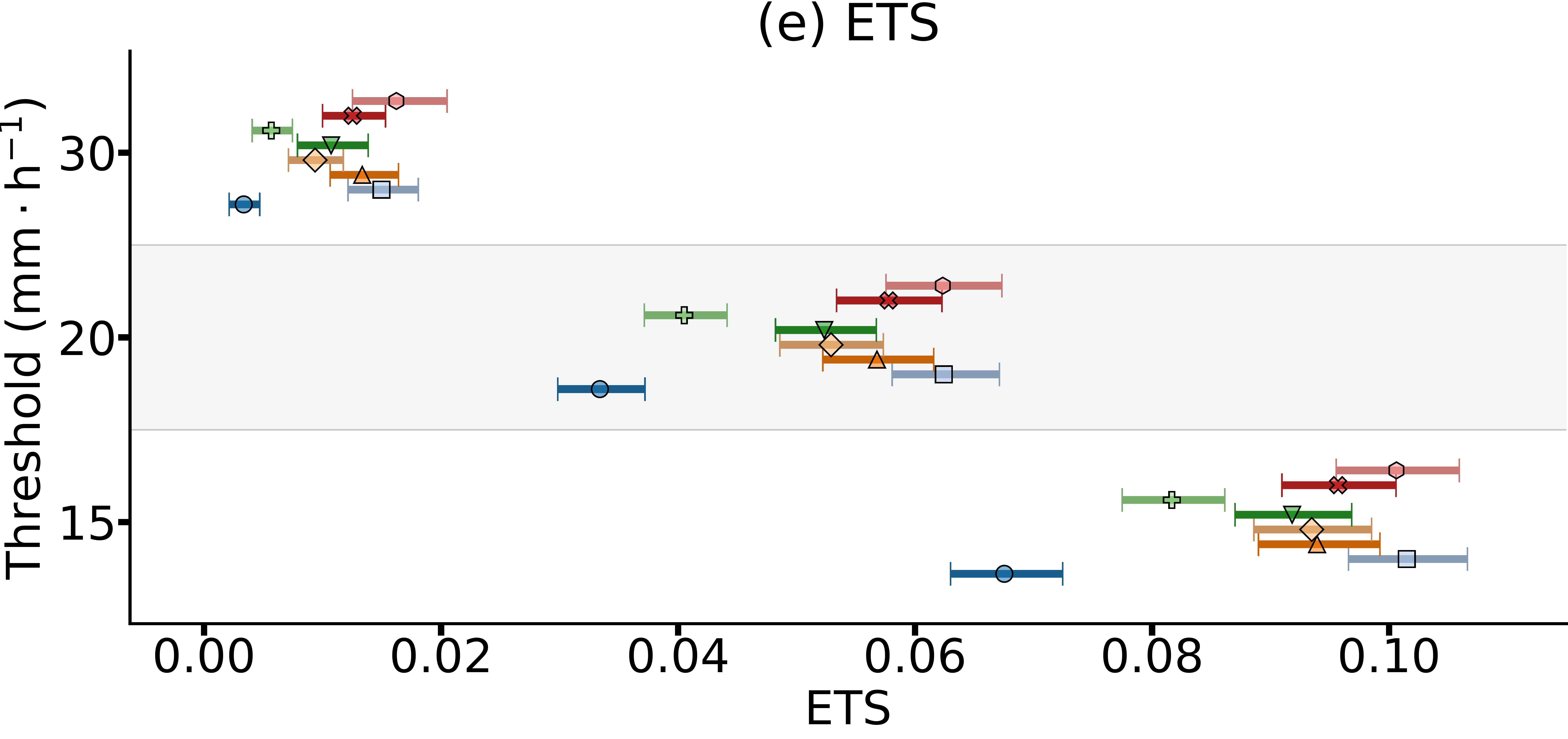}
    \includegraphics[width=0.46\textwidth]{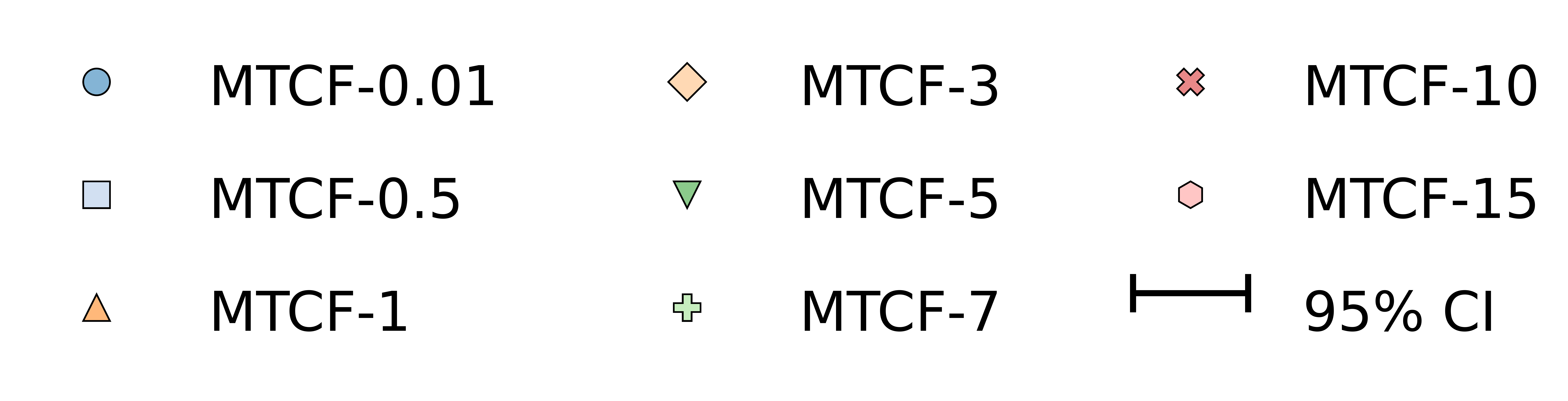}
    \caption{Performance of MTCF with different decision thresholds at rainfall thresholds of 15, 20, and 30 $\mathrm{mm}\cdot\mathrm{h}^{-1}$: (a) RMSE, (b) ME, (c) POD, (d) FAR, and (e) ETS. Error bars denote 95\% paired-bootstrap confidence intervals.}
    \label{fig:mtcf_ci}
\end{figure*}

The performance of MTCF is sensitive to its decision threshold. To select a representative configuration for comparison with Hurdle--RMIL, MTCF is evaluated with decision thresholds of 0.01, 0.5, 1, 3, 5, 7, 10, and 15 over the pooled test set. The point estimates in Fig.~\ref{fig:mtcf_ana} show that no single setting dominates all metrics and rainfall thresholds. MTCF-0.01 yields the highest ETS at rainfall thresholds of 0.1--3 $\mathrm{mm}\cdot\mathrm{h}^{-1}$, whereas MTCF-15 yields the lowest RMSE at 3--30 $\mathrm{mm}\cdot\mathrm{h}^{-1}$ and the smallest magnitude of negative ME at 0.1--30 $\mathrm{mm}\cdot\mathrm{h}^{-1}$. However, the improved RMSE and ME obtained with MTCF-15 are accompanied by a pronounced reduction in ETS at the 0.1 $\mathrm{mm}\cdot\mathrm{h}^{-1}$ rainfall threshold, from 0.301 for MTCF-0.01 and 0.279 for MTCF-0.5 to 0.153. MTCF-0.5 provides more balanced detection performance across the evaluated range: it achieves the highest ETS at rainfall thresholds of 15 and 20 $\mathrm{mm}\cdot\mathrm{h}^{-1}$ and an ETS of 0.015 at 30 $\mathrm{mm}\cdot\mathrm{h}^{-1}$, close to the highest value of 0.016 obtained with MTCF-15, while retaining substantially higher ETS at low rainfall thresholds than MTCF-15. As shown in Fig.~\ref{fig:mtcf_ci}, the RMSE confidence intervals for MTCF-15 and MTCF-0.5 are separated at rainfall thresholds of 15 and 20 $\mathrm{mm}\cdot\mathrm{h}^{-1}$ but overlap at 30 $\mathrm{mm}\cdot\mathrm{h}^{-1}$, indicating that the RMSE advantage of MTCF-15 is better supported at the former two thresholds. For ME, the intervals are separated at 15 $\mathrm{mm}\cdot\mathrm{h}^{-1}$ but overlap at 20 and 30 $\mathrm{mm}\cdot\mathrm{h}^{-1}$, so the reduction in underestimation indicated by MTCF-15 is more clearly supported at 15 $\mathrm{mm}\cdot\mathrm{h}^{-1}$. Their POD, FAR, and ETS intervals overlap at all three rainfall thresholds, so the uncertainty estimates do not establish a consistent detection-skill advantage for either setting. MTCF-0.5 is therefore adopted in the main comparison as a representative configuration that maintains comparatively strong ETS across both low and high rainfall thresholds, rather than as the optimum for every individual metric.
 
\section*{Acknowledgment}
AHI/Himawari-8 data are distributed by the National Institute of Information and Communications Technology, Japan (\url{https://sc-web.nict.go.jp/himawari/himawari-data-archive.html}). The rain-gauge observations were obtained from the Tianqing Meteorological Big Data Cloud Platform of the China Meteorological Administration. The authors thank the editor and the anonymous reviewers for their useful advice on this research.

\ifCLASSOPTIONcaptionsoff
  \newpage
\fi

\IEEEtriggeratref{16}


\clearpage
\makeatletter
\global\@colroom\@colht
\global\vsize\@colroom
\makeatother
\bibliographystyle{IEEEtran}
\bibliography{ref}
\end{document}